\documentclass[10pt,journal,compsoc]{IEEEtran}
\ifCLASSOPTIONcompsoc
  \usepackage[nocompress]{cite}
\else
  \usepackage{cite}
\fi

\ifCLASSINFOpdf
\else
\fi
\usepackage{graphicx}
\usepackage[colorlinks,linkcolor=blue]{hyperref}
\usepackage{amsmath}
\usepackage{multirow}

\usepackage{array}
\newcolumntype{C}[1]{>{\centering\let\newline\\\arraybackslash\hspace{0pt}}m{#1}}
\usepackage{amssymb}
\usepackage{amsfonts}
\usepackage{algorithmic}
\usepackage{xcolor}
\newcommand{\specialcell}[2][c]{%
  \begin{tabular}[#1]{@{}c@{}}#2\end{tabular}}
\ifCLASSOPTIONcompsoc
 \usepackage[caption=false,font=footnotesize,labelfont=sf,textfont=sf]{subfig}
\else
  \usepackage[caption=false,font=footnotesize]{subfig}
\fi
\begin{document}
\title{Learning Content-Weighted Deep Image Compression}
%
%
%
%

\author{Mu~Li,
        Wangmeng~Zuo,~\IEEEmembership{Senior Member,~IEEE},
        Shuhang~Gu,
        Jane~You,~\IEEEmembership{Member,~IEEE},
        and~David~Zhang,~\IEEEmembership{Fellow,~IEEE}
\IEEEcompsocitemizethanks{\IEEEcompsocthanksitem Mu Li and Jane You are with the Department
of Computing, Hong Kong Polytechnic University, Hong Kong, e-mail: (csmuli@comp.polyu.edu.hk; csyjia@comp.polyu.edu.hk).
\IEEEcompsocthanksitem Wangmeng Zuo is with the School
of Computer Science and Technology, Harbin Institute of Technology, Harbin, 150001, China and is also with Peng Cheng Lab, Shenzhen, China, e-mail: (cswmzuo@gmail.com).
\IEEEcompsocthanksitem Shuhang Gu is with Computer Vision Laboratory, ETH Zurich, Switzerland, e-mail: (shuhang.gu@vision.ee.ethz.ch).
\IEEEcompsocthanksitem David Zhang is with the School of Science and Engineering, The Chinese University of Hong Kong (Shenzhen), Shenzhen, China, e-mail: (davidzhang@cuhk.edu.cn).}
\thanks{Manuscript received xxx; revised xxx}}

%
%

\markboth{The IEEE Transactions on Pattern Analysis and Machine Intelligence }%
{Shell \MakeLowercase{\textit{et al.}}: Bare Advanced Demo of IEEEtran.cls for IEEE Computer Society Journals}

\IEEEtitleabstractindextext{%
\begin{abstract}
Learning-based lossy image compression usually involves the joint optimization of rate-distortion performance.
Most existing methods adopt spatially invariant bit length allocation, and incorporate discrete entropy approximation to constrain compression rate.
Nonetheless, the information content is spatially variant, where the regions with complex and salient structures generally are more essential to image compression.
{Taking} {the} spatial variation of image content into account, this paper presents a content-weighted encoder-decoder model, which involves an importance map subnet to {produce the importance mask for} locally adaptive bit rate allocation.
Consequently, the summation of importance {mask} can thus be utilized as {an} alternative of entropy estimation for compression rate control.
Furthermore, the quantized representations of the learnt code and importance map are still spatially dependent, which can be losslessly compressed using arithmetic coding.
To compress the codes effectively and efficiently, we propose a trimmed convolutional network to predict the conditional probability of quantized codes.
Experiments show that the proposed method can produce visually much better results, and performs favorably in comparison with deep and traditional lossy image compression approaches.
%
%
\end{abstract}

}

\maketitle

\IEEEdisplaynontitleabstractindextext

%
\IEEEpeerreviewmaketitle

\ifCLASSOPTIONcompsoc
\IEEEraisesectionheading{\section{Introduction}\label{sec:introduction}}
\else
\section{Introduction}
\label{sec:introduction}
\fi
Inspired by the unprecedented success of deep learning, {deep} image compression models recently has received considerable attention from the vision and learning communities.
Traditional image encoding standards, e.g., JPEG~\cite{wallace1992jpeg}, JPEG 2000~\cite{skodras2001jpeg}, and BPG (intra-frame encoding of HEVC)~\cite{sullivan2012overview}, generally adopt handcrafted transform and separate optimization on codecs{, thus are limited in compression performance and are inflexible in adapting to image content and tasks}.
In comparison, deep networks provide a flexible and end-to-end manner to learn nonlinear analysis and synthesis transforms jointly by optimizing rate-distortion performance, thereby expectantly surpassing existing codecs by compression performance and visual quality.
Moreover, recent years have witnessed the consistent progress and pursuit for the acquisition and sharing of higher definition images and videos.
And the population of the smart phones and Internet further increases the burden {on} storage and network bandwidth, {which further makes} it demanding to develop better image compression methods.



Learning-based lossy image compression is usually formulated as a joint rate-distortion optimization problem, and cannot be readily solved by deep networks.
On the one hand, quantization operation generally is indispensable to generate discrete codes.
However, its gradient is zero almost everywhere except it is infinite for several threshold points, making it challenging to jointly optimize encoder and decoder via back-propagation.
To handle this issue, several continuous proxy methods have been presented, including variational relaxation~\cite{toderici2015variable,balle2016end}, smoothed~\cite{theis2017lossy} and soft-to-hard approximation~\cite{agustsson2017soft}.
On the other hand, for joint rate-distortion optimization, rate loss usually is introduced for modeling the {entropy} of quantized codes, and also requires continuous approximation~\cite{balle2016end,theis2017lossy,agustsson2017soft}.

\begin{figure*}
\centering
\begin{minipage}{1.0\textwidth}
\begin{minipage}{0.242\textwidth}
\centering{\scriptsize{bpp / PSNR (dB) / MS-SSIM}}
\end{minipage}
\begin{minipage}{0.242\textwidth}
\centering{\scriptsize{0.217 / 31.41 / 0.956}}
\end{minipage}
\begin{minipage}{0.242\textwidth}
\centering{\scriptsize{0.208 / 32.69 / 0.968}}
\end{minipage}
\begin{minipage}{0.242\textwidth}

\end{minipage}
\end{minipage}\par\medskip

\begin{minipage}{1.0\textwidth}
\begin{minipage}{0.242\textwidth}
\includegraphics[width=1.0\textwidth]{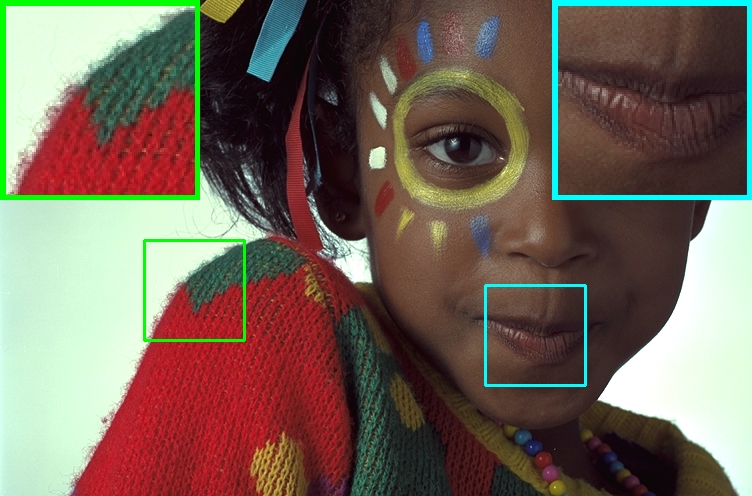}
\end{minipage}
\begin{minipage}{0.242\textwidth}
\includegraphics[width=1.0\textwidth]{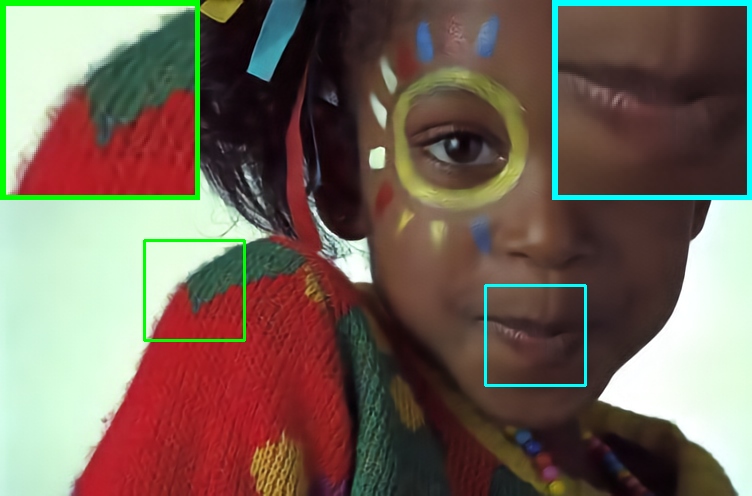}
\end{minipage}
\begin{minipage}{0.242\textwidth}
\includegraphics[width=1.0\textwidth]{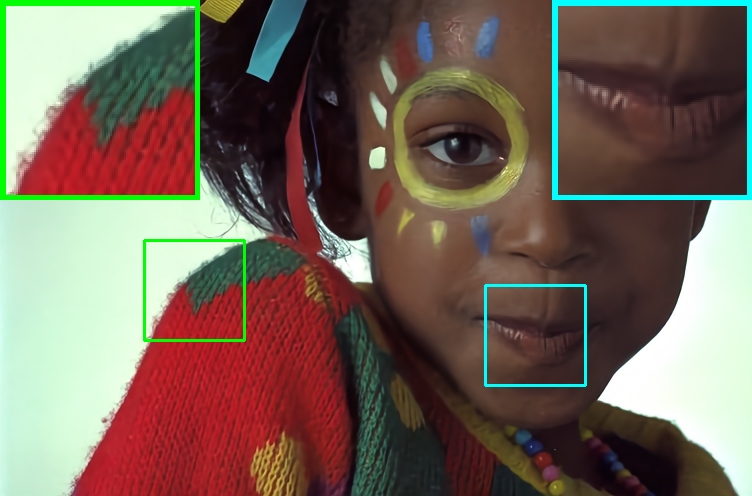}
\end{minipage}
\begin{minipage}{0.242\textwidth}
\includegraphics[width=1.0\textwidth]{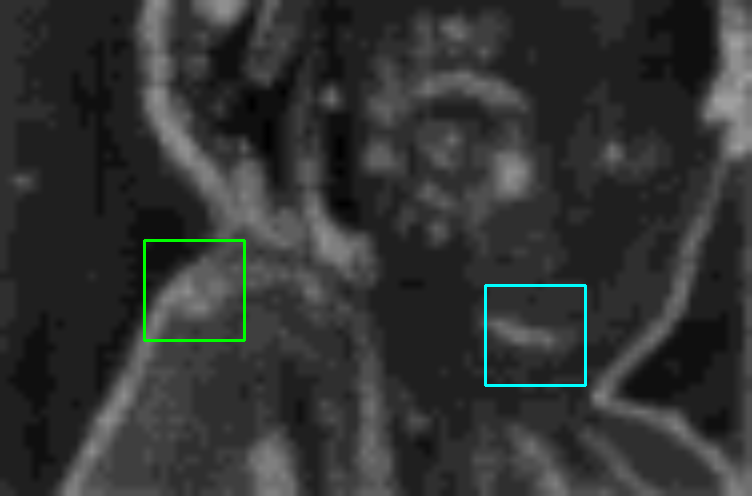}
\end{minipage}
\end{minipage}\par\medskip

\begin{minipage}{1.0\textwidth}
\begin{minipage}{0.242\textwidth}
\centering{\scriptsize{(a) Original}}
\end{minipage}
\begin{minipage}{0.242\textwidth}
\centering{\scriptsize{(b) Ball{\'e} et al.~\cite{balle2016end}}}
\end{minipage}
\begin{minipage}{0.242\textwidth}
\centering{\scriptsize{(c) Ours(MSE)}}
\end{minipage}
\begin{minipage}{0.242\textwidth}
\centering{\scriptsize{(d) Importance map}}
\end{minipage}
\end{minipage}
\caption{Decoding images of deep compression approach Ball{\'e} et al.~\cite{balle2016end} and our content-weighted image compression method.}\label{fig1}
\end{figure*}

Albeit their significant progress, existing learning-based image compression methods are still limited in modeling and exploiting spatial variation and dependency of image content.
First, spatially invariant bit length allocation is generally adopted in existing methods, whereas undoubtedly the content of an image is spatially variant.
That is, the regions with complex and salient structures generally are more essential to constitute an image.
For example, it can be seen from Fig. \ref{fig1} that Ball{\'e} et al.~\cite{balle2016end} fail to recover the details of the mouth and sweater at lower bits per pixel (bpp).
With the guidance of the importance map (see Fig. \ref{fig1}(d)) learnt from the image content, our method can assign more bits to encode the areas with more details and structural information and thus can generate visually better decoding {image} (see Fig. \ref{fig1}(c)).
Second, the {entropy} is usually calculated by assuming that quantized codes follow some specific form of distribution.
The mismatch between the assumption and the real distribution will inevitably have an adverse effect on compression performance.
Last but not the least, the codes after quantization are still spatially dependent.
Ball{\'e} et al.~\cite{balle2016end} simply adopt the context-based adaptive binary arithmetic coding framework (CABAC)~\cite{marpe2003context} for entropy encoding.
Considering the practical feasibility of maintaining conditional probability tables, CABAC only exploits the two nearest codes as context.
Nonetheless, better compression performance can be attained by incorporating larger context in entropy encoding.

To address the above issues, we in this paper take both spatial variation and dependency of image content into account, and present a content-weighted encoder-decoder model for deep image compression (i.e., CWIC).
For handling spatially variant image content, the encoder of our CWIC involves an encoder subnet, a learnt quantization operation and an importance map subnet.
In particular, the quantization operation is learnt to quantize each {element} from one of the $n$ feature maps into $T$ discrete levels by minimizing the quantization error.
And the importance map subnet is deployed to estimate the informative importance $p_{i,j}$ of local image content at location $(i,j)$.
Specifically, when $\frac{l}{L} \leq {p}_{i,j} < \frac{l+1}{L}$, we only encode and save the first $\frac{nl}{L}$ channels at spatial location $(i,j)$, where $L$ is the number of the importance level.
The learnt importance map can be exploited to {produce the importance mask $\mathbf{m}$} for guiding locally adaptive bit rate allocation.
{Codes with the importance mask $m_{kij}=1$ are saved, while those $m_{kij}=0$ are ignored.}
Thus, we can allocate more {codes} to the region with rich and salient structures and less {codes} to the smooth region, thereby benefiting the reconstruction of texture details with less sacrifice of bit rate {({see} Fig.~\ref{fig1}(c))}.
Moreover, the summation of importance map {$\sum_{k,i,j} {m}_{kij}$ naturally serves as an estimation} of compression rate, and our CWIC model can be learnt without any assumption on the distribution of quantized codes.

For exploiting local spatial context, we adopt the arithmetic coding framework, and present a convolutional entropy {prediction model} to predict the current {symbol} from its context for quantized codes as well as importance map.
Existing methods generally suffer from either storage or computational burden, and are limited in large context modeling and compression performance.
To tackle this dilemma, we present a trimmed convolutional network for arithmetic encoding (TCAE), where convolutional kernels are specially trimmed to respect the compression order and context dependency.
Then, the probability prediction of all symbols can be efficiently performed in one single forward pass via a fully convolutional network.
By stacking several trimmed convolution layers, TCAE can model large context while maintaining computational efficiency.
Furthermore, an inclined TCAE model is presented to divide the codes from a 3D code map into several inclined planes.
Parallel decoding can then be safely conducted to the {symbols} inner each inclined plane, thereby significantly speeding up the decoding process.

Experiments are conducted to evaluate our CWIC model on the Kodak PhotoCD image dataset\footnote[1]{http://r0k.us/graphics/kodak/} {and the Tecnick dataset\footnote[2]{https://testimages.org/}}.
In terms of MS-SSIM, our CWIC clearly outperforms existing image encoding standards, i.e., JPEG~\cite{skodras2001jpeg}, JPEG 2000~\cite{wallace1992jpeg} and BPG~\cite{sullivan2012overview}, and several deep image compression methods, e.g.,~\cite{balle2016end, rippel2017real, johnston2017improved, mentzer2018conditional1}.
In terms of PSNR, our CWIC performs on par with BPG~\cite{sullivan2012overview} and surpasses the other competing methods.
As for visual quality, our CWIC is promising in retaining fine salient details and suppressing visual artifacts in comparison to the competing methods.

This paper is a substantial extension of our pioneer work~\cite{li2017learning}.
Compared with~\cite{li2017learning}, an improved network structure combining dense blocks is introduced in encoder and decoder.
And binary quantization is substituted with a learnt channel-wise multi-value quantization for adaptive discretion of encoder feature.
Moreover, a two-stage relaxation scheme {is} adopted for better learning of the importance map subnet.
Finally, we introduce a TCAE as well as an inclined TCAE to effectively and efficiently model large context in arithmetic coding.
The contributions of this work are summarized as follows:
\begin{itemize}
  \item A content-weighted encoder-decoder model is introduced for lossy image compression.
        Here, a learnt channel-wise quantization is deployed for the discretion of {the encoder features}, an importance map subnet is introduced to {guide} locally adaptive bit allocation, the summation of {the generated} importance {mask} is used as an estimation of compression rate, and a two-stage relaxation scheme is deployed for learning importance map subnet.
  \item A TCAE network is presented for large context modeling in arithmetic encoding.
        With trimmed convolution, the conditional probability of quantized codes can be efficiently predicted via fully convolutional network.
        And an inclined TCAE model is further {introduced} to accelerate the decoding process.
  \item Experiments show that our method is effective in recovering salient structures and rich details while suppressing visual artifacts at lower bpp.
      Moreover, our method performs favorably in comparison to existing image encoding standards~\cite{skodras2001jpeg,wallace1992jpeg,sullivan2012overview} and deep models~\cite{balle2016end,rippel2017real,johnston2017improved,mentzer2018conditional1}.
\end{itemize}

The remainder of this paper is organized as follows.
Section \ref{sec:related_work} briefly reviews the relevant works of deep networks for lossy image compression and context modeling.
Sections \ref{sec:cwic} and \ref{sec:tcae} respectively present our CWIC and TCAE models for handling
the spatial variation and dependency issues in image compression.
Section \ref{sec:experiment} gives the experimental results, and Section \ref{sec:conclusion} provides some concluding remarks.

\section{Related Work}\label{sec:related_work}
%
%
%
%
%
%
%
%
Deep networks have achieved unprecedented success in many low level vision tasks such as image denoising~\cite{xie2012image,zhang2017beyond}, single image super-resolution (SISR)~\cite{dong2014learning,dong2016image} and image compression artifacts removal~\cite{dong2015compression}.
In order to utilize deep networks in image compression, considerable studies have been given to relax the quantization operation and rate loss, and to model spatial context in arithmetic coding, which will be briefly surveyed in this section.

\subsection{Deep Networks for Lossy Image Compression}
%
In the encoder-decoder framework, both recurrent neural network (RNN) and conventional network (CNN) based models have been developed for lossy image compression.
%
%
In~\cite{toderici2015variable}, Toderici et al. suggest a RNN architecture to compress the residual of a $32 \times 32$ images in a progressive manner.
Subsequently, they~\cite{toderici2016full} present a new variant of gated recurrent unit (GRU) as well as content-based residual scaling for RNN-based full-resolution image compression.
Three improvements are further introduced in~\cite{johnston2017improved}, i.e., improved network architecture, spatially adaptive bit allocation, and SSIM-weighted loss.
In contrast to our pioneer work~\cite{li2017learning}, they~\cite{johnston2017improved} simply adopt a handcrafted spatially adaptive bit allocation post-processing scheme.


Using variational auto-encoder (VAE), Ball{\'e} et al.~\cite{balle2016end} adopt {a} uniform noise approximation for modeling quantization error, and present a continuous and differentiable proxy of {the} rate-distortion loss.
%
%
Supposing that all the codes in one feature map are independent, a linear piece-wise probability density function (PDF) is learnt for each channel to estimate the entropy of the codes.
In~\cite{balle2018variational}, they further introduce a scale hyperprior as side information for capturing spatial dependency, and then model the entropy of the codes conditioned on the learnt hyperprior.
Furthermore, Minnen et al.~\cite{minnen2018joint} utilize a single PixelCNN layer for modeling autoregressive priors, and combine it with hyperprior for boosting rate-distortion performance.
%

Convolutional auto-encoders have also been studied and applied to lossy image compression.
Theis et al.~\cite{theis2017lossy} provide a continuous upper bound of entropy rate, and replace rounding function with its smooth approximation in backward propagation.
For modeling the distribution of encoder feature, they adopt Gaussian scale mixtures.
Agustsson et al.~\cite{agustsson2017soft} introduce a soft relaxation of {the} quantization and entropy, and adopt a soft-to-hard annealing scheme in training.
Rippel et al.~\cite{rippel2017real} incorporate {a} deep auto-encoder with pyramidal decomposition and adaptive code length regularization to develop a real-time high performance compression system.
Adversarial loss~\cite{goodfellow2014generative} is also considered in~\cite{rippel2017real,agustsson2018extreme} to generate visually realistic decoding images for low bit rates.
%
%
Soon after our pioneer work~\cite{li2017learning} and concurrently with~{\cite{li2018efficient}}, Mentzer et al.~\cite{mentzer2018conditional1} introduce a masked {convolutional} network for capturing spatial dependency in entropy model, and suggest a learning scheme by alternatingly updating the entropy network and auto-encoder in training.


\subsection{Deep Context Modeling in Entropy Encoding}
According to Shannon's source coding theorem~\cite{Shannon1948}, the optimal code length for a symbol should be $-\log_b P$.
Here, $b$ is the number of symbols used to generate output codes and $P$ is the probability of the input symbol.
In most deep network-based image compression systems, the codes are assumed to be i.i.d. and follow some specific form of distribution to ease the entropy computation~\cite{balle2016end,theis2017lossy,agustsson2017soft,rippel2017real}.
Nonetheless, it is obvious that the codes are spatially dependent and are stored in a sequential order.
Thus, better $P$ can be attained by predicting $P$ conditioned on the preceding encoded symbols (i.e., context).
%
In~\cite{balle2016end}, CABAC~\cite{marpe2003context} is adopted to further compress the codes in a lossless manner, but it {only considers} the two nearest codes and is unable in modeling large context.
%


Deep networks have made considerable progress for modeling large context of natural language and images.
For natural language, RNNs~\cite{mikolov2010recurrent,sundermeyer2012lstm} have exhibited great power in modeling long-range dependency.
As for images, PixelCNN~\cite{oord2016conditional} and PixelRNN~\cite{oord2016pixel} are also introduced to capture highly complex and long-range dependency between pixels.
In deep image compression system, the codes usually can be represented as {a} {3D cuboid}.
Toderici et al.~\cite{toderici2016full} adopt a BinaryRNN combining CNN and LSTM to model the context of the codes from both the current coding plane and the previous ones.
However, RNN based models generally require one forward pass to estimate the probability of each pixel (or bit), and are computationally very heavy in image generation and compression.
%

In our pioneer work~\cite{li2017learning}, we extract a fixed length context for each position, and learn a shallow convolutional {network} to predict the probability of current bit.
Even though { the} convolutional entropy {prediction model}~\cite{li2017learning} is more efficient than RNN based models, it performs probability prediction independently without shared computation and remains inefficient.
Subsequently, trimmed (masked) convolutional networks are suggested in our work~\cite{li2017learning} and Mentzer et al.~\cite{mentzer2018conditional1} to perform probability prediction of all symbols within one single forward pass.
We further extend trimmed convolution to its multi-group form, and present an inclined TCAE model which allows parallel decoding inner one inclined plane to speed up the decoding process.
%

\section{Content-weighted Image Compression}\label{sec:cwic}
In this section, we present a content-weighted encoder-decoder network for image compression (i.e., CWIC).
To begin with, we first describe the network structure of our CWIC, including encoder, decoder, and importance map subnets.
Then, distortion and rate losses are defined on the {decoding image} as well as importance map.
Finally, to easy the difficulty caused by quantization in CWIC, continuous relaxations are introduced for learning the encoder and importance map subnets.
\begin{figure*}[!t]
\center
\includegraphics[width=1.0\textwidth]{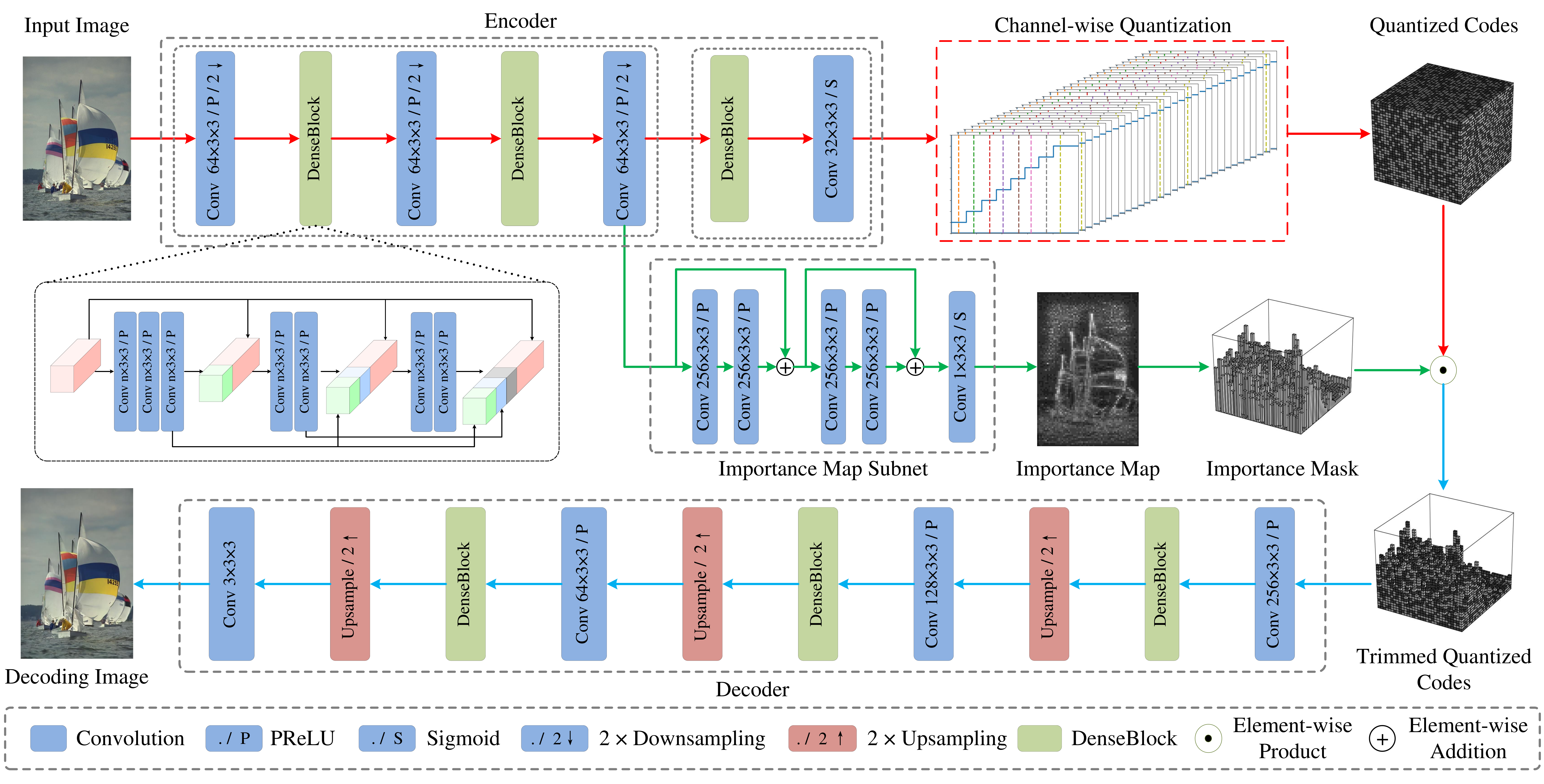}
\caption{{Illustration of our content weighted image compression model.
The whole framework involves an encoder, a learnt channel-wise multi-valued quantization, an importance map subnet and a decoder.
The encoder produces $32$ feature maps which {are} further quantized by the channel-wise multi-valued quantization function to generate quantized codes.
%
%
The importance map subnet {estimates} the informative importance of local image content and generate an importance map with only $1$ channel.
With the quantized importance map, an importance mask is further generated for guiding spatially variant bit rate allocation.
By multiplying quantized codes with importance mask in an element-wise manner, the trimmed quantized codes {are} produced as the input of the decoder to generate the decoding image.}}\label{fig:framework}

\end{figure*}

\subsection{Network Architecture}\label{network_architecture}
As illustrated in Fig.~\ref{fig:framework}, our CWIC network consists of three subnetworks, i.e., encoder, importance map and decoder subnets.
In particular, the encoder is further divided into the shared and encoding-specific parts.
To generate discrete codes, quantization operations are deployed to the outputs of encoder and importance map subnets.
In the following, we introduce the main network components, quantization operations, and encoding/decoding procedure.



\subsubsection{Encoder and decoder subnets}\label{encoder_and_decoder}
Given an image $\mathbf{x}$, the encoder subnet $E(\mathbf{x})$ is comprised of a shared part $E_s$ and {an} encoding-specific part $E_p$.
Concretely, {$E_s$  has three convolution layers with stride 2.}
And the feature map channels of the three layers are 64, 128, and 256, respectively.
{Moreover, a dense block is deployed right after each of the first two strided convolution layers to increase the nonlinearity of the encoder.
For $E_p$, it only contains one dense block.}
From Fig.~\ref{fig:framework}, each dense block involves three sub-blocks, where the first sub-block consists of three convolution layers and each of the other two sub-blocks consists of two convolution layers.
Following DenseNet~\cite{huang2017densely}, skip connections are introduced from any sub-block to all successive sub-blocks to improve information flow and ease the training of deep networks, thereby benefiting compression performance.
Analogous to~\cite{lim2017enhanced} in SISR, we remove the batch normalization operations from the sub-blocks, and empirically find that it is helpful in suppressing visual compression artifacts in smooth areas.
After the dense block in $E_p$, we further add an extra convolution layer with sigmoid nonlinearity to reduce the channels to $32$ and constrain the output within the interval $(0,1)$.
For simplicity, $3 \times 3$ convolution is adopted in all convolution layers.

%

The structure of the decoder subnet $D(\mathbf{c})$ is a mirror of the encoder.
In particular, convolution layer with stride 2 in the encoder subnet is substituted by a upsampling convolution layer, which involves a basic convolution layer followed by a depth-to-width operation~\cite{toderici2016full}.
And the last convolution layer produces the decoding image with 3 channels and linear activation is used.

\subsubsection{Importance map subnet} \label{sec:importance_map}
\begin{figure}[!ht]
\center
\includegraphics[width=1.0\linewidth]{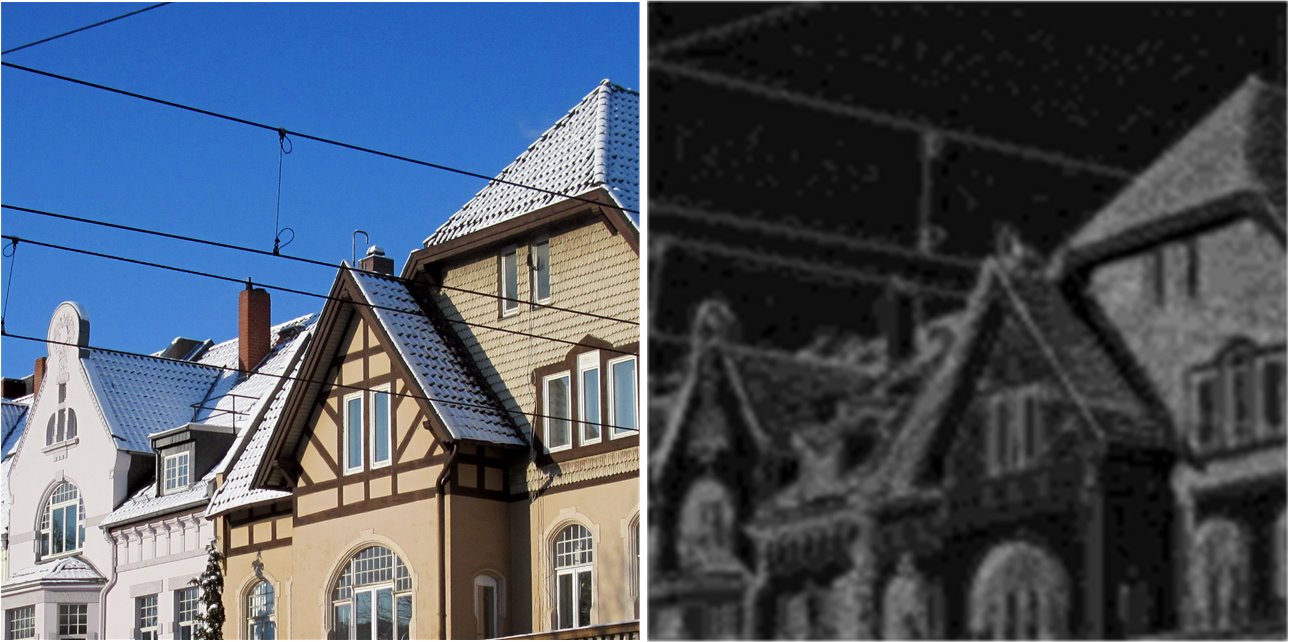}

\begin{minipage}{0.49\linewidth}
\centering
\scriptsize{(a) Original image}
\end{minipage}
\begin{minipage}{0.49\linewidth}
\centering
\scriptsize{(b) Importance map (0.213 bpp)}
\end{minipage}
\caption{Illustration of the importance map. The regions with sharp edge and rich texture generally have higher values and should be allocated with more bits.}\label{fig:imp_illustration}
\end{figure}

In general, an image conveys spatially variant informative content.
From Fig.~\ref{fig:imp_illustration}, the regions with house are more salient and content-intensive, while the regions with sky are simple and contain little informative content.
Most deep image compression methods~\cite{balle2016end,agustsson2017soft,theis2017lossy} allocate spatially invariant code length and exploit entropy coding to further compress the the codes.
Although entropy coding is able to encode quantized codes into bits with different length, it is deployed after quantization operation, and cannot recover the information loss caused by spatially invariant bit allocation in quantization.
As a result, such solutions usually are inferior in preserving salient structure and fine details at lower bpp.

To alleviate this issue, we suggest to utilize spatially variant bit allocation which can be more advantageous by emphasizing salient structures of image.
As shown in Fig.~\ref{fig:imp_illustration}, it is reasonable to allocate more bits to region \emph{house} and fewer bits to region \emph{sky}.
Therefore, we introduce an importance map subnet to produce importance map from the input image.
It can be seen from Fig.~\ref{fig:imp_illustration}(b) that, importance map provides a reasonable estimation of the informative importance of local image content, and can be exploited to guide locally adaptive bit rate allocation.
In comparison to spatially invariant bit rate allocation, we can add more channels of feature maps and incorporate with importance map to preserve more salient structure and fine details without the increase of code length.
%
%
Note that the code length at each position is controlled by the importance map.
Thus, the summation of {quantized importance map, i.e., importance mask,} can serve as an estimation of compression rate, and be computed without any assumption on the distribution of quantized codes.
It is worth noting that just noticeable distortion (JND) models~\cite{yang2005just} have also been suggested for spatially variant bit allocation and rate control in video coding.
In contrast to JND~\cite{yang2005just}, the importance map is learnt from training data via rate-distortion optimization.

The architecture of importance map subnet is shown in Fig.~\ref{fig:framework}.
In particular, it takes the intermediate feature map $E_s(\mathbf{x})$ as input, involves {two residual blocks~\cite{He_2016_CVPR} and an extra convolution layer} with sigmoid nonlinearity to produce the 1-channel importance map $\mathbf{p} = P(E_s(\mathbf{x}))$ which has the same spatial size, i.e., $h \times w$, {as} the encoder feature map $\mathbf{e}$ with the values in the range $(0, 1)$.


\subsubsection{Quantization}
Both the encoder feature map $\mathbf{e}$ and importance map $\mathbf{p}$ are continuous values in the range $(0,1)$, and quantization is required.
For $\mathbf{e}$, we adopt the channel-wise multi-valued quantization $Q$ parameterized by $\Theta_k = \{s_{k,0}, \ldots, s_{k,t}, \ldots, s_{k,T-1}\}$, where $s_{k,t}$s denote non-negative weights with each represents a quantization interval, and $T$ is the number of quantization levels.
With $\Theta_k$, the $t$-th quantization center $q_{k,t}$ of the $k$-th channel is defined as,
\begin{equation}
q_{k,t}=\sum\nolimits_{t^{\prime}=0}^t s_{k,t^{\prime}}\;.
\end{equation}
The quantized level $t^{*}(e_{kij})$ of an element $e_{kij}$ of the $k$-th channel can be obtained by,
\begin{gather}
\label{eq:quntization}
t^{*} = \arg \min_{t} \|e_{kij} - q_{k,t}\|^2,\;\; t=0, \ldots, T-1.
\end{gather}
Then, $e_{kij}$ is discretely represented as $Q(e_{kij})= q_{k,t^*}$, {and its quantization index  is represented as $QL(e_{kij})=t^{*}$}.



As for importance map $\mathbf{p}$, we define the following quantization function to quantize the importance value $p_{ij}$ at position $(i,j)$,
\begin{equation}\label{eq:imp_quantization}
QI(p_{ij})=l, \mbox{if}\;\frac{l}{L}\leq p_{ij} < \frac{l+1}{L} \;,\; l \in \{0,\ldots,L-1\},
\end{equation}
where $L$ is the number of quantization levels for importance map.
We note that the quantized importance map is also required to {be stored} in our encoding scheme.
Denote by $n$ the number of channels of encoder feature map $\mathbf{e}$.
Without loss of generality, we assume that $(n \mod L) = 0$.

For guiding spatially variant bit allocation, we further introduce a binary importance mask $M(\mathbf{p})$ with the same size as the quantized encoder feature map $Q(\mathbf{e})$.
In particular, the $(k,i,j)$-th element ${m}_{kij}$ of $\mathbf{m} = M(\mathbf{p})$ is defined as,
\begin{equation}\label{eq:mask}
{m}_{kij}=\begin{cases}
1, \;\; \mbox{if} \; k < \frac{n}{L}QI(p_{ij}),\\
0, \;\; \mbox{otherwise}.
\end{cases}
\end{equation}
Guided by $M(\mathbf{p})$, all the codes with ${m}_{kij} = 0$ are discarded from $Q(\mathbf{e})$.
When $QI(p_{ij})=0$, no code {needs to be stored} at position $(i,j)$, and all of its information is predicted from its context.
To sum up, instead of $n \times h \times w$, only {$\sum_i\sum_j \frac{n}{L}QI(p_{ij})=\sum_{k,i,j} {m}_{kij}$ codes from} $Q(\mathbf{e})$  {need to be stored}, and the summation of {importance mask} can thus be used as an indicator of compression rate.

\subsubsection{Procedure of encoding and decoding}
Finally, we summarize the procedure of encoding and decoding based on the encoder, importance map, and decoder subnets.
Given an input image $\mathbf{x}$, the shared part of encoder subnet is first deployed to generate intermediate feature map $E_s(\mathbf{x})$.
Then, both {the} encoder-specific part of encoder {subnet and the}  importance map subnet take $E_s(\mathbf{x})$ as input to produce encoder feature map $\mathbf{e} = E_p(E_s(\mathbf{x}))$ and importance map $\mathbf{p} = P(E_s(\mathbf{x}))$, respectively.

By quantizing $\mathbf{e}$ and $\mathbf{p}$, we obtain the discrete encoder representation $Q(\mathbf{e})$, quantized importance map $QI(\mathbf{p})$, and binary importance mask $M(\mathbf{p})$.
The encoding result of $\mathbf{x}$ can then be represented as $\mathbf{z} = M(\mathbf{p}) \circ Q(\mathbf{e})$, where $\circ$ denotes the element-wise product.
{The corresponding quantization index of $\mathbf{z}$ is represented as $\mathbf{o} =  M(\mathbf{p}) \circ QL(\mathbf{e})$. 
Then $\mathbf{o}$ and $QI(\mathbf{p})$ are stored as the codes of $\mathbf{x}$.
In the decoding stage, $\mathbf{z}$ is reconstructed by $z_{kij}=q_{k,o_{kij}}$ if $m_{kij}=1$, otherwise $z_{kij}=0$.
and the decoder subnet takes $\mathbf{z}$ as input to obtain the decoding image $D(\mathbf{z})$.
}

%

\subsection{Loss Functions}\label{loss}
In general, both distortion and rate losses should be included in model objective of content-weighted image compression.
Moreover, a quantization loss is also introduced to guide the learning of channel-wise multi-valued quantization.
In the following, we explain these loss functions and give the overall model objective.

\textbf{Distortion loss.} Distortion loss is used to measure the distortion between the input image and decoding image.
Concretely, we consider two types of distortion metrics.
The first is based on the mean squared error (MSE),
%
\begin{equation}
\mathcal{L}_{\text{MSE}}(\mathbf{z}, \mathbf{x}) = \frac{1}{3HW} \|D(\mathbf{z}) - \mathbf{x}\|^2_2 \;,
\end{equation}
%
where $H$ and $W$ are the height and width of $\mathbf{x}$, respectively.
The other is based on the multi-scale structural similarity (MS-SSIM)~\cite{wang2003multiscale},
\begin{equation}
\mathcal{L}_{\text{MS-SSIM}}(\mathbf{z}, \mathbf{x}) = 100 \left(1 - \mbox{MS-SSIM} \left( D(\mathbf{z}), \mathbf{x} \right) \right) \;.
\end{equation}
%
%
In our implementation, $\mathcal{L}_{\text{MS-SSIM}}$ is adopted as the default distortion loss $\mathcal{L}_{D}$, and we denote our method with $\mathcal{L}_{\text{MSE}}$ as Ours(MSE).

\textbf{Rate loss.} Benefited from importance map subnet, we define the rate loss directly on approximate code length.
Suppose the size of encoder feature map $E(\mathbf{x})$ is $n \times h \times w$.
The code by our model includes two parts:
(i) quantized importance map $QI(\mathbf{p})$ with the size $h \times w$;
(ii) the trimmed code with the size $\frac{n}{L}\sum_{i,j} QI({p}_{ij})$.
Note that the size of $QI(\mathbf{p})$ is constant given an image size.
Thus $\frac{n}{L}\sum_{i,j} QI({p}_{ij})$ can be used as an indicator of code length.


For better rate control, we introduce a threshold $r$ based on the expected code length for a given compression rate, and penalize the rate loss only when it is higher than $r$.
Then, we define the rate loss in our model as,
\begin{equation}
\mathcal{L}_{R}(\mathbf{x}) \!=\! \max \left\{0, (\frac{n}{L}\sum\nolimits_{i,j} QI({p}_{ij})- r \cdot nhw )\right\} \; .
\end{equation}
By this way, rate loss only penalizes the code length higher than {$r \cdot nhw$}, making the learnt compression system exhibit a comparable compression rate around the given one.

Considering that the number of trimmed codes is exactly equal to the number of $1$s in the importance mask, the ratio loss can be equivalently rewritten as:
{
\begin{equation}
\mathcal{L}_{R}(\mathbf{x}) \!=\! \max \left\{0, (\sum\nolimits_{k,i,j} m_{kij}- r \cdot nhw) \right\} \; .
\end{equation}
}

\textbf{Quantization loss.}
For better quantization of encoder feature map, we employ $\Theta_k$ to parameterize the multi-valued quantization for the $k$-th channel {\and incorporate a quantization loss for minimizing the squared $\ell_2$ error caused by quantization,}
%
%
\begin{equation}\label{eq:quantize-error}
\mathcal{L}_{Quant}=\frac{1}{nhw}\sum\nolimits_{k,i,j} \|Q(e_{kij})-e_{kij}\|^2.
\end{equation}
%


\textbf{Model objective.}
Let $\mathcal{X}$ be a set of training data, and $\mathbf{x} \in \mathcal{X}$ be an image from the set.
The overall learning objective is then defined as the combination of distortion, rate, and quantization losses,
\begin{equation}\label{equ:total_loss}
\mathcal{L} = \sum_{\mathbf{x} \in \mathcal{X}}\{\mathcal{L}_{D}(\mathbf{z}, \mathbf{x}) + \gamma \mathcal{L}_{R}(\mathbf{x}) + \eta \mathcal{L}_{Quant} \} ,
\end{equation}
where $\gamma$ and $\eta$ are tradeoff parameters for balancing the three loss terms.
Considering that quantization loss is not directly related with the rate-distortion performance, we empirically set $\eta = 1$ and $\mathcal{L}_{Quant}$ is deployed to only update quantization parameters in training.


\subsection{Relaxation of Quantization for Model Learning}
As noted above, due to the quantization operations, conventional back-propagation algorithm is not applicable to learn the encoder and importance map subnets.
To circumvent this issue, two relaxation approaches are presented.
To relax the quantization of feature map, a proxy function based on straight-through estimator is introduced to approximate the channel-wise quantization in backward propagation.
To relax the quantization of importance map, {a} two-stage relaxation shceme is adopted to train importance map subnet.

\subsubsection{Relaxation and learning of channel-wise multi-valued quantization}\label{sec:lcmq}							
The gradients of the learnt channel-wise multi-valued quantization function are zeros almost everywhere and are infinite at several threshold points.
Such non-differentiable property inevitably restricts the backward propagation of gradients from decoder to encoder, and gives rise to the difficulty in learning the deep image compression system.
As a result, any layers before the quantization function (i.e., the whole encoder) are never updated during training.

Fortunately, some recent works on binarized neural networks (BNN)~\cite{zhou2016dorefa,rastegari2016xnor,courbariaux2016binarized} have studied the issue of propagating gradient through binarization, which can also be extended to relax multi-valued quantization.
Based on the straight-through estimator on the gradient~\cite{courbariaux2016binarized}, we introduce a linear proxy $\tilde{Q}(e_{kij})$ to approximate $Q(e_{kij})$,
\begin{equation}
\tilde{Q}(e_{kij})=e_{kij}.
\end{equation}
In particular, $Q(e_{kij})$ is still adopted in forward propagation, while $\tilde{Q}(e_{kij})$ is only used in backward-propagation.
The gradient of $\tilde{Q}(e_{kij})$ can then be easily obtained by,
\begin{equation}
\tilde{Q}^\prime(e_{kij})=1.
\end{equation}

Albeit that the proxy function can ease the difficulty of model learning, its effectiveness actually depends on the values of $e_{kij}$s in training.
When the quantization error is $0$, $Q(e_{kij})$ equals to $\tilde{Q}(e_{kij})$, and it is safe to use $\tilde{Q}(e_{kij})$ as proxy function.
Moreover, even the equality does not hold, the quantization loss in Eqn.~(\ref{eq:quantize-error}) can constrain that $\tilde{Q}(e_{kij})$ approximates $Q(e_{kij})$.
Thus, it is reasonable to use $\tilde{Q}(e_{kij})$ as  proxy of the learnt quantization function in practice.

\textbf{Initialization and re-initialization in learning.}
By assuming that encoder feature follows a uniform distribution in the range $[0, 1]$, we simply initialize $s_{k,t}$s as $s_{k,0}=\frac{1}{2T}$ and $s_{k,t}=\frac{1}{T}$ for $t>0$.
However, we empirically find that such initialization scheme may suffer from the dead point problem.
For example, when all $Q(e_{kij})$s from the $k$-th channel are lower than $q_{k,t_0}$, the gradients with respect to $s_{k,t_0},\ldots,s_{k,T-1}$ will be always zero, and the last few quantization levels, i.e., $q_{k,t_0},\ldots,q_{k,T-1}$, will never be optimized and used during training.
For handling this issue, we store the histogram of $q_{k,t}$s of each mini-batch.
When all the counts of $q_{k,t}$s with $t \geq t_0$ are zero in 50 successive mini-batches, we re-initialize the weights {$s_{k,t} = \frac{s_{k,t_0-1}}{T - t_0 + 1}$ for $t = t_0-1, \ldots, T-1$}.
As a result, $Q(e_{kij})$s with {$e_{kij} > q_{k,t_0-2}$} are more likely to be quantized to the last few quantization levels in future training, and the non-increasing quantization loss can also be guaranteed.

\subsubsection{Relaxation and learning of importance map}
Analogous to channel-wise quantization, straight-through estimator can also be used to relax the quantization of importance map which includes the generation of both quantized importance map and binarized importance mask.
However, we empirically find that such solution works well only when binarization is adopted to quantize encoder feature map. 
%
%
For better learning importance map subnet in general setting, we introduce a two-stage relaxation scheme,
Here, an alternative loss is used in the first stage to update $QI({p}_{ij})$, and another alternative loss is adopted in the second stage to update importance map subnet.

In the first stage, given the current code $\mathbf{z}^*$, by approximating $Q(e_{kij})-Q(e^*_{kij})$ with $e_{kij}-e^*_{kij}$, the distortion loss $\mathcal{L}_D$ is relaxed with its Taylor expansion w.r.t. $\mathbf{z}^*$,
\begin{eqnarray}
\mathcal{L}^\prime_D
&=&	\mathcal{L}_D(\mathbf{z}^*)+\sum_{k,i,j} m_{kij}\frac{\partial \mathcal{L}_D}{\partial z_{kij}}|_{\mathbf{z}=\mathbf{z}^*}(e_{kij}-e^*_{kij})\nonumber\\
s.t.&& |e_{kij}-e^*_{kij}|\leq \xi
\end{eqnarray}
where $\xi$ is a small positive value, and it is empirically set as $\xi=0.1$. By replacing the $\mathcal{L}_D$ with the proxy $\mathcal{L}^\prime_D$, we define the proxy function for $QI(\mathbf{p})$ as,
\begin{equation}
\mathcal{L}^\prime(QI(\mathbf{p})) = \mathcal{L}^\prime_D+\gamma \mathcal{L}_R.
\end{equation}
It is worth noting that, the minimization of $\mathcal{L}^\prime(QI(\mathbf{p}))$ w.r.t. $QI(\mathbf{p})$ can be decomposed into $h \times w$ subproblems on $QI({p}_{ij})$.
The loss function for each $QI({p}_{ij})$ can be represented as,
\begin{equation}				
\mathcal{L}^\prime_{ij}=\begin{cases}\sum_{k=0}^{n-1} m_{kij}t_{kij},\;\;\mbox{if}\;\; \sum_{k,i,j} m_{kij}< r \cdot nhw\\
\sum_{k=0}^{n-1} m_{kij}(t_{kij}+\gamma)-r \cdot n,\;\;\mbox{otherwise}
\end{cases}
\end{equation}
where $t_{kij}=(e_{kij}-e^*_{kij})\frac{\partial \mathcal{L}_D}{\partial z_{kij}}|_{\mathbf{z}=\mathbf{z}^*}$ with $|e_{kij}-e^*_{kij}|\leq \xi$.
Note that $\mathcal{L}^\prime_{ij}$ is a function of both $e_{kij}$ and $QI(p_{ij})$.
First, we only consider $\mathcal{L}^\prime_{ij}$ w.r.t. $e_{kij}$.
Due to that $m_{kij}$ is non-negative, it is obvious that the minimum of $t_{kij}$ should be $-\xi \left | \frac{\partial \mathcal{L}_D}{\partial z_{kij}}|_{\mathbf{z}=\mathbf{z}^*} \right |$.
Then, given $t_{kij} = -\xi \left | \frac{\partial \mathcal{L}_D}{\partial z_{kij}}|_{\mathbf{z}=\mathbf{z}^*} \right |$,
%
%
the minimization of $\mathcal{L}^\prime_{ij}$ w.r.t. $QI({p}_{ij})$ can be rewritten as:
\begin{equation}				
\mathcal{L}^{\prime}_{ij} \!\!=\!\! \begin{cases}-\xi \sum_{k=0}^{n-1}  m_{kij} \left|\frac{\partial \mathcal{L}_D}{\partial z_{kij}}|_{\mathbf{z} = \mathbf{z}^*} \right|,\;\;\mbox{if}\;\; \sum_{k,i,j} m_{kij} \!\!<\!\! r \!\cdot\! nhw\\
 -\xi\sum_{k=0}^{n-1}  m_{kij}( \left|\frac{\partial \mathcal{L}_D}{\partial z_{kij}}|_{\mathbf{z}=\mathbf{z}^*}\right|-\frac{\gamma}{\xi})-r \cdot n,\;\;\mbox{otherwise}
\end{cases}
\end{equation}
It is noted that $QI(p_{ij})$ only has $L$ possible values, i.e., $QI(p_{ij}) \in \{0,\ldots,L-1\}$.
Then, we define the $L$ different importance masks as $\{s_l = (\underbrace{1,\ldots,1}_{nl/L}, \underbrace{0,\ldots,0}_{n-nl/L})\;|\;l=0,\ldots, L-1\}$.
%
%
Thus, we simply test all possible $QI(p_{ij})$ values to find the optimal one for minimizing $\mathcal{L}^\prime_{ij}$,
\begin{equation}
l^* = \arg \min_l \mathcal{L}^\prime_{ij}.
\end{equation}

In the second stage, we introduce a continuous proxy based on $l^*$ for updating importance map subnet,
%
%
\begin{equation}
\mathcal{L}_{imp}=\alpha |l^* - p_{i,j} \cdot L| ,
\end{equation}
where $\alpha$ is a trade-off parameter and we set it to be 0.001 in our implementation.
Thus, the gradient w.r.t. $p_{i,j}$ can be obtained by,
\begin{equation}
\frac{\partial\mathcal{L}_{imp}}{\partial p_{i,j}}=\begin{cases}
-\alpha, \; \mbox{if} \; p_{i,j}<l^*/L\\
\alpha, \;\;\;\; \mbox{if} \; p_{i,j}>l^*/L\\
0, \;\;\;\;\; \mbox{otherwise}.
\end{cases}
\end{equation}

\vspace{-.1in}
\subsection{Implementation and Learning}

In our implementation, we set the {channel number} of code maps $n = 32$.
For the learnt channel-wise quantization function, we set the number of quantized values $T = 8$.
As for the importance map, the number quantization levels $L$ is set to $16$.
%
%
Without importance map and entropy coding, the whole code maps corresponds to a compression rate of $1.5$ bpp.
By setting specific threshold value $r$, our CWIC framework can be adapted to different compression rates without changing the network structure.
For the setting of $r$, we simply let $r = \frac{2}{3}r_0$, where {$r_0 \in \{ 0.1, 0.2, 0.3, 0.45, 0.6, 0.8, 1.00\}$} is the expected compression rate.
The parameter $\gamma$ controls the tradeoff between distortion and rate losses.
According to {$r_0 \in \{  0.1, 0.2, 0.3, 0.45, 0.6, 0.8, 1.00\}$}, we set {$\gamma \in \{ 1 \times 10^{-3}, 5 \times 10^{-4}, 2\times 10^{-4}, 1 \times 10^{-4}, 5\times 10^{-5}, 2\times 10^{-5}, 1\times 10^{-5} \}$} to make the gradient of $\mathcal{L}_R$ term comparable to that of $\mathcal{L}_D$ term during training.

{
The whole CWIC model is trained using the ADAM solver~\cite{kingma2014adam}. We initialize the model with the parameters pre-trained on the the training set $\mathcal{X}$ without the importance map subnet.
The model is further trained with the learning rate of $1\times 10^{-4}$, $1\times 10^{-5}$ and $1\times 10^{-6}$.
The smaller learning rate is adopted until the objective with the larger one keeps non-decreasing in five successive epochs.
}

\section{Trimmed Convolutional Network for Arithmetic Encoding}\label{sec:tcae}
{The code $\mathbf{o}$} and quantized importance map $QI(\mathbf{p})$ by the above CWIC model are still spatially dependent, and can be further losslessly compressed.
{For the code} $\mathbf{o}$, there are two kinds of $0$s, i.e., the $0$ in the quantization {index} and the $0$ generated by the element-wise product with importance mask.
Nonetheless, the former $0$s are informative, while the later should be ignored in entropy prediction but still can be used as the context of {the} other symbols.
To distinguish these two kinds of $0$s, we simply adopt  {$\mathbf{o}^\prime=(\mathbf{o}+1)\circ \mathbf{m}$} in our implementation.
For lossless compression, arithmetic coding~\cite{witten1987arithmetic} predicts the probability of the
current symbol to be encoded from its context, and is proved to be the optimal coding in approximating entropy-based compression rate~\cite{said2004introduction}.
Thus, we adopt the arithmetic coding framework, and present a trimmed convolutional network model for efficient modeling of large context.

\subsection{Coding Schedule and Context of 3D Cuboid}
\begin{figure}[tbp]
 \begin{center}
 \includegraphics[width=1.0\linewidth]{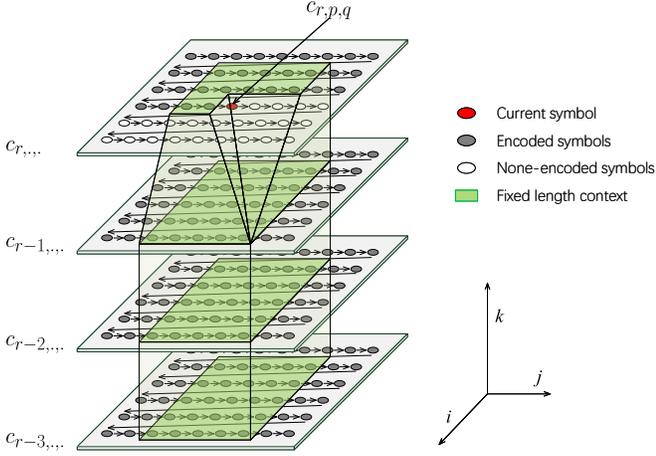}
 \end{center}
    \caption{{Coding schedule and context of 3D cuboid. The arrows indicate the encoding order of the cuboid. The green areas are the fixed length context of the symbol $c_{r,p,q}$ with $h_t=2$ and $w_t=2$. The red circle represents the current symbol $c_{r,p,q}$, and the gray (white) circles represent the encoded (non-encoded) symbols.}}\label{schedule}
\end{figure}

To begin with, we note that both {$\mathbf{o}^\prime$} and $QI(\mathbf{p})$ can be represented as a 3D cuboid {$\mathbf{C} = \{c_{k,i,j} | 0 \leq k \leq n-1, 0 \leq i \leq h-1, \mbox{and} \; 0 \leq j \leq w-1 \}$}.
Before TCAE, we first introduce the coding schedule and two types of context based on $\mathbf{C}$.
As illustrated in Fig.~\ref{schedule}, beginning at ${c}_{0,0,0}$, we adopt the following order to encode $\mathbf{C}$:
(i) ${c}_{k,i,j+1}$ is encoded after ${c}_{k,i,j}$ until $j = w-1$;
(ii) when $j = w-1$, ${c}_{k,i+1,0}$ is encoded after ${c}_{k,i,w-1}$ until $i = h-1$;
(iii) when $j = w-1$ and $i = h-1$, ${c}_{k+1,0,0}$ is encoded after ${c}_{k,h-1,w-1}$.

For a position $(r,p,q)$, we define its full context as,
$\mbox{CTX}({c}_{r,p,q}) = \{{c}_{k,i,j}| \{k<r\} \vee \{ k=r, i<p \} \vee \{k=r,i=p,j<q\}\}$, {i.e., all the gray circle in Fig.~\ref{schedule}}.
Unfortunately, the length of the full context $\mbox{CTX}({c}_{r,p,q})$ is unfixed and varies by the position $(r,p,q)$, making it difficult to learn {a} CNN-based probability prediction model based on $\mbox{CTX}({c}_{r,p,q})$.

Naturally, the context spatially close to the current symbol ${c}_{r,p,q}$ plays {a} more important role in probability prediction.
%
%
Taking these aspects into account, we give a fixed length context defined as $\mbox{CTX}_f({c}_{r,p,q}) = \{{c}_{k,i,j}| \{ r-c_t \leq k < r, |i-p| \leq h_t, |j-q| \leq w_t\} \vee \{ k=r, p-h_t \leq i < p, |j-q| \leq w_t \} \vee \{k=r,i=p, q-w_t \leq j < q\}\}$.
Furthermore, considering that different channels are not totally independent, we empirically suggest to set a larger $c_t$ or even include all channels in context modeling.
Fig.~\ref{schedule} gives an example of $\mbox{CTX}_f({c}_{r,p,q})$ with $h_t=2$ and $w_t=2$ for intuitive illustration.
%


%
In our pioneer work~\cite{li2017learning}, we extract a $(c_t+1) \times (2h_t+1) \times (2w_t+1)$ cuboid {$\mbox{CTX}_f^\prime({c}_{r,p,q})=\{{c}_{k,i,j} | \{ r-c_t \leq k \leq r, |i - p| \leq w_t, |j-q| \leq h_t \}$} with $c_t=3,h_t=2,w_t=2$.
For context modeling, the non-encoded symbols in $\mbox{CTX}_f^\prime({c}_{r,p,q})$ are replaced with 0s.
Then, a convolutional entropy {prediction model} of three convolution layers followed by three fully connected layers is introduced to predict ${c}_{r,p,q}$ from its context cuboid {$\mbox{CTX}_f^\prime({c}_{r,p,q})$}.

In \cite{li2017learning}, the contexts and non-encoded symbols are dynamically changed along with the encoding process.
Thus, convolutional entropy {prediction model} requires to perform probability prediction independently without shared computation, thereby remaining computationally expensive.
%
%
%
Consequently, {even though} the introduction of non-encoded symbols is necessary for context modeling, {it} also brings new difficulties to exploit fully convolutional network (FCN) for shared computation.
Next, we will present a group of trimmed convolutions to circumvent the inefficiency issue.

\vspace{-.1in}
\subsection{Trimmed Convolution}
The fixed length context in convolutional entropy encoder has two appealing properties, which can be exploited to perform probability prediction via trimmed convolutions.
(i) Given the current symbol ${c}_{r,p,q}$, the positions of all non-encoded symbols are fixed.
(ii) The default value for all non-encoded symbols is also a fixed number, and without loss of generality we can set it to be 0.

We first begin with the definition and analysis of standard convolution operator.
Denote by $\mathbf{w}^0$ a group of $n$ convolution kernels $\mathbf{w}^0 = \{ w_{t,k,i,j}| -w_0 \leq j \leq w_0, -h_0 \leq i \leq h_0, 0 \leq k < n, 0 \leq t < n\}$.
Then the convolution result $(\mathbf{C}*\mathbf{w}^0)$ at the location $(r,p,q)$ can be written as,
\begin{equation}
\label{eqn:convolution}
(\mathbf{C}*\mathbf{w}^0)(r,p,q) = \sum_{k, l-i = p, m-j = q} c_{k,l,m}w^0_{r,k,i,j}.
\end{equation}
where $*$ denotes the convolution operator.
However, such convolution treats the context and non-encoded symbols equally and cannot be applicable to context modeling.

\begin{figure}[!tbp]
\center
\begin{minipage}{1.0\linewidth}
\begin{minipage}{0.24\linewidth}
\center
\includegraphics[width=1.0\linewidth]{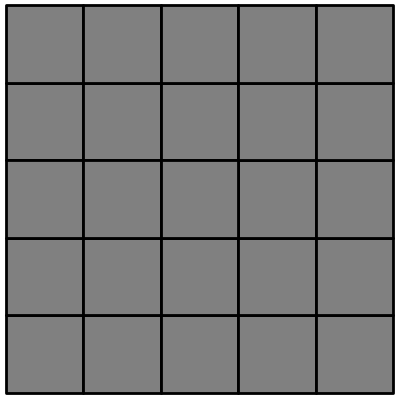}
\scriptsize{(b)}
\end{minipage}
\begin{minipage}{0.24\linewidth}
\center
\includegraphics[width=1.0\linewidth]{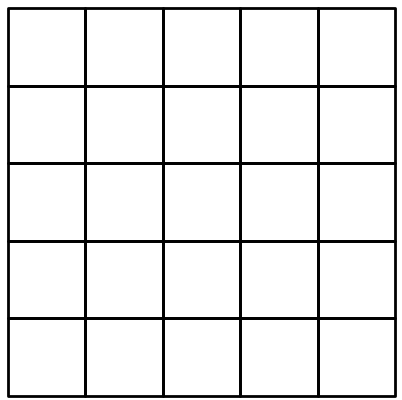}
\scriptsize{(b)}
\end{minipage}
\begin{minipage}{0.24\linewidth}
\center
\includegraphics[width=1.0\linewidth]{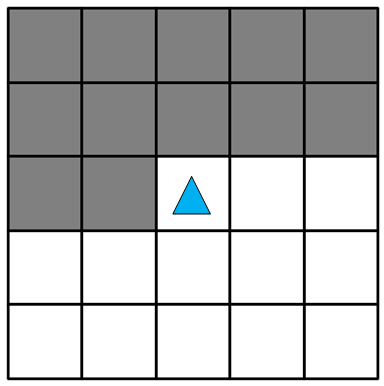}
\scriptsize{(c)}
\end{minipage}
\begin{minipage}{0.24\linewidth}
\center
\includegraphics[width=1.0\linewidth]{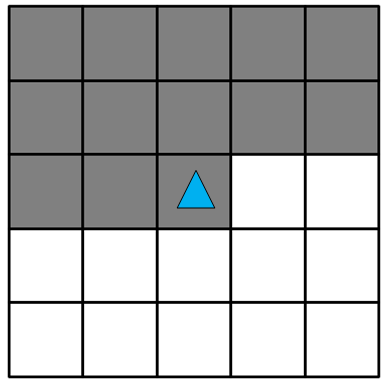}
\scriptsize{(d)}
\end{minipage}
\end{minipage}\par\medskip
   \caption{Mask planes with respect to $w_t$ for trimmed convolution kernels with the size of $5\times5$. The gray value denotes 1 and the white value denotes 0. The blue triangle represents the position of the codes to be encoded with respect to the mask.
    (a) $k<t$, (b) $k>t$, (c) $k = t$ for the input layer, (d) $k = t$ for the hidden layers.
}\label{mask}
\end{figure}

We then present our trimmed convolution by taking the properties of non-encoded symbols into account.
From Property (ii), we can keep voxel values unchanged, and introduce zeros in convolutional kernel $\mathbf{w}^0$ to exclude the effect of non-encoded symbols in convolution.
From Property (i), the positions of non-encoded symbols are fixed and pre-defined w.r.t. $\mathbf{w}^0$, thereby allowing us to employ a mask $\mathbf{\hat{m}}$ of $\{0, 1\}$ for correctly setting zeros.
Here, $\hat{m}_{k,i,j}$ is defined as $1$ if ${c}_{k,p+i,q+j}$ is encoded before {${c}_{k,p,q}$}, and 0 otherwise.
Trimmed convolution is then defined as,
\begin{equation}
\label{eqn:trimmed_convolution}
\mathbf{C}^{1} = \mathbf{C}*(\mathbf{\hat{m}} \circ \mathbf{w}^0),
\end{equation}
where $\circ$ denotes the element-wise product.
With trimmed convolution, we can safely exclude the effect of non-encoded symbols in context modeling while maintaining the efficiency of FCN for predicting probabilities of all voxels.

In the following, we first use single convolution kernel as an example to explain the settings of $\mathbf{\hat{m}}$,
%
%
which are different for the input layer and the hidden layers.
For the input layer, when predicting the probability of ${c}_{r,p,q}$, both ${c}_{r,p,q}$ and the symbols encoded after ${c}_{r,p,q}$ should be masked out in trimmed convolution.
Following the definition of context, we define the mask $\mathbf{\hat{m}}^0$ for the input layer as,
{
\begin{equation}
\hat{m}^0_{tkij}=\begin{cases}
1,  \mbox{if} \left\{k \!<\! t\right\} \!\vee\! \left\{k \!=\! t, i \!<\! 0\right\} \!\vee\! \left\{k \!=\! t, i \!=\! 0, j \!<\! 0\right\} \; , \\
0, \mbox{otherwise} \; .
\end{cases}
\end{equation}}
When it comes to the hidden layer $\mathbf{C}^{d}$ ($d \geq 1$), we note that the feature ${c}^{d}_{r,p,q}$ only conveys the context information of $\mathbf{c}_{r,p,q}$ and should not be excluded in the successive context modeling.
Therefore, we modify the definition of the mask $\mathbf{\hat{m}}^d$ ($d \geq 1$) for hidden layer as,
{
\begin{equation}
\hat{m}^d_{tkij}=\begin{cases}
1,  \mbox{if} \left\{k \!<\! t\right\} \!\vee\! \left\{k \!=\! t, i \!<\! 0\right\} \!\vee\! \left\{k \!=\! t, i \!=\! 0, j \!\leq\! 0\right\} , \\
0, \mbox{otherwise}.
\end{cases}
\end{equation}}

Using the convolution kernel $\mathbf{w}$ with the size of $5 \times 5 \times n$ as an example, Fig.~\ref{mask} illustrates the representative mask planes w.r.t. the kernel planes $\mathbf{w}_{t,k,\cdot,\cdot}$.
As shown in Fig.~\ref{mask}(a) (Fig.~\ref{mask}(b)), when $k < t$ ($k > t$), the $k$-th mask plane is a matrix of 1s (0s) for both the input and hidden layers.
When $k = t$, the center position should be masked out in the mask plane for the input layer (see Fig.~\ref{mask}(c)), but can be retained for the hidden layers (see Fig.~\ref{mask}(d)).
%

{\bf Multi-group trimmed convolution.} The trimmed convolution in Eqn. (\ref{eqn:trimmed_convolution}) only uses one group of convolution kernels in each layer, which is still limited in complicated probability prediction.
Thus, we extend the trimmed convolution to the multi-group form. 
Suppose there are $g_{in}$ groups of feature maps $\mathcal{C}^d = \{\mathbf{C}^{d,0}, ...,  \mathbf{C}^{d,g_{in}-1}\}$ in the $d$-th layer and $g_{out}$ groups of feature maps $\mathcal{C}^{d+1} = \{\mathbf{C}^{d+1,0}, ...,  \mathbf{C}^{d+1,g_{out}-1}\}$ in the $(d+1)$-th layer.
Each group of feature map has the same size with the input cuboid $\mathcal{C}$.
The group trimmed convolution is defined as,
\begin{equation}
\label{eqn:group_trim}
\mathbf{C}^{d+1, g^{\prime}} = \sum_{g = 0}^{g_{in}-1} \mathbf{C}^{d,g}*(\mathbf{\hat{m}}^d \circ \mathbf{w}^{d,g,g^{\prime}}),
\end{equation}
where $\mathbf{C}^{d,g}$ denotes the $g$-th group of feature map in $\mathcal{C}^d$, $\mathbf{C}^{d+1, g^{\prime}}$ denotes the $g^{\prime}$-th group of feature map in $\mathcal{C}^{d+1}$.
$\mathbf{\hat{m}}^d$ is the mask for the $d$-th layer, and $\mathbf{w}^{d,g,g^{\prime}}$ is the convolution kernel to connect $\mathbf{C}^{d,g}$ and $\mathbf{C}^{d+1, g^{\prime}}$.

\vspace{-.1in}
\subsection{TCAE and Learning Objective}
Our proposed TCAE is constructed by stacking several $5\times5$  trimmed convolution layers to enlarge the context and increase the nonlinearity of the model.
Given all the model parameters $\mathcal{W} = \{ \mathbf{w}^{d,g,g^{\prime}} \}$ , the output of TCAE can be written as $F(\mathbf{C}; \mathcal{W}) = \{ \left(F(\mathbf{C}; \mathcal{W})\right)^b_{r,p,q} | b=0,\ldots,m-1 \}$.
Here, $\left(F(\mathbf{C}; \mathcal{W})\right)^b_{r,p,q}$ denotes the predicted probability of ${c}_{r,p,q} = b$, and $m$ is the number of quantization levels of the input code map.
{Using $\mathbf{o}^\prime$ as an example,}
we adopt the code length after arithmetic encoding as the learning objective,
\begin{equation}
\label{eqn:objective}
\ell(\mathcal{W}; \mathbf{C}) = \sum_{{r,p,q}}\sum_{b=0}^{m-1} - {m}_{rpq}s(c_{r,p,q},b) \log_2 \left(F(\mathbf{C}; \mathcal{W})\right)^b_{r,p,q}
\end{equation}
where $s({c}_{r,p,q},b)=1$ when ${c}_{r,p,q} = b$, and $s({c}_{r,p,q},b)=0$ otherwise. ${m}_{rpq}$ is an importance mask to exclude those codes with the ${m}_{rpq}=0$ according to Eqn. (\ref{eq:mask}) during training.

\vspace{-.1in}
\subsection{Inclined TCAE}
					
The above TCAE can only accelerate the encoding process.
In the decoding stage, the codes should still be decoded in a sequential order and cannot be speeded up with GPU and parallel computation.
To alleviate this issue, we present an inclined TCAE model by introducing another kind of coding schedule and context.
Concretely, we divide the 3D cuboid $\mathbf{C}$ into $n+h+w-2$ inclined planes, where the $t$-th inclined plane is defined as $CB_t(\mathbf{C}) = \{c_{k,i,j}|k+i+j=t\}$ ($t = 0, 1, \ldots, n+h+w-3$).
In inclined TCAE, the context of the current symbol $c_{r,p,q}$ is then defined as $\mbox{CTX}_b(c_{r,p,q}) = \{ CB_0(\mathbf{C}), CB_1(\mathbf{C}), \ldots, CB_{r+p+q-1}(\mathbf{C}) \}$.
In terms of coding schedule, we simply begin with $CB_0(\mathbf{C})$ and then gradually encode $CB_t(\mathbf{C})$ after $CB_{t-1}(\mathbf{C})$.

To accelerate the encoding process, we define the mask for the input and hidden layers as follows,
\begin{equation}
\hat{m}^0_{tkij}=\begin{cases}
1,  \mbox{if}\; i+k+j <0 \; , \\
0, \mbox{otherwise} \; .
\end{cases}
\end{equation}
\begin{equation}
\hat{m}^d_{tkij}=\begin{cases}
1,  \mbox{if}\; i+k+j \leq 0 , \\
0, \mbox{otherwise} \;.
\end{cases}
\end{equation}
Benefited from the new coding schedule and context, all $c_{r,p,q}$s in $CB_t(\mathbf{C})$ share the same context and can be decoded in parallel, which can then be utilized to speed up decoding process with GPU parallel computation.
							
\vspace{-.1in}
\subsection{Implementation and Learning}\label{sec:implementation}
Two inclined TCAE models are deployed for context modeling of {the code $\mathbf{o}^\prime$} and quantized importance map $QI(\mathbf{p})$, respectively.

As for network structure, our inclined TCAE is comprised of 2 trimmed convolution layers, 3 residual blocks with each consisting of 2 trimmed convolution layers, and a final trimmed convolution layer {followed by a softmax function}.
For context modeling of {$\mathbf{o}^\prime$}, $8$ groups are used for the first 8 trimmed convolution layers and 9 groups are used for the last layer.
For context modeling of $QI(\mathbf{p})$, $32$ groups are used for the first 8 layers and $16$ groups for last layer.

To train inclined TCAE, we adopt the ADAM solver~\cite{kingma2014adam}.
The model is trained with the learning rate of $3 \times 10^{-4}$, $1 \times 10^{-4}$, $3.33 \times 10^{-5}$ and $1.11 \times 10^{-5}$.	
The smaller learning rate is adopted until the objective with the larger learning rate keeps non-decreasing in five successive epochs.

\vspace{-.1in}
\section{Experiments}\label{sec:experiment}

In this section, we compare our full CWIC method with both existing image encoding standards and state-of-the-art deep image compression models.
A number of ablation studies are also given to assess the effect of importance map, channel-wise multi-valued quantization, and {inclined} TCAE.
The pre-trained models will be available at \url{https://github.com/limuhit/CWIC}.

\vspace{-.1in}
\subsection{Experimental Setup}\label{sec:expriment_setup}
By setting different $r$ values, we train 7 models based on MS-SSIM distortion loss and and 7 models based on MSE loss.
%
%
All the models are trained with 10,000 high quality images crawled from the photo sharing website \emph{Flickr}.
Each image is downsampled to its 1/3 size for removing the possible compression artifacts caused by JPEG compression and save the downsampled image with the lossless \emph{PNG} format.
Finally, 500,000 patches with the size of $384\times384$ are randomly cropped from the 10,000 images for training.
For performance evaluation, we adopt two datasets, i.e., Kodak PhotoCD and Tecnick.
The compression rate of our model is evaluated by bits per pixel (bpp), which is calculated as the total amount of bits used to code the image divided by the number of pixels.
The image distortion is evaluated by the Multi-Scale Structural Similarity (MS-SSIM) and the Peak Signal-to-Noise Ratio (PSNR).
%


\vspace{-.1in}
\subsection{Quantitative Evaluation}

Using MS-SSIM and PSNR as performance metrics, we evaluate the rate-distortion performance of our CWIC, existing image encoding standards, and state-of-the-art deep image compression models.
In terms of image encoding standards, we consider JPEG~\cite{wallace1992jpeg}, JPEG 2000~\cite{skodras2001jpeg}, and BPG~\cite{bellard2016bpg}.
Among different variants of JPEG, the optimized JPEG with 4:2:0 chroma subsampling is adopted\footnote{http://libjpeg.sourceforge.net/}.
JPEG 2000 is based on the optimized implementation in MATLAB 2015, and the implemented BPG is based on the 4:2:0 chroma format\footnote{https://bellard.org/bpg/}.
In terms of deep image compression models, their source codes generally are not available.
Following the strategy adopted in~\cite{mentzer2018conditional1}, we carefully digitalize and collect rate-distortion curves from the {related literatures~\cite{rippel2017real,theis2017lossy,agustsson2017soft}} in our comparative experiments.

\begin{figure}
\center
\begin{minipage}{0.485\linewidth}
\center
\includegraphics[width=1.0\linewidth]{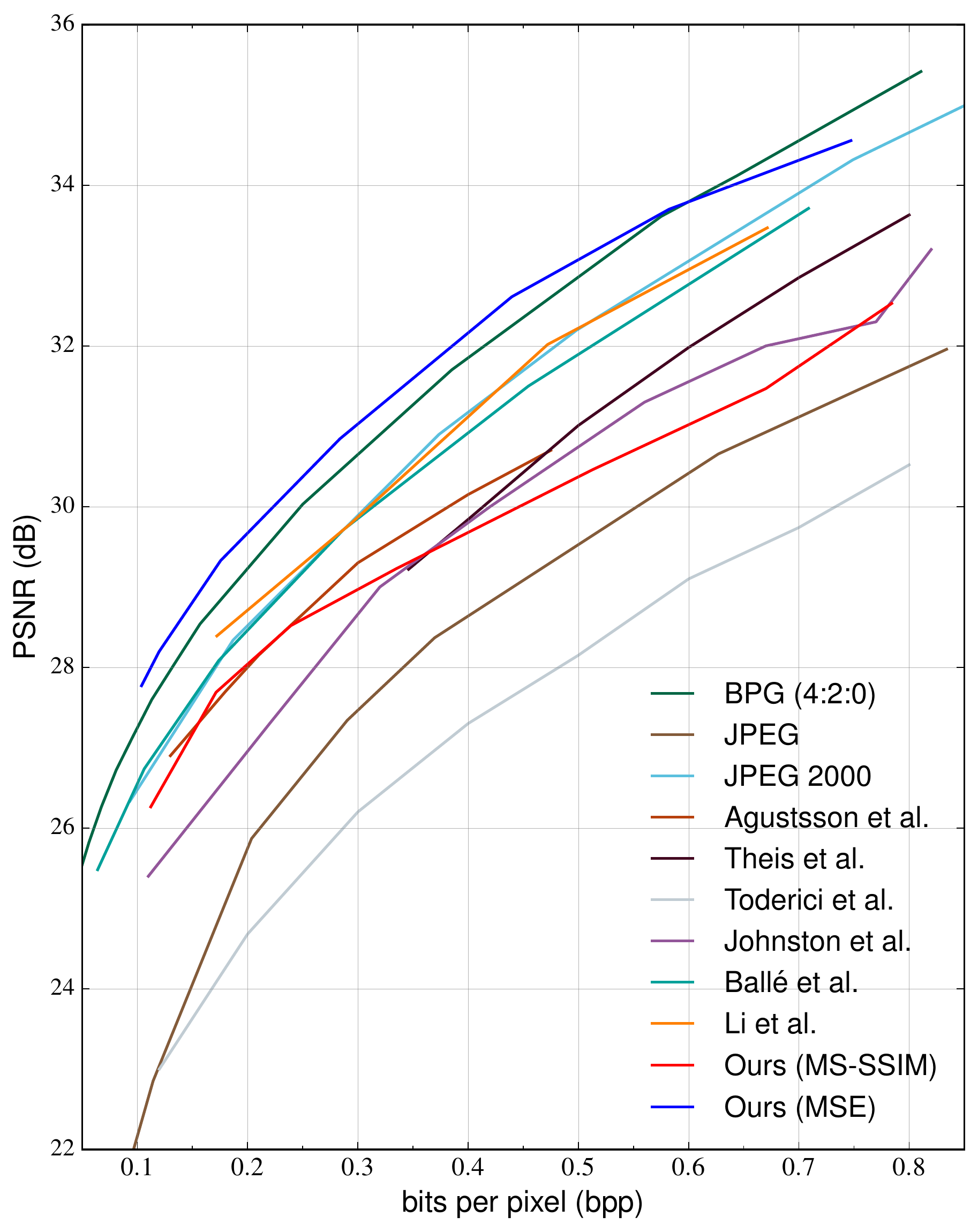}
\scriptsize{(a) PSNR}
\end{minipage}
\begin{minipage}{0.495\linewidth}
\center
\includegraphics[width=1.0\linewidth]{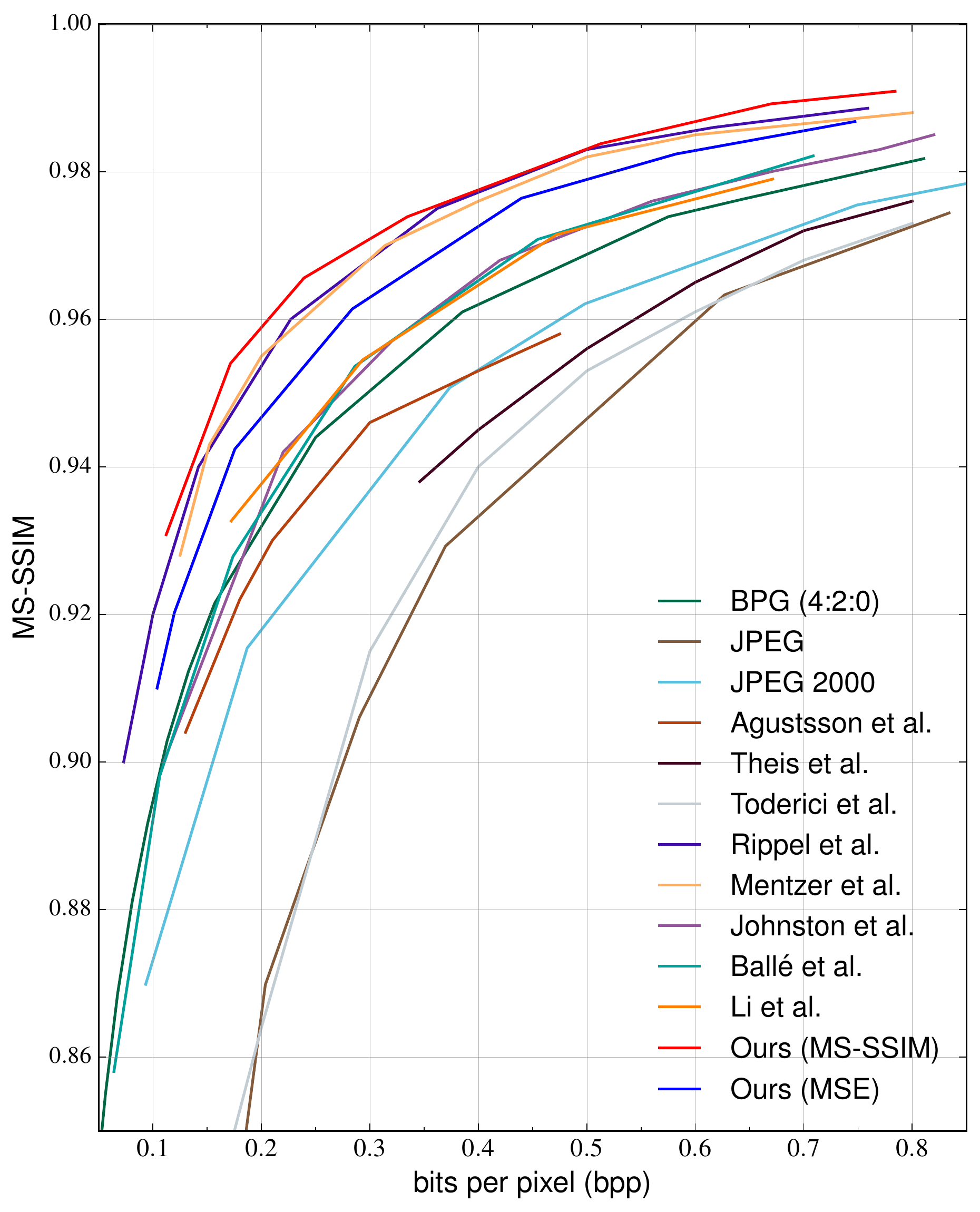}
\scriptsize{(b) MS-SSIM}
\end{minipage}\par\medskip
\caption{{Rate-distortion curves of different compression algorithms w.r.t. (a) PSNR and (b) MS-SSIM on the Kodak PhotoCD image dataset.}}\label{fig:psnr_kodak}
\end{figure}

\begin{figure}
\center
\begin{minipage}{0.485\linewidth}
\center
\includegraphics[width=1.0\linewidth]{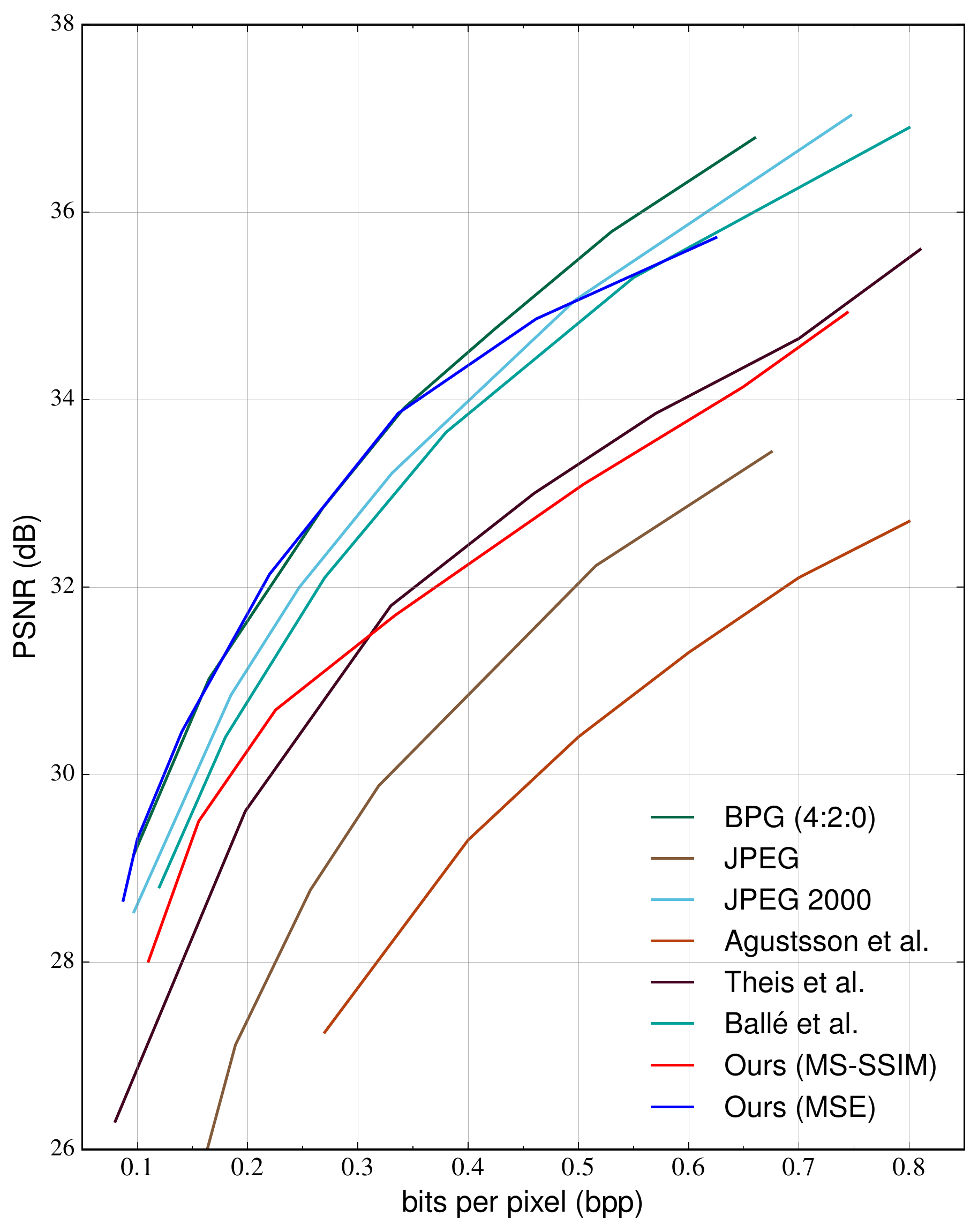}
\scriptsize{(a) PSNR}
\end{minipage}
\begin{minipage}{0.495\linewidth}
\center
\includegraphics[width=1.0\linewidth]{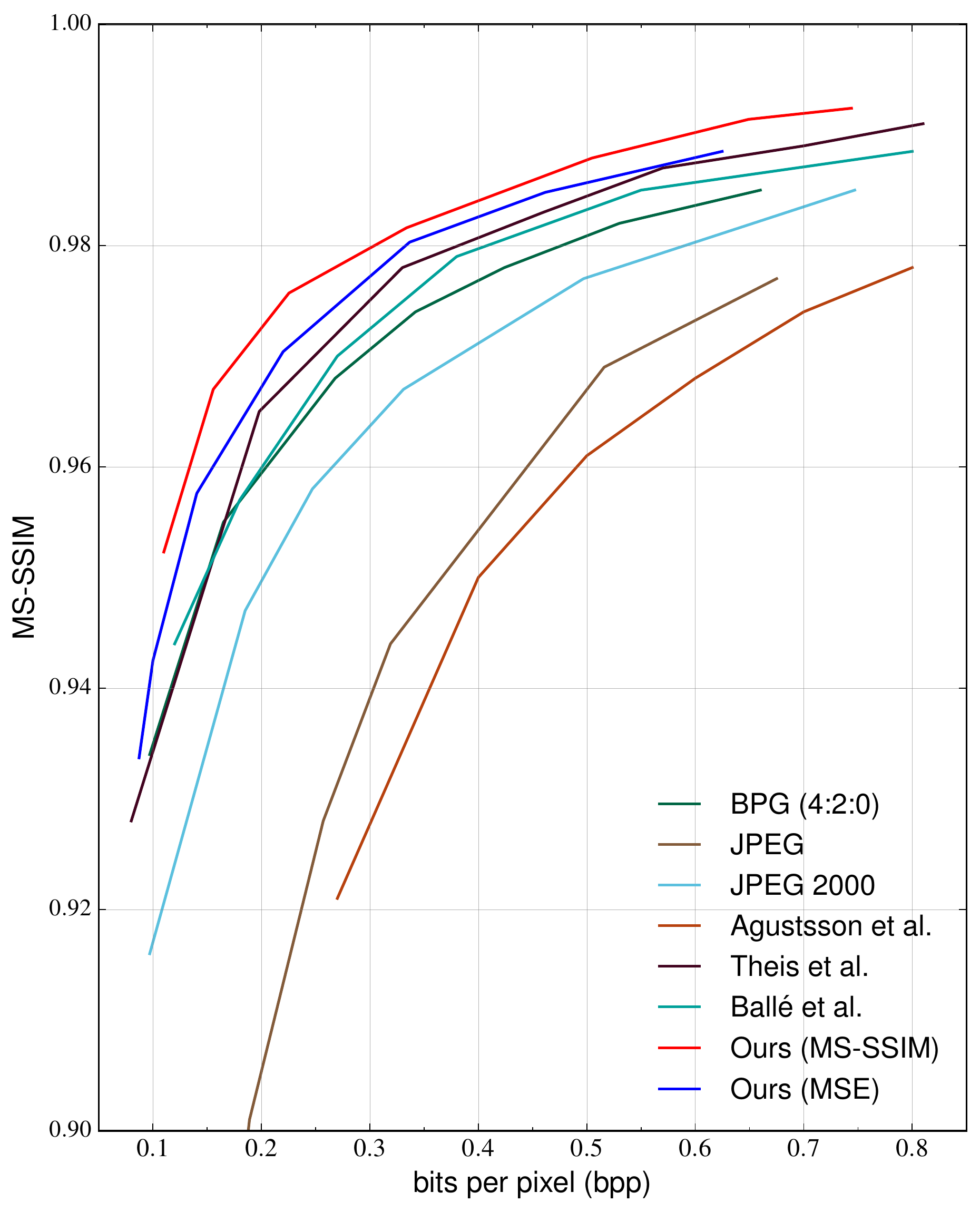}
\scriptsize{(b) MS-SSIM}
\end{minipage}\par\medskip
\caption{{Rate-distortion curves of different compression algorithms w.r.t. (a) PSNR and (b) MS-SSIM on the Tecnick dataset.
}}\label{fig:psnr_tec}
\end{figure}

Fig.~\ref{fig:psnr_kodak} shows the rate-distortion curves of competing methods on the Kodak datasets.
In terms of MS-SSIM, we compare our method with JPEG, JPEG 2000, BPG, Ball{\'e} et al.~\cite{balle2016end}, Rippel et al.~\cite{rippel2017real}, Theis et al.~\cite{theis2017lossy}, Johnston et al.~\cite{johnston2017improved}, Toderici et al.~\cite{toderici2015variable}, Agustsson et al.~\cite{agustsson2017soft} and Mentzer et al.~\cite{mentzer2018conditional1}.
In terms of PSNR, we exclude Rippel et al. and Mentzer et al. due to that the PSNR result is not reported in their paper~\cite{rippel2017real,mentzer2018conditional1}.
From Fig.~\ref{fig:psnr_kodak}(a), Ours(MS-SSIM) performs on par with Rippel et al.~\cite{rippel2017real} and outperforms the other methods by MS-SSIM.
It is worth noting that, Ours(MSE) also exhibits competitive MS-SSIM performance, which is only inferior to Rippel et al.~\cite{rippel2017real} {and Mentzer et al.~\cite{mentzer2018conditional1}} among the competing methods, probably being explained by that the use of importance map benefits the reconstruction of salient and structural information at lower bpp.
From Fig.~\ref{fig:psnr_kodak}(b), Ours(MSE) is comparable with BPG, and is much better {than} the other methods.
Furthermore, we give the rate-distortion curves of competing methods on Tecnick~\cite{asuni2014testimages,asuni2013testimages} in Fig.~\ref{fig:psnr_tec}, and get the similar observations with the Kodak dataset.
It is noted that the results of Rippel et al.~\cite{rippel2017real}, Theis  et al.~\cite{theis2017lossy}, Agustsson et al.~\cite{agustsson2017soft} and Mentzer et al.~\cite{mentzer2018conditional1} are unavailable on Tecnick.


\begin{figure*}[!tbp]
\centering
\hfill\vline\hfill
\begin{minipage}{1.0\textwidth}
\begin{minipage}{0.195\textwidth}\center{\scriptsize {Original}}\end{minipage}
\begin{minipage}{0.195\textwidth}\center{\scriptsize {JPEG 2000}}\end{minipage}
\begin{minipage}{0.195\textwidth}\center{\scriptsize {Ball{\'e} et al.}}\end{minipage}
\begin{minipage}{0.195\textwidth}\center{\scriptsize {BPG}}\end{minipage}
\begin{minipage}{0.195\textwidth}\center{\scriptsize {Ours}}\end{minipage}
\end{minipage}\par\medskip

\begin{minipage}{1.0\textwidth}
\begin{minipage}{0.195\textwidth}
\includegraphics[width=1.0\textwidth]{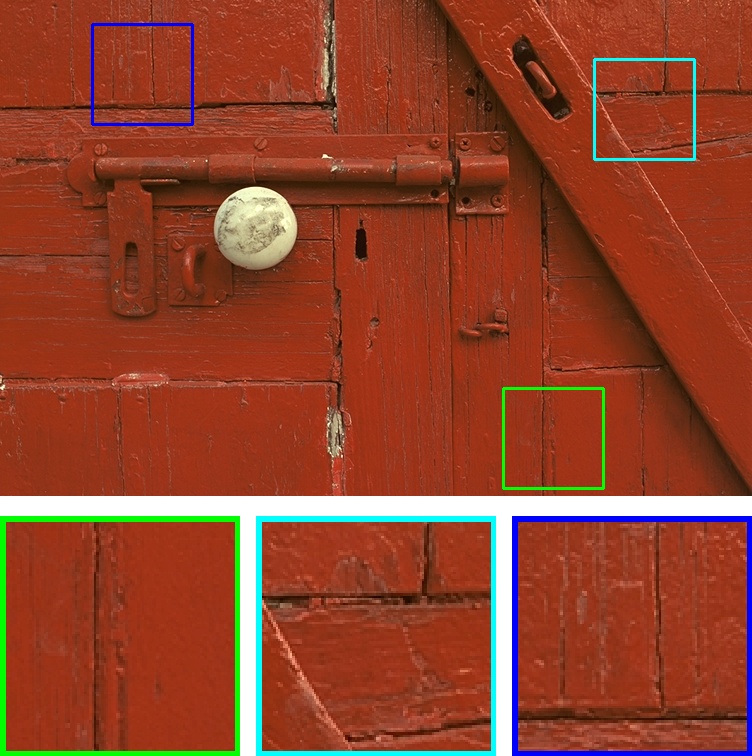}
\end{minipage}
\begin{minipage}{0.195\textwidth}
\includegraphics[width=1.0\textwidth]{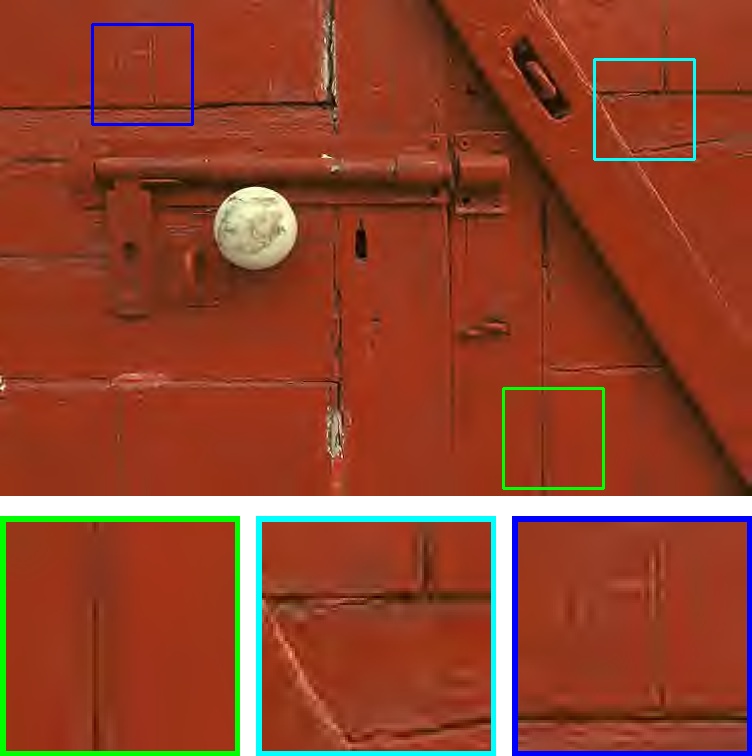}
\end{minipage}
\begin{minipage}{0.195\textwidth}
\includegraphics[width=1.0\textwidth]{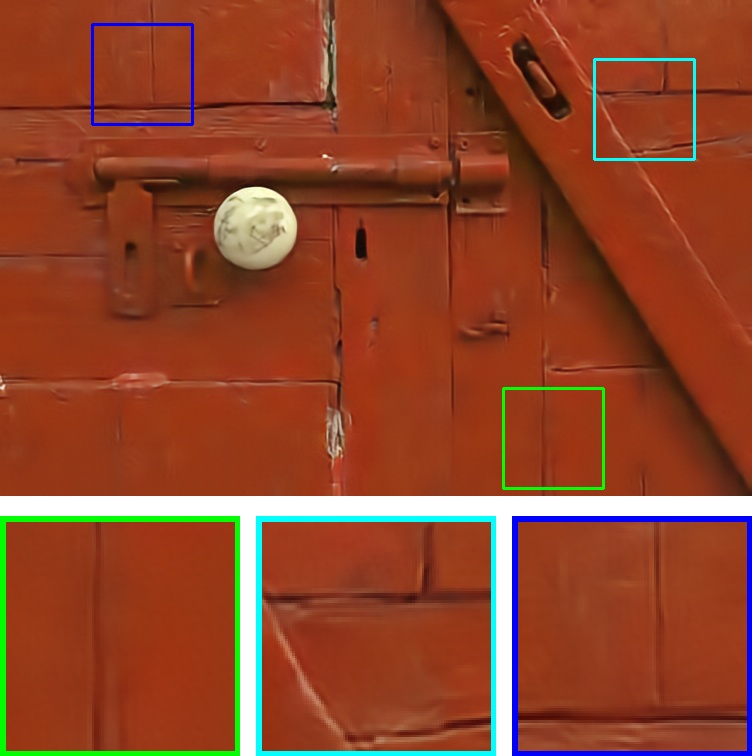}
\end{minipage}
\begin{minipage}{0.195\textwidth}
\includegraphics[width=1.0\textwidth]{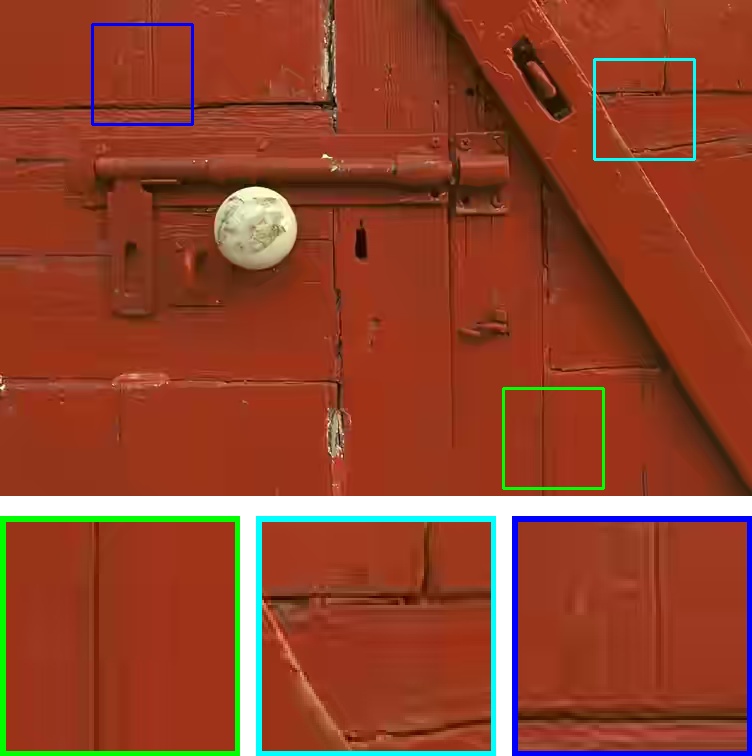}
\end{minipage}
\begin{minipage}{0.195\textwidth}
\includegraphics[width=1.0\textwidth]{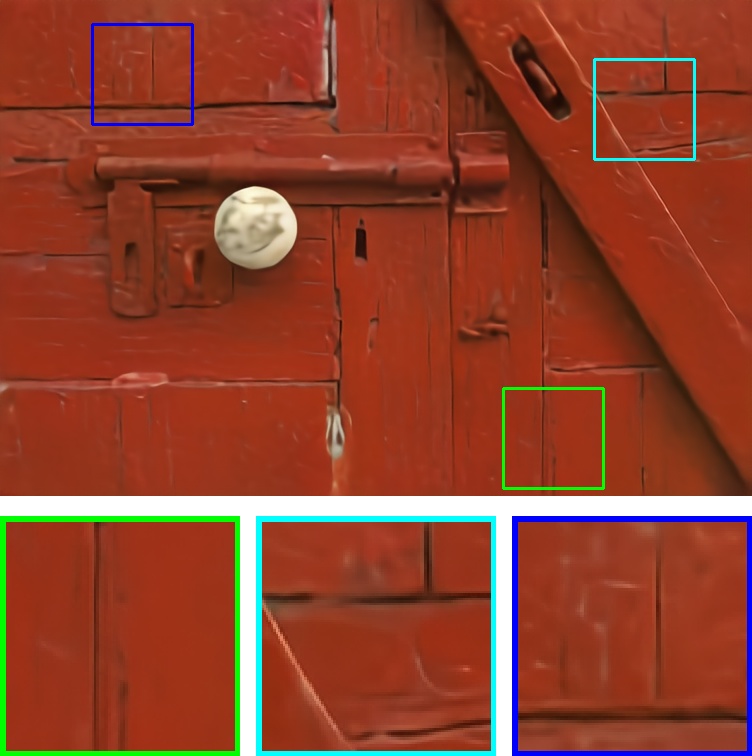}
\end{minipage}
\end{minipage}\par\medskip

\begin{minipage}{1.0\textwidth}
\begin{minipage}{0.195\textwidth}\center{\scriptsize {bpp / PSNR / MS-SSIM}}\end{minipage}
\begin{minipage}{0.195\textwidth}\center{\scriptsize{0.094 / 29.69 / 0.882}}\end{minipage}
\begin{minipage}{0.195\textwidth}\center{\scriptsize{0.107 / 29.83 / 0.899}}\end{minipage}
\begin{minipage}{0.195\textwidth}\center{\scriptsize{0.093 / 30.62 / 0.900}}\end{minipage}
\begin{minipage}{0.195\textwidth}\center{\scriptsize{0.091 / 29.39 / 0.924}}\end{minipage}
\end{minipage}\par\medskip

\begin{minipage}{1.0\textwidth}
\begin{minipage}{0.195\textwidth}
\includegraphics[width=1.0\textwidth]{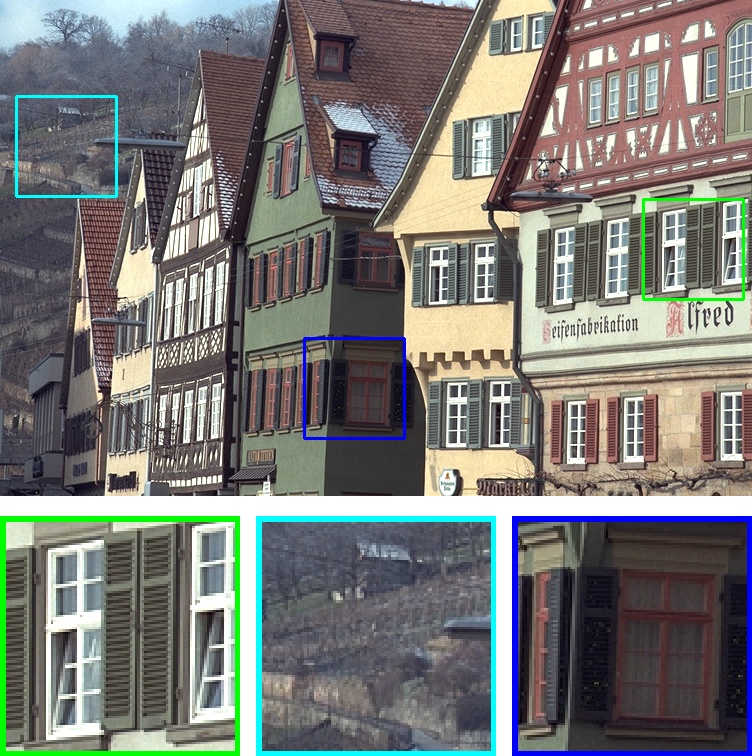}
\end{minipage}
\begin{minipage}{0.195\textwidth}
\includegraphics[width=1.0\textwidth]{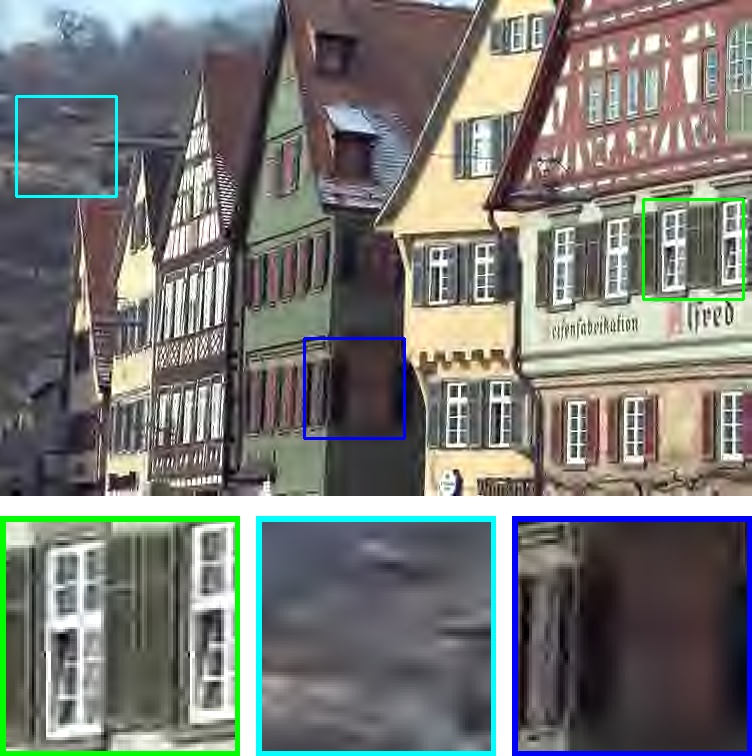}
\end{minipage}
\begin{minipage}{0.195\textwidth}
\includegraphics[width=1.0\textwidth]{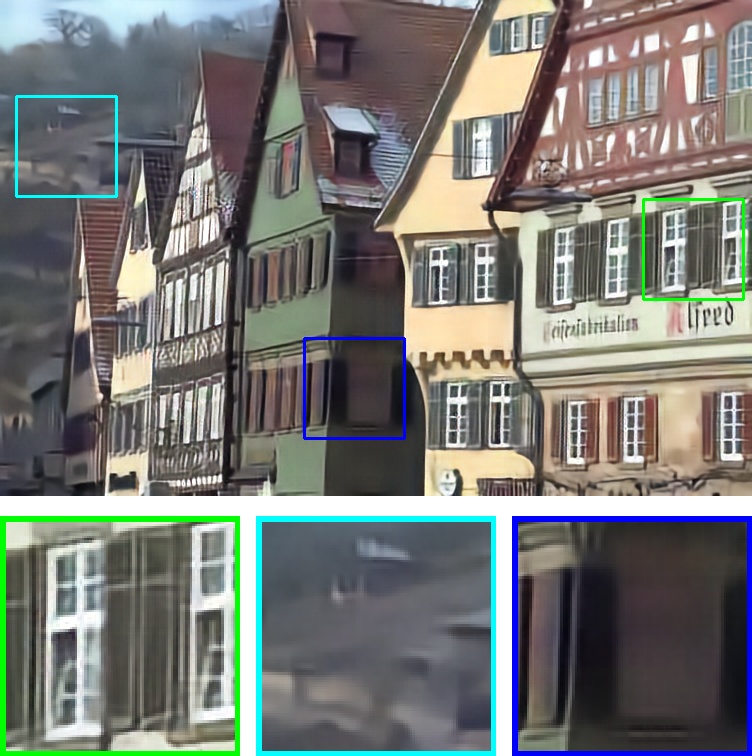}
\end{minipage}
\begin{minipage}{0.195\textwidth}
\includegraphics[width=1.0\textwidth]{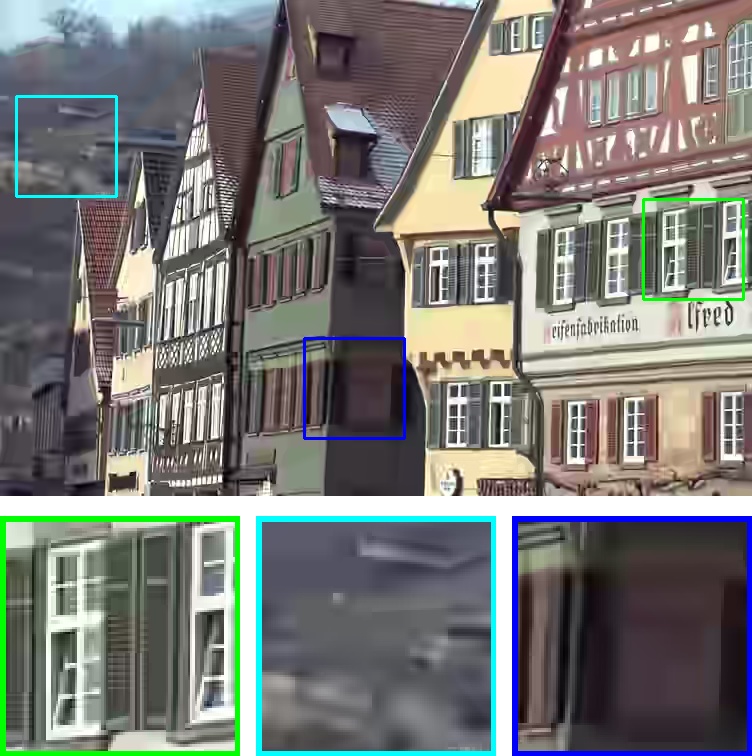}
\end{minipage}
\begin{minipage}{0.195\textwidth}
\includegraphics[width=1.0\textwidth]{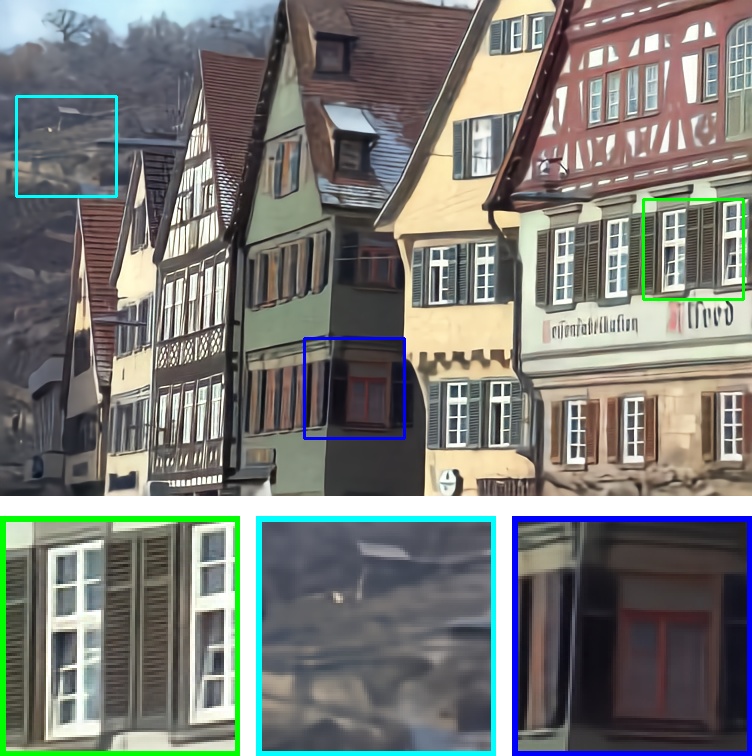}
\end{minipage}
\end{minipage}\par\medskip

\begin{minipage}{1.0\textwidth}
\begin{minipage}{0.195\textwidth}\center{\scriptsize {bpp / PSNR / MS-SSIM}}\end{minipage}
\begin{minipage}{0.195\textwidth}\center{\scriptsize{0.187 / 21.97 / 0.885}}\end{minipage}
\begin{minipage}{0.195\textwidth}\center{\scriptsize{0.203 / 22.44 / 0.919}}\end{minipage}
\begin{minipage}{0.195\textwidth}\center{\scriptsize{0.192 / 23.74 / 0.921}}\end{minipage}
\begin{minipage}{0.195\textwidth}\center{\scriptsize{0.180 / 24.21 / 0.955}}\end{minipage}
\end{minipage}\par\medskip

\begin{minipage}{1.0\textwidth}
\begin{minipage}{0.195\textwidth}
\includegraphics[width=1.0\textwidth]{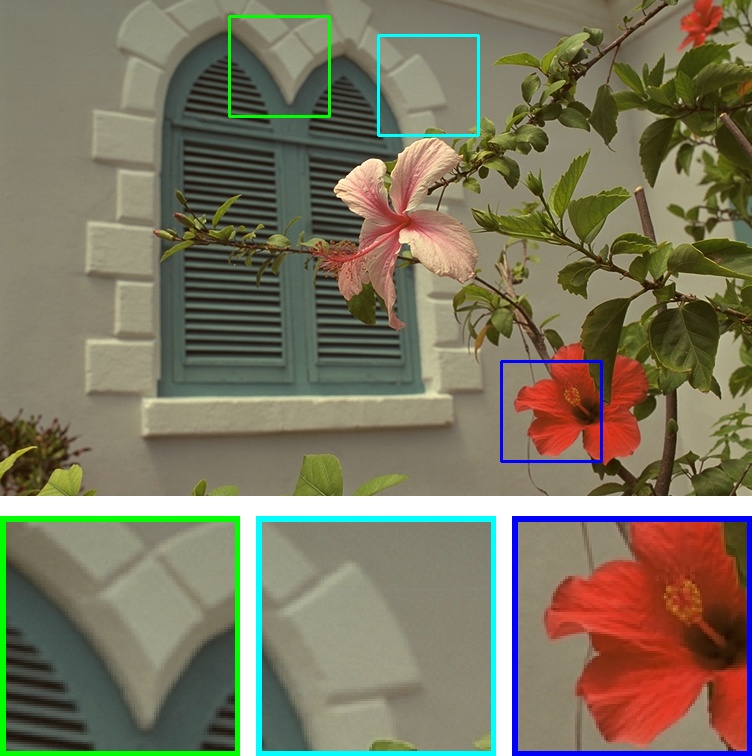}
\end{minipage}
\begin{minipage}{0.195\textwidth}
\includegraphics[width=1.0\textwidth]{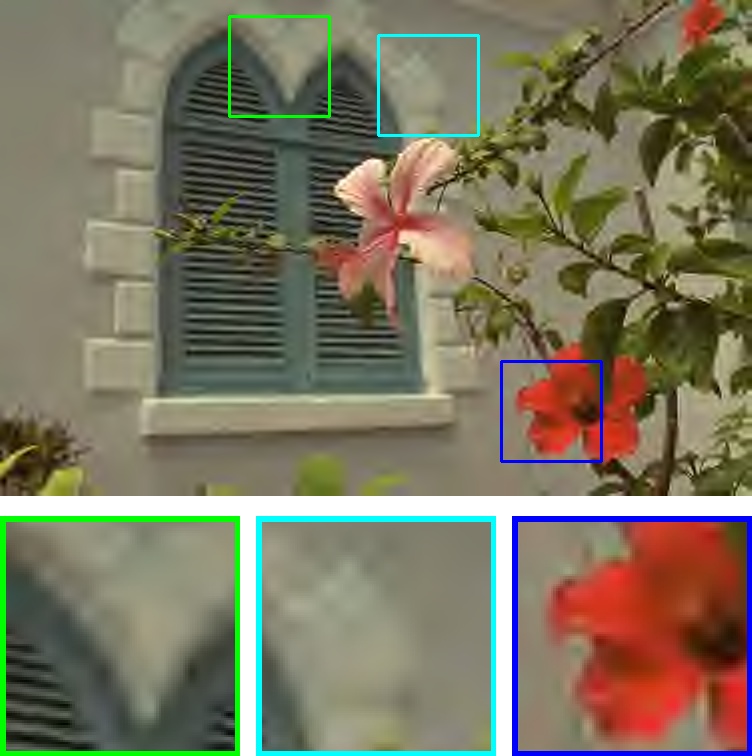}
\end{minipage}
\begin{minipage}{0.195\textwidth}
\includegraphics[width=1.0\textwidth]{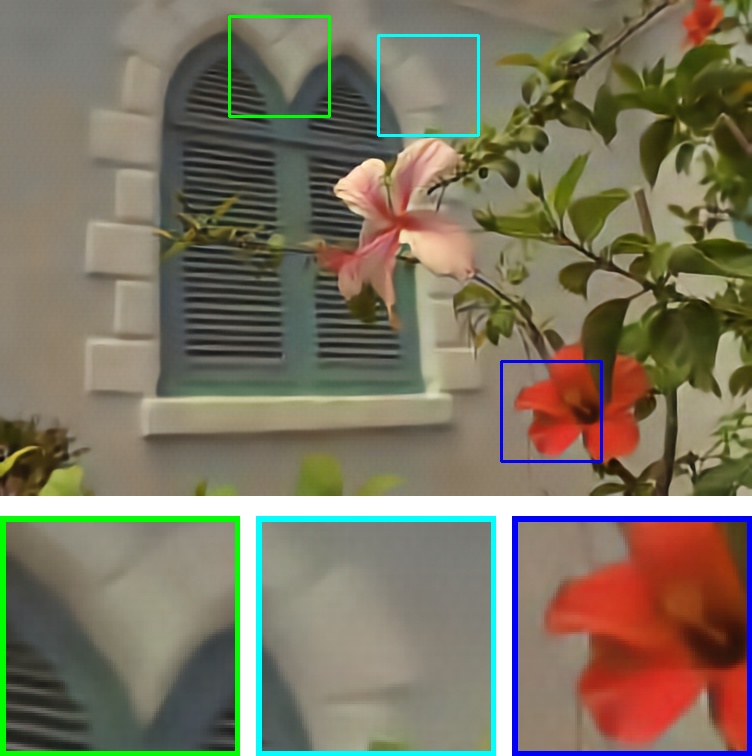}
\end{minipage}
\begin{minipage}{0.195\textwidth}
\includegraphics[width=1.0\textwidth]{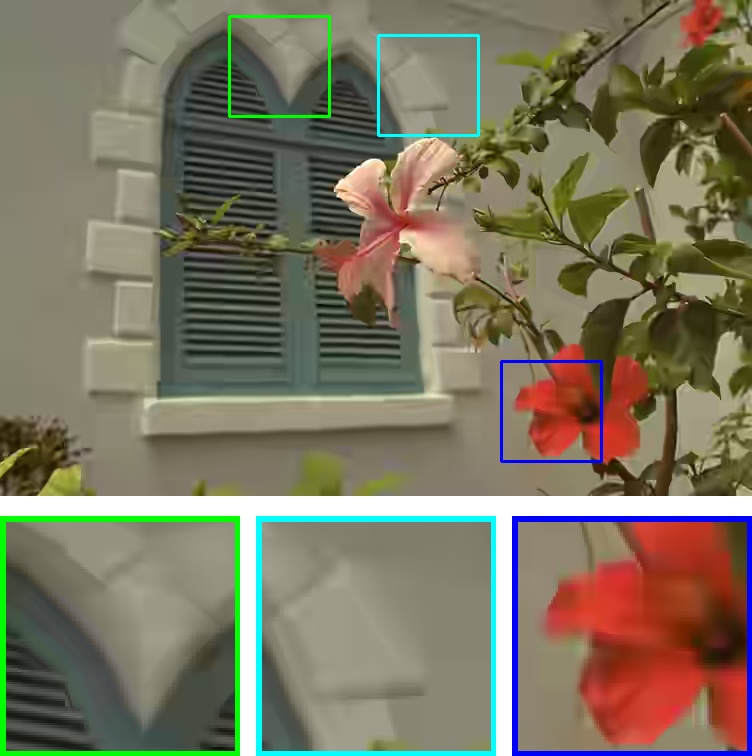}
\end{minipage}
\begin{minipage}{0.195\textwidth}
\includegraphics[width=1.0\textwidth]{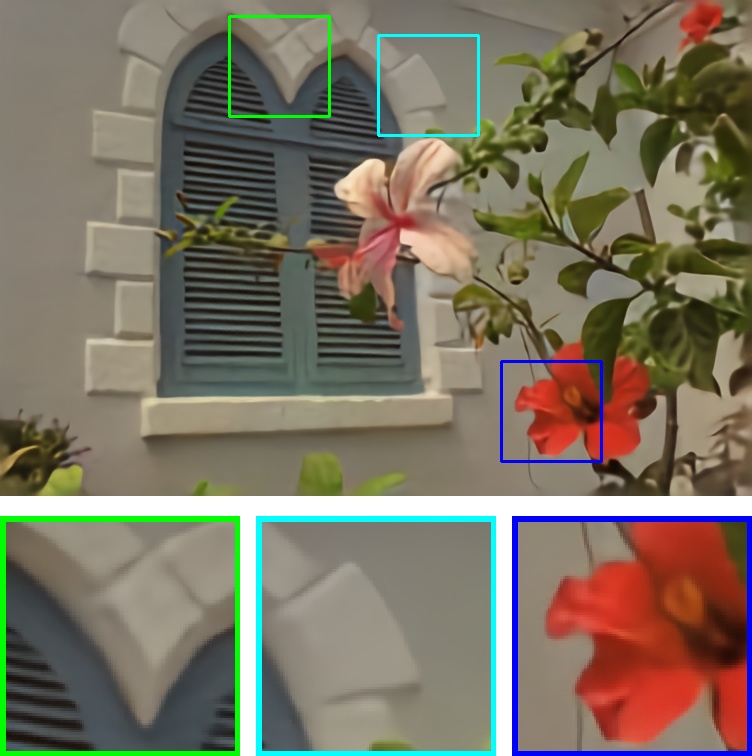}
\end{minipage}
\end{minipage}\par\medskip

\begin{minipage}{1.0\textwidth}
\begin{minipage}{0.195\textwidth}\center{\scriptsize {bpp / PSNR / MS-SSIM}}\end{minipage}
\begin{minipage}{0.195\textwidth}\center{\scriptsize{0.094 / 26.84 / 0.911}}\end{minipage}
\begin{minipage}{0.195\textwidth}\center{\scriptsize{0.113 / 28.34 / 0.942}}\end{minipage}
\begin{minipage}{0.195\textwidth}\center{\scriptsize{0.101 / 28.79 / 0.946}}\end{minipage}
\begin{minipage}{0.195\textwidth}\center{\scriptsize{0.092 / 27.38 / 0.957}}\end{minipage}
\end{minipage}\par\medskip

\begin{minipage}{1.0\textwidth}
\begin{minipage}{0.195\textwidth}
\includegraphics[width=1.0\textwidth]{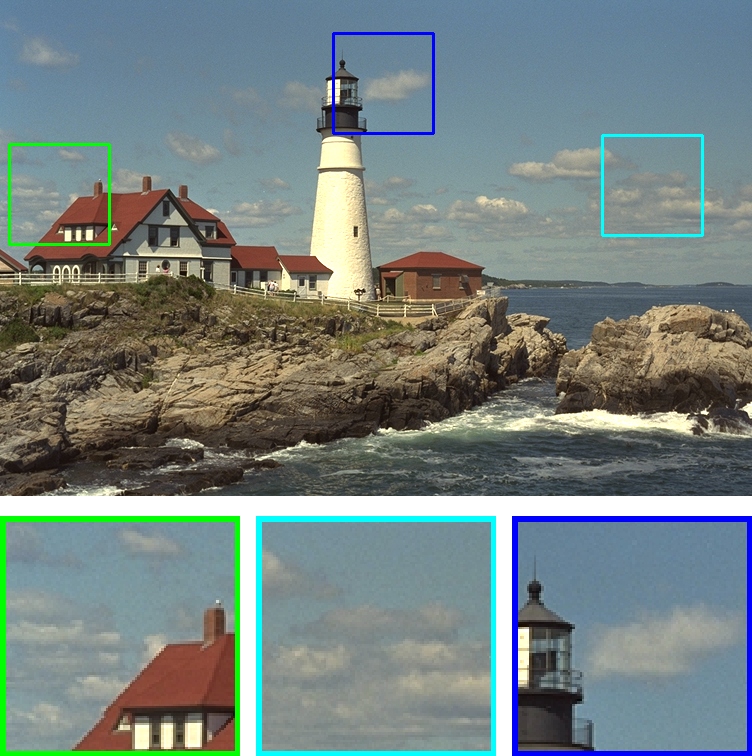}
\end{minipage}
\begin{minipage}{0.195\textwidth}
\includegraphics[width=1.0\textwidth]{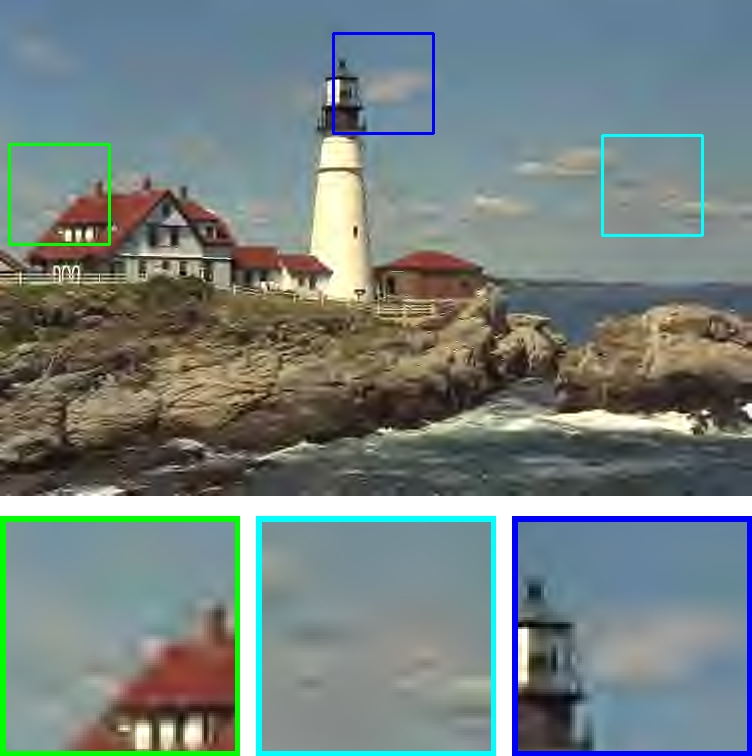}
\end{minipage}
\begin{minipage}{0.195\textwidth}
\includegraphics[width=1.0\textwidth]{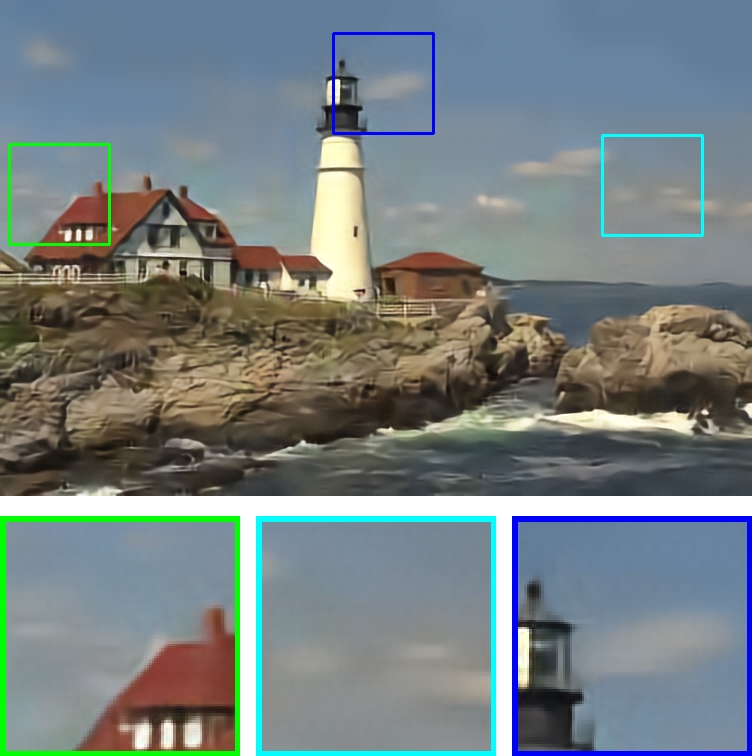}
\end{minipage}
\begin{minipage}{0.195\textwidth}
\includegraphics[width=1.0\textwidth]{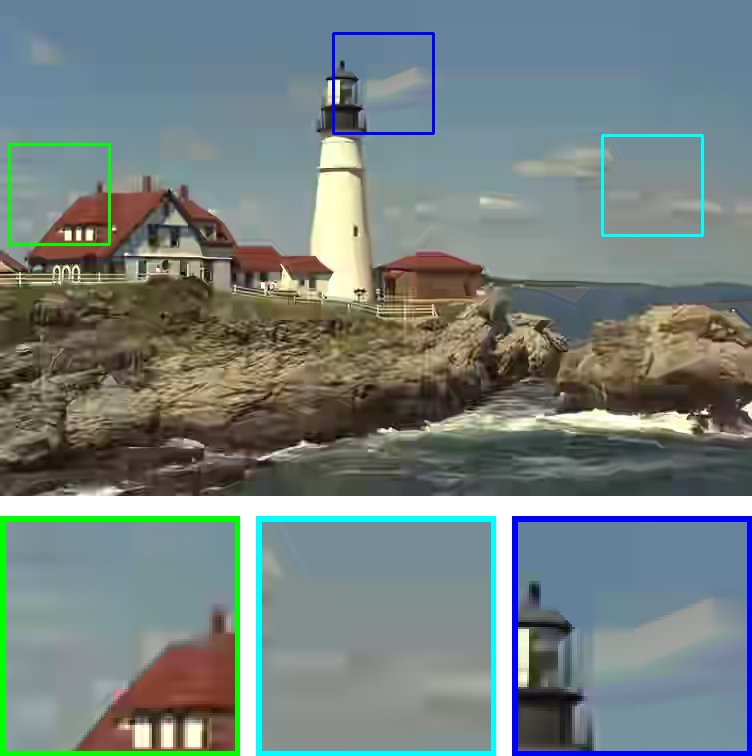}
\end{minipage}
\begin{minipage}{0.195\textwidth}
\includegraphics[width=1.0\textwidth]{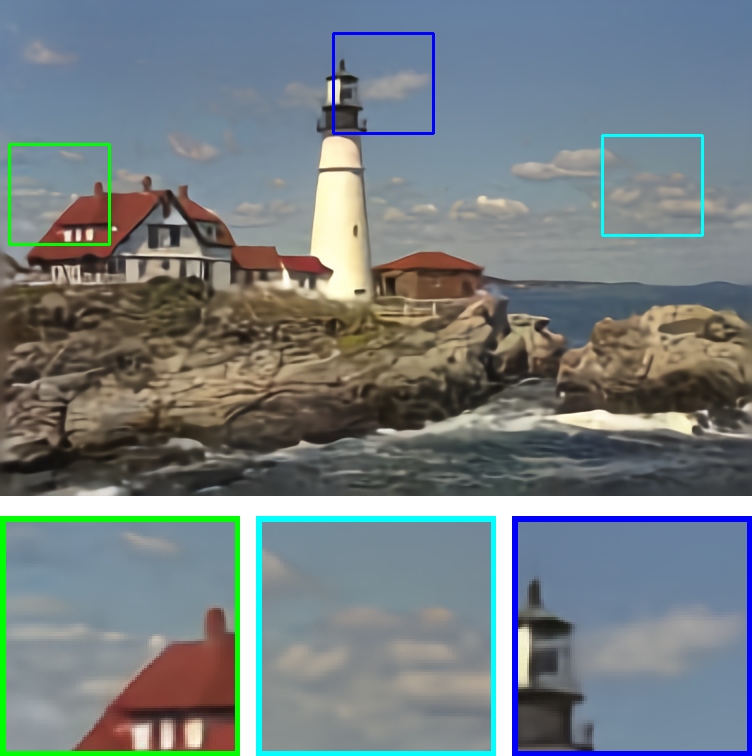}
\end{minipage}
\end{minipage}\par\medskip

\begin{minipage}{1.0\textwidth}
\begin{minipage}{0.195\textwidth}\center{\scriptsize {bpp / PSNR / MS-SSIM}}\end{minipage}
\begin{minipage}{0.195\textwidth}\center{\footnotesize{0.092 / 24.93 / 0.892}}\end{minipage}
\begin{minipage}{0.195\textwidth}\center{\scriptsize{0.106 / 25.65 / 0.918}}\end{minipage}
\begin{minipage}{0.195\textwidth}\center{\scriptsize{0.091 / 25.74 / 0.907}}\end{minipage}
\begin{minipage}{0.195\textwidth}\center{\scriptsize{0.092 / 24.43 / 0.929}}\end{minipage}
\end{minipage}\par\medskip

\begin{minipage}{1.0\textwidth}
\begin{minipage}{0.195\textwidth}
\includegraphics[width=1.0\textwidth]{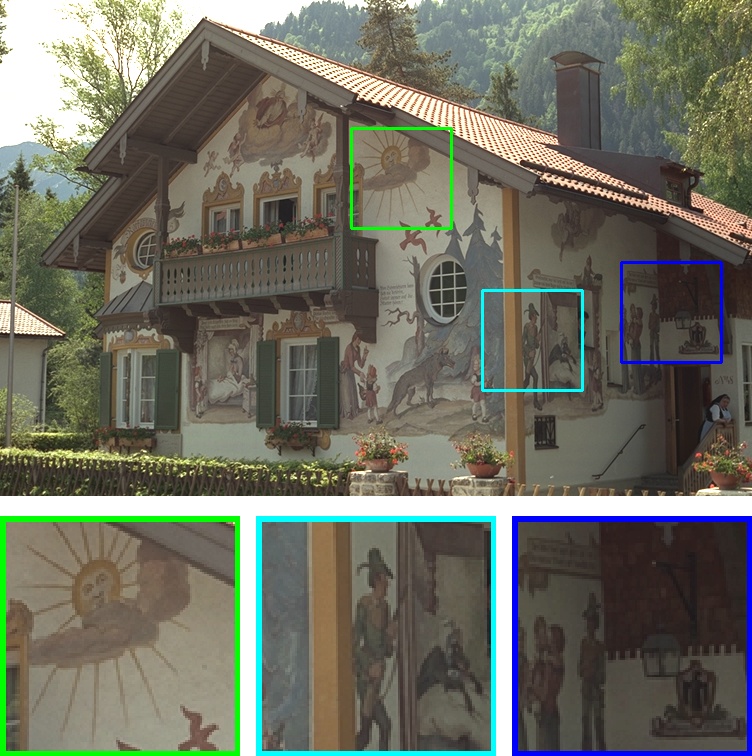}
\end{minipage}
\begin{minipage}{0.195\textwidth}
\includegraphics[width=1.0\textwidth]{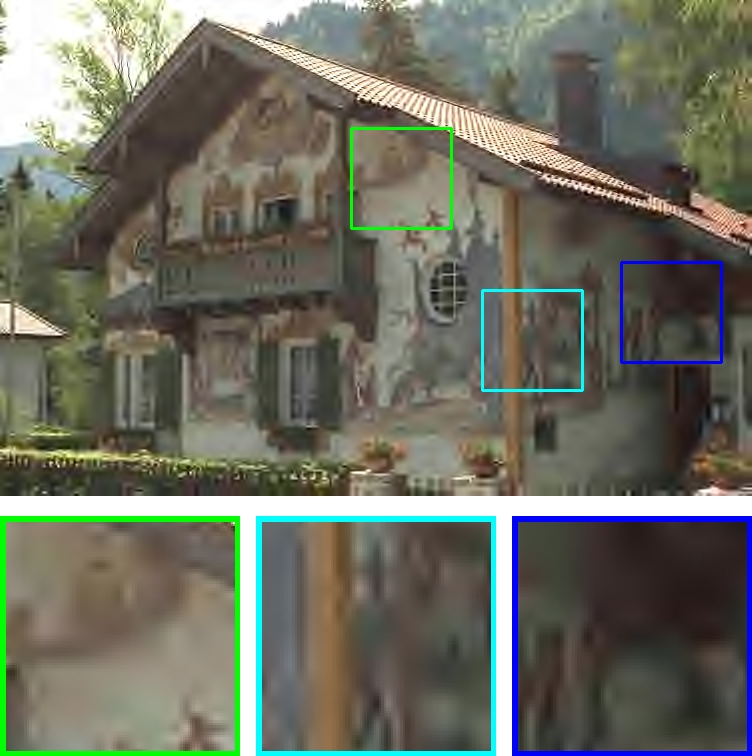}
\end{minipage}
\begin{minipage}{0.195\textwidth}
\includegraphics[width=1.0\textwidth]{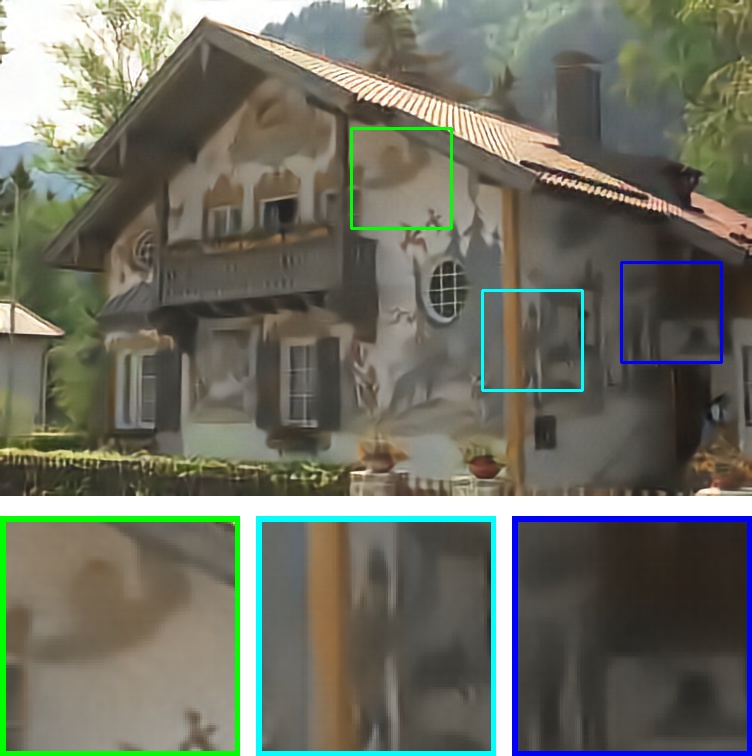}
\end{minipage}
\begin{minipage}{0.195\textwidth}
\includegraphics[width=1.0\textwidth]{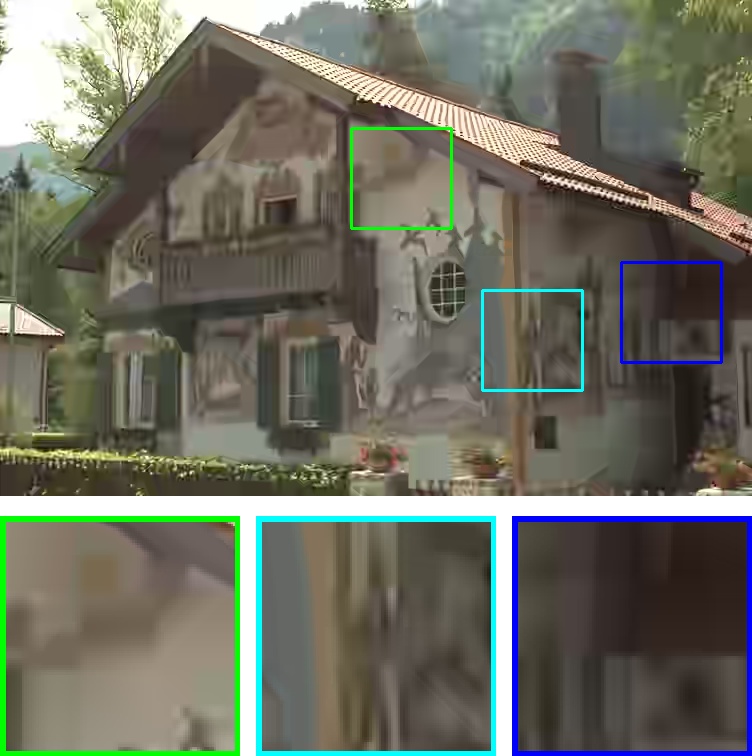}
\end{minipage}
\begin{minipage}{0.195\textwidth}
\includegraphics[width=1.0\textwidth]{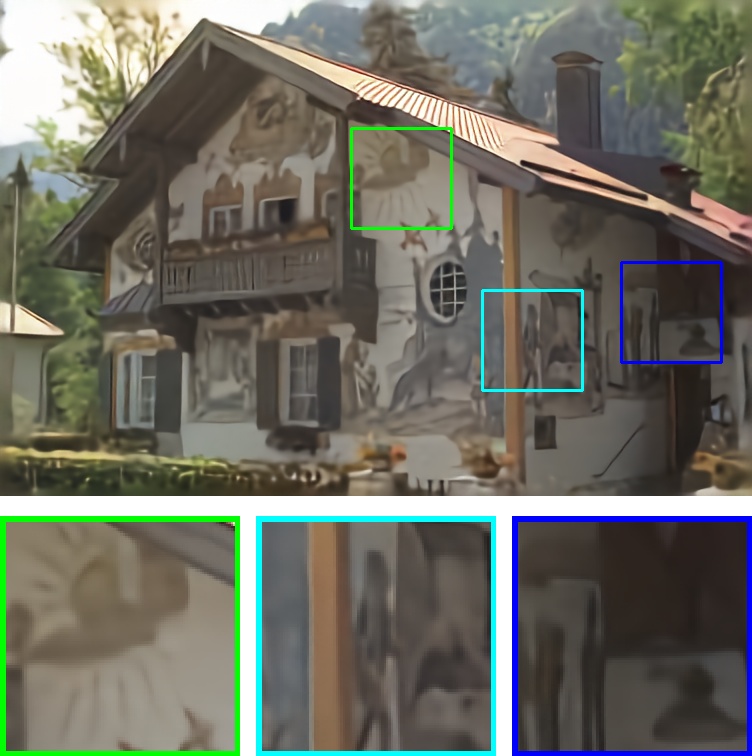}
\end{minipage}
\end{minipage}\par\medskip

\begin{minipage}{1.0\textwidth}
\begin{minipage}{0.195\textwidth}\center{\scriptsize {bpp / PSNR / MS-SSIM}}\end{minipage}
\begin{minipage}{0.195\textwidth}\center{\scriptsize{0.122 / 23.82 / 0.852}}\end{minipage}
\begin{minipage}{0.195\textwidth}\center{\scriptsize{0.137 / 24.10 / 0.889}}\end{minipage}
\begin{minipage}{0.195\textwidth}\center{\scriptsize{0.119 / 24.70 / 0.869}}\end{minipage}
\begin{minipage}{0.195\textwidth}\center{\scriptsize{0.105 / 23.26 / 0.912}}\end{minipage}
\end{minipage}\par\medskip
\caption{Decoding images produced by different compression systems. From the left to right: ground-truth, JPEG 2000, Ball{\'e} et al.~\cite{balle2016end}, BPG and ours. In general, our model achieves the best visual quality, demonstrating the superiority of our model in preserving both sharp edges and detailed textures. (Best viewed on screen in color)} \label{visual_results}
\end{figure*}

\vspace{-.1in}
\subsection{Visual Quality Evaluation}
Quantitative evaluation is conducted to assess the visual quality of decoding images by different methods.
Among deep models, most existing methods except Ball{\'e} et al.~\cite{balle2016end} do not provide either source codes or decoding images.
Among image coding standards, JPEG 2000 and BPG are superior to JPEG by quantitative metrics.
Thus, we compare Ours(MS-SSIM) with JPEG 2000, BPG, Ball{\'e} et al.~\cite{balle2016end} in our experiment.

\begin{figure}[htp]
\centering
\begin{minipage}{1.0\linewidth}
\begin{minipage}{0.32\textwidth}\center{\scriptsize{Original}}\end{minipage}
\begin{minipage}{0.32\textwidth}\center{\scriptsize{Ours(MSE)}}\end{minipage}
\begin{minipage}{0.32\textwidth}\center{\scriptsize{Ours(MS-SSIM)}}\end{minipage}
\end{minipage}\par\medskip
\begin{minipage}{1.0\linewidth}
\begin{minipage}{0.32\textwidth}
\includegraphics[width=1.0\textwidth]{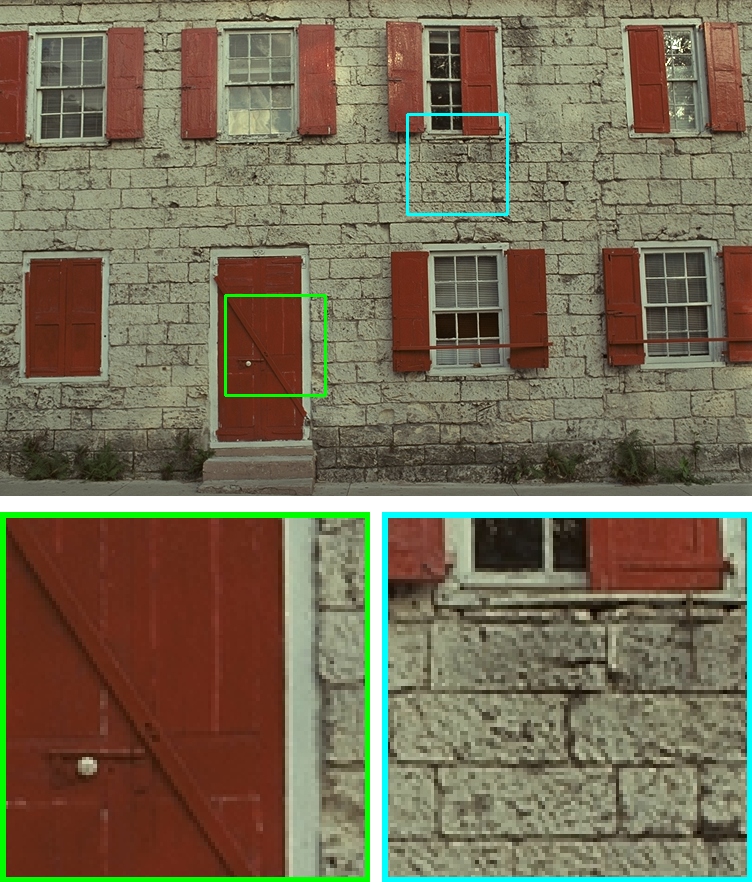}
\end{minipage}
\begin{minipage}{0.32\textwidth}
\includegraphics[width=1.0\textwidth]{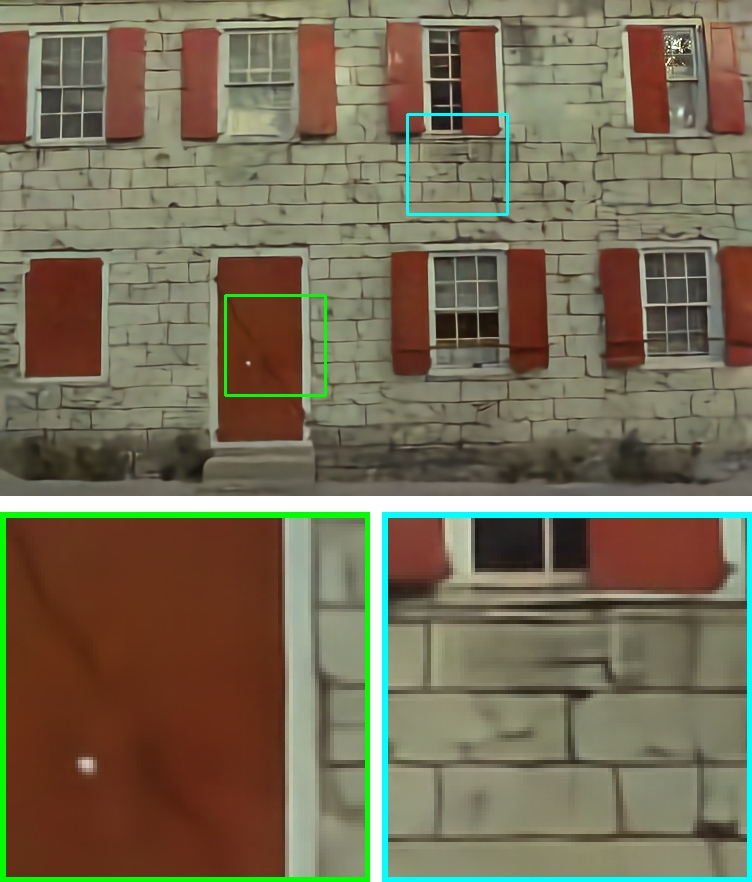}
\end{minipage}
\begin{minipage}{0.32\textwidth}
\includegraphics[width=1.0\textwidth]{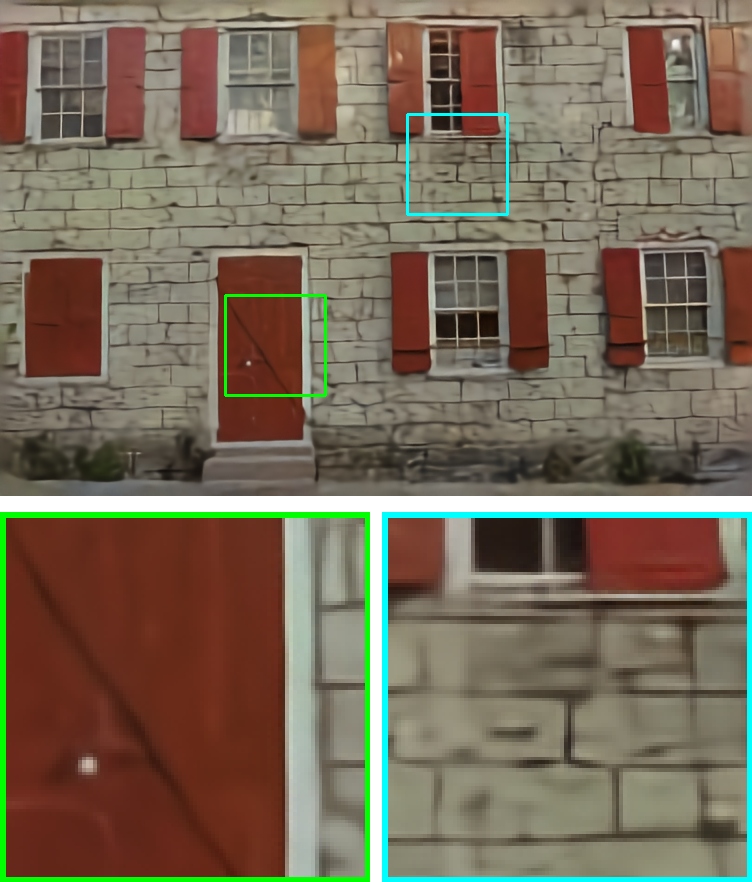}
\end{minipage}
\end{minipage}\par\medskip

\begin{minipage}{1.0\linewidth}
\begin{minipage}{0.32\textwidth}\center{\tiny{bpp / PSNR / MS-SSIM}}\end{minipage}
\begin{minipage}{0.32\textwidth}\center{\tiny{0.104 / 24.29 / 0.876}}\end{minipage}
\begin{minipage}{0.32\textwidth}\center{\tiny{0.107 /23.35 / 0.892}}\end{minipage}
\end{minipage}\par\medskip

\begin{minipage}{1.0\linewidth}
\begin{minipage}{0.32\textwidth}
\includegraphics[width=1.0\textwidth]{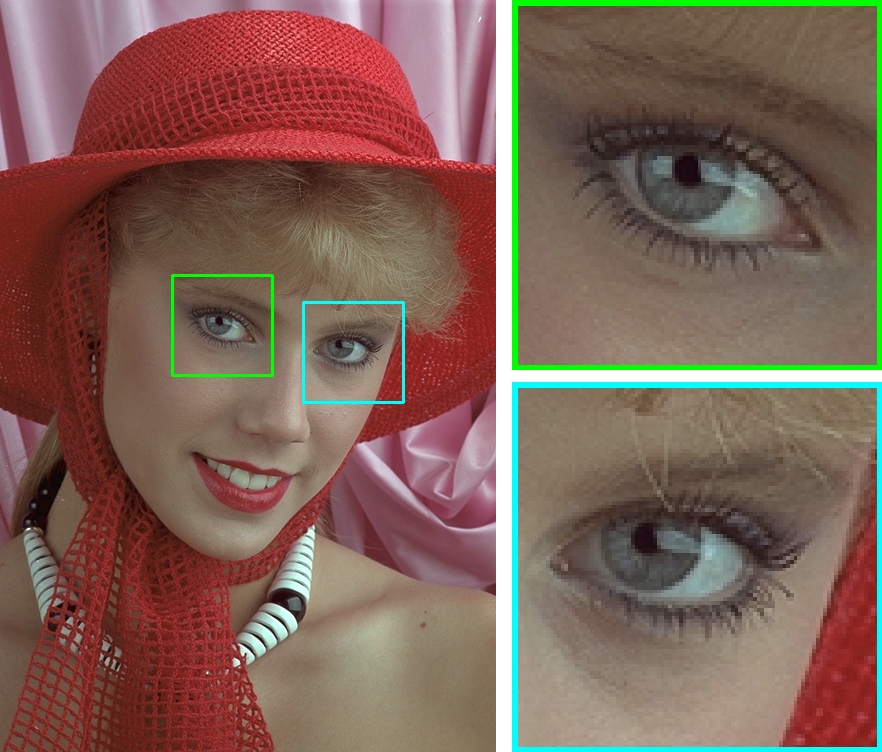}
\end{minipage}
\begin{minipage}{0.32\textwidth}
\includegraphics[width=1.0\textwidth]{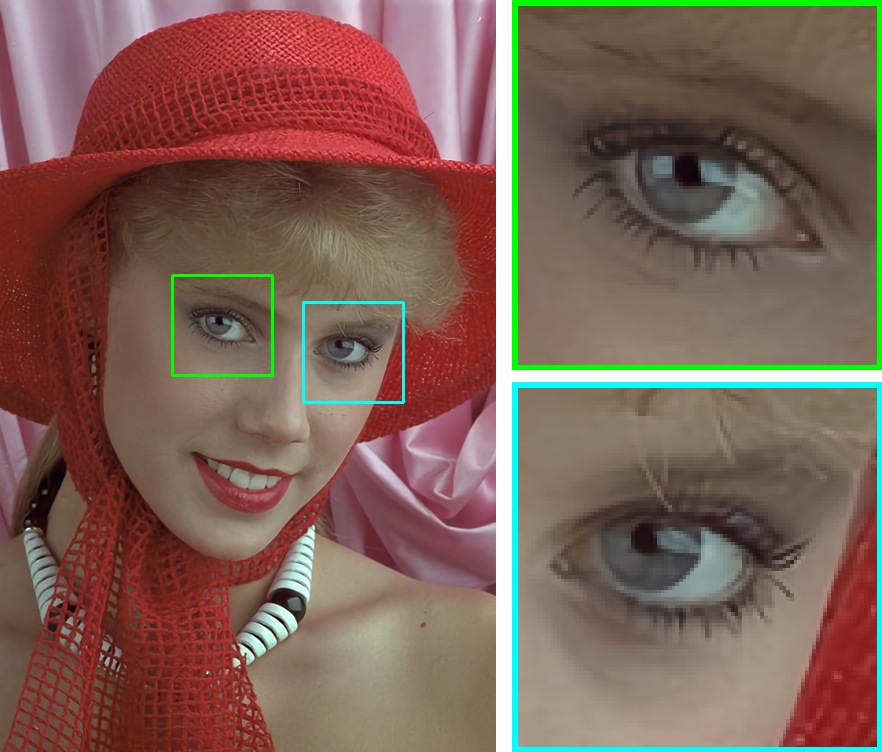}
\end{minipage}
\begin{minipage}{0.32\textwidth}
\includegraphics[width=1.0\textwidth]{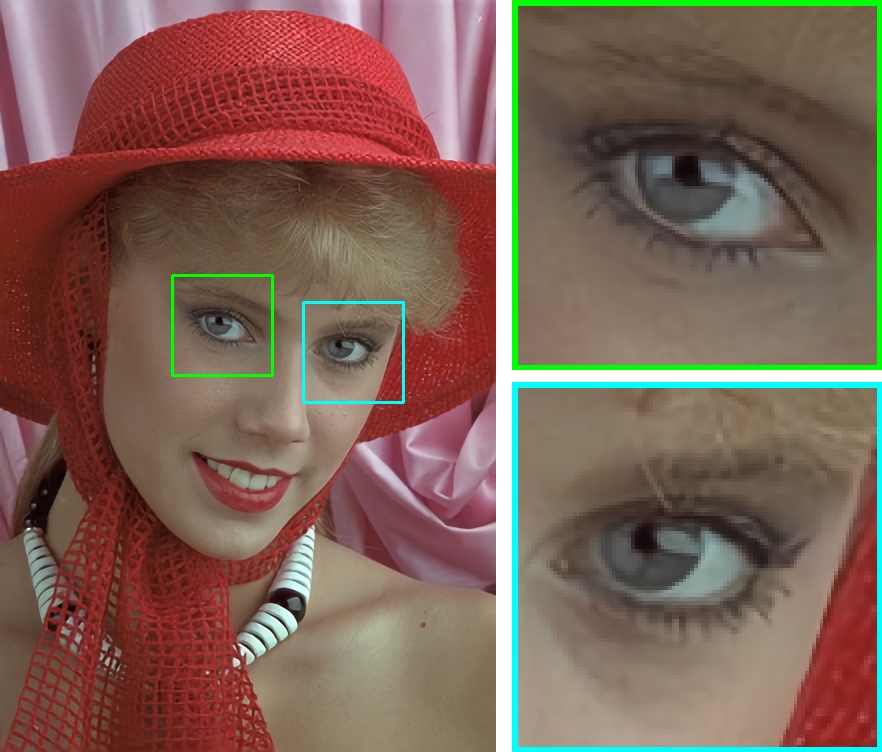}
\end{minipage}
\end{minipage}\par\medskip

\begin{minipage}{1.0\linewidth}
\begin{minipage}{0.32\textwidth}\center{\tiny{bpp / PSNR / MS-SSIM}}\end{minipage}
\begin{minipage}{0.32\textwidth}\center{\tiny{0.560 / 34.11 / 0.979}}\end{minipage}
\begin{minipage}{0.32\textwidth}\center{\tiny{0.527 / 33.08 / 0.984}}\end{minipage}
\end{minipage}
\caption{{Decoding} images produced by our models optimized with MSE and MS-SSIM, respectively. Ours(MS-SSIM) exhibits better textures  at lower bpp but may slightly obscure small sharp edges.} \label{fig:cmp_visual}
\end{figure}

Fig.~\ref{visual_results} shows the decoding images of competing methods on five Kodak images.
Visual artifacts, e.g., blurring and ringing, can still be observed from the results of JPEG 2000 and BPG.
Ball{\'e} et al.{~\cite{balle2016end}} is effective in suppressing ringing artifacts, but is limited in handling small-scale details, and also suffers from blurring and smoothing effect at lower bpp.
In contrast, the results by our method exhibit much less noticeable artifacts and are visually much
more pleasing.
More importantly, due to the introduction of importance map based {bit length} allocation, our method is more effective in retaining salient structure and fine details in comparison to the competing methods.

{
Fig.~\ref{fig:cmp_visual} shows the visual comparison between the proposed methods optimized by MS-SSIM and MSE, respectively.
At lower bpp, Ours(MSE) performs well in preserving sharp strong edges and smooth textures, while Ours(MS-SSIM) is superior in  keeping small-scale textures and weak edges.
However, at higher bpp, Ours(MS-SSIM) fails to reconstruct parts of small-scale edges, e.g., the eyelash in bottom-right of Fig.~\ref{fig:cmp_visual}.
One possible explanation is that MS-SSIM is designed for measuring multi-scale similarity.
As a result, the small edges are usually ignored at the large scale, which inevitably diminishes the contribution of small-scale edges in the metric. %
%
}

%
%
%


\subsection{Ablation Studies}
In this section, we separately test the effect of three components, i.e., the channel-wise multi-valued quantization, importance map and TCAE, with ablation studies.
For a fair comparison, we simply reuse all the parameters for training the 7 CWIC model in Sec.~\ref{sec:expriment_setup}.
$\mathcal{S}=\{\gamma,r,\mathbf{w}_{e}^0,\mathbf{w}_{d}^0\}$ denotes a set of parameters to train a CWIC.
Here, $\mathbf{w}_{e}^0$ and $\mathbf{w}_{d}^0$ are separately the initial weights of encoder and decoder.
In the experiments, all the ablation variant models are trained on the 7 parameter sets, i.e. $\mathcal{S}_1,\ldots,\mathcal{S}_7$, by MS-SSIM distortion loss and tested on Kodak dataset.

\subsubsection{Channel-wise multi-valued quantization}

We consider three other variants for the learnt channel-wise multi-valued quantization (LCMQ), i.e., (1) the learnt multi-value quantization (LMQ) with all the channels sharing the same quantization function, (2) the fixed multi-valued quantization (FMQ) with all the quantization levels are fixed and (3) the binarization function (BIN) used in~\cite{li2017learning}.
The quantization levels for all the variants except for BIN are set to be $8$.
For FMQ, we adopt the uniform multi-valued quantization used to initialize LCMQ in Sec.~\ref{sec:lcmq}.
For BIN, 1 bit instead of 3 bits is used to represent a code in $\mathbf{c}$ before entropy coding, and we increase the number of channel of $\mathbf{o}$ from $32$ to $96$ to compensate for the total number of bits to represent $\mathbf{o}$.
%

\begin{table}[!htb]
\scriptsize
\begin{center}
\caption{Quantization error of four quantization functions, i.e., LCMQ, LMQ, FMQ and BIN, on 7 parameter sets.}
\begin{tabular}{|c|c|c|c|c|}
\hline
Set & \specialcell{LCMQ} & \specialcell{LMQ }& \specialcell{FMQ} &  \specialcell{BIN }\\
\hline
$\mathcal{S}_1$&\textbf{$0.97\times 10^{-3}$}&$1.12\times 10^{-3}$&$2.21\times 10^{-3}$&$3.29\times 10^{-2}$\\
$\mathcal{S}_2$&\textbf{$0.92\times 10^{-3}$}&$1.01\times 10^{-3}$&$1.78\times 10^{-3}$&$3.13\times 10^{-2}$\\
$\mathcal{S}_3$&\textbf{$0.79\times 10^{-3}$}&$1.03\times 10^{-3}$&$1.93\times 10^{-3}$&$3.24\times 10^{-2}$\\
$\mathcal{S}_4$&\textbf{$0.78\times 10^{-3}$}&$0.93\times 10^{-3}$&$2.09\times 10^{-3}$&$3.15\times 10^{-2}$\\
$\mathcal{S}_5$&\textbf{$0.92\times 10^{-3}$}&$1.00\times 10^{-3}$&$1.89\times 10^{-3}$&$3.07\times 10^{-2}$\\
$\mathcal{S}_6$&\textbf{$1.01\times 10^{-3}$}&$1.20\times 10^{-3}$&$2.08\times 10^{-3}$&$3.21\times 10^{-2}$\\
$\mathcal{S}_7$&\textbf{$0.93\times 10^{-3}$}&$1.03\times 10^{-3}$&$1.81\times 10^{-3}$&$3.18\times 10^{-2}$\\
\hline
\hline
AVE&\textbf{$0.90\times 10^{-3}$}&$1.05\times 10^{-3}$&$1.97\times 10^{-3}$&$3.18\times 10^{-2}$\\
\hline
\end{tabular}
\label{table:quantization}
\end{center}
\end{table}

By replacing LCMQ with each of the three other variants, we re-train the CWIC model on the 7 parameter sets with the MS-SSIM distortion loss.
Two metrics, i.e.,  distortion of the decoding images and quantization error, are reported based on the Kodak dataset.
The quantization error is defined as the MSE between the output of encoder $\mathbf{e}$ and the quantized code $Q(\mathbf{e})$.
Table~\ref{table:quantization} lists the quantization error of the four quantization functions.
Unsurprisingly, the BIN in~\cite{li2017learning} obtains the largest quantization error due to that it has only two quantization levels.
Among the multi-valued quantization functions, LCMQ and LMQ get much lower quantization error than FQM, indicating that the learning of quantization is indeed helpful in decreasing quantization error when the quantization levels are the same.
Nonetheless, LCMQ achieves the lowest quantization error on all the 7 parameter sets, demonstrating the usefulness of learning quantization function for each channel.

%
\begin{figure}
\center
\begin{minipage}{0.49\linewidth}
\center
\includegraphics[width=1.0\linewidth]{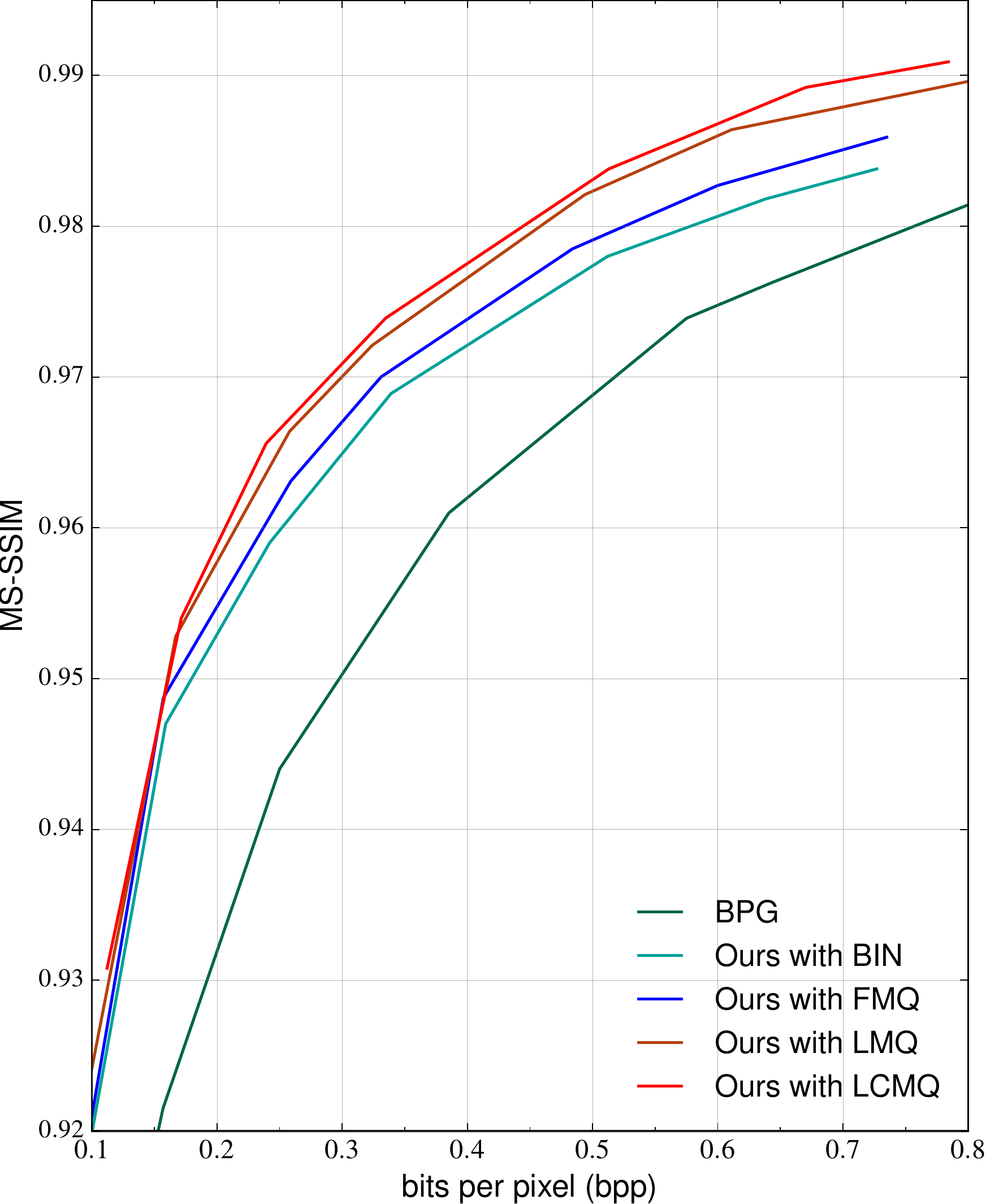}
\scriptsize{(a) Quantization}
\end{minipage}
\begin{minipage}{0.48\linewidth}
\center
\includegraphics[width=1.0\linewidth]{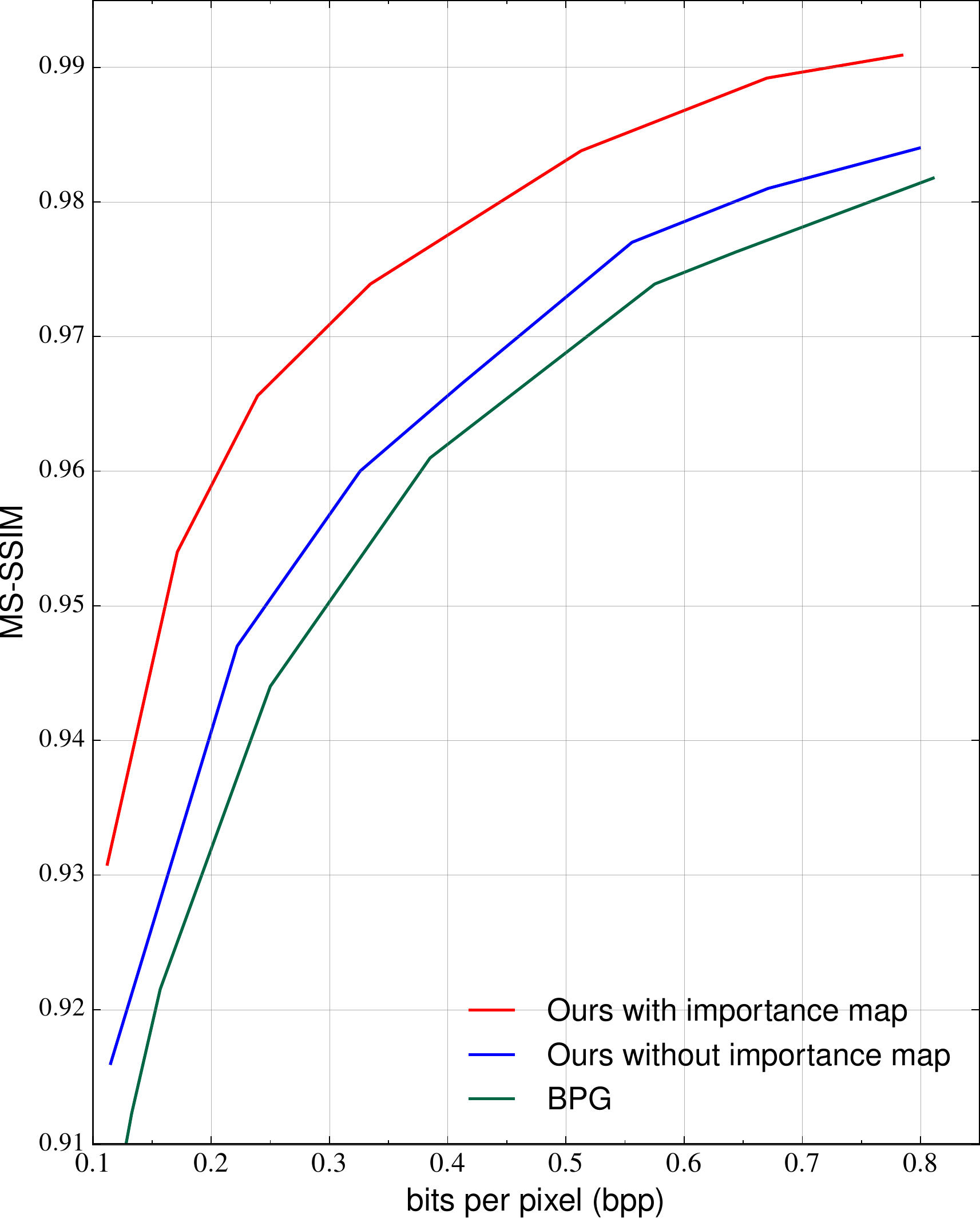}
\scriptsize{(b) Importance Map}
\end{minipage}\par\medskip
\caption{{Rate-distortion curves for ablation studies on Kodak. (a) comparison of four quantization variants, i.e., LCMQ, LMQ, FMQ and BIN. (b) comparison of our CWIC models with and without importance map.}}\label{fig:ablation}
\end{figure}
\begin{figure}
\center
\begin{minipage}{0.410\linewidth}
\includegraphics[width=1.0\linewidth]{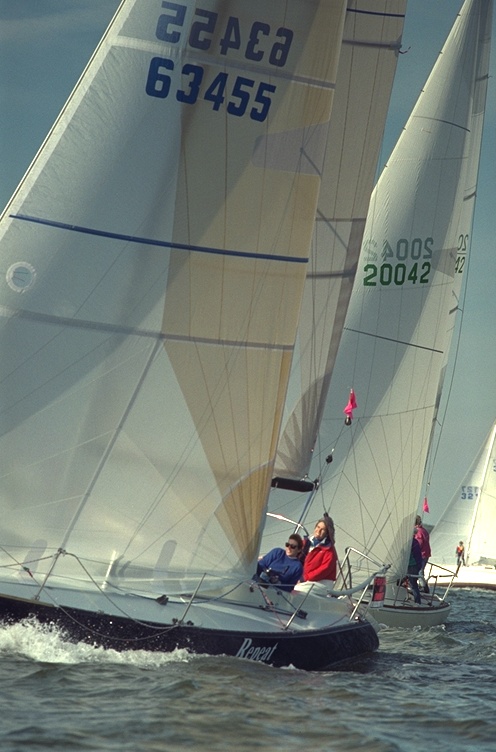}
\centering{\scriptsize{Original}}
\end{minipage}
\begin{minipage}{0.563\linewidth}
\begin{minipage}{1.0\linewidth}
\begin{minipage}{0.32\linewidth}
\includegraphics[width=1.0\linewidth]{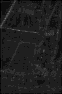}
\end{minipage}
\begin{minipage}{0.32\linewidth}
\includegraphics[width=1.0\linewidth]{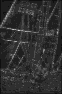}
\end{minipage}
\begin{minipage}{0.32\linewidth}
\includegraphics[width=1.0\linewidth]{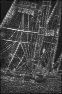}
\end{minipage}
\end{minipage}\par\medskip

\begin{minipage}{0.32\linewidth}\centering{\scriptsize{0.151 bpp}}\end{minipage}
\begin{minipage}{0.32\linewidth}\centering{\scriptsize{0.239 bpp}}\end{minipage}
\begin{minipage}{0.32\linewidth}\centering{\scriptsize{0.347 bpp}}\end{minipage}\par\medskip

\begin{minipage}{1.0\linewidth}
\begin{minipage}{0.32\linewidth}
\includegraphics[width=1.0\linewidth]{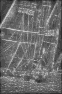}
\end{minipage}
\begin{minipage}{0.32\linewidth}
\includegraphics[width=1.0\linewidth]{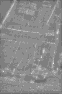}
\end{minipage}
\begin{minipage}{0.32\linewidth}
\includegraphics[width=1.0\linewidth]{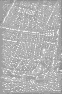}
\end{minipage}
\end{minipage}\par\medskip

\begin{minipage}{0.32\linewidth}\centering{\scriptsize{0.510 bpp}}\end{minipage}
\begin{minipage}{0.32\linewidth}\centering{\scriptsize{0.639 bpp}}\end{minipage}
\begin{minipage}{0.32\linewidth}\centering{\scriptsize{0.814 bpp}}\end{minipage}
\end{minipage}\par\medskip
\caption{Visualization of the importance maps at 6 kinds of bpps. Left: ground-truth. Right: importance maps ranging from $0.151$ to $0.814$ bpp.}\label{fig:imp}
\end{figure}

\begin{figure*}
\center
\begin{minipage}{0.24\textwidth}
\center
\includegraphics[width=1.0\linewidth]{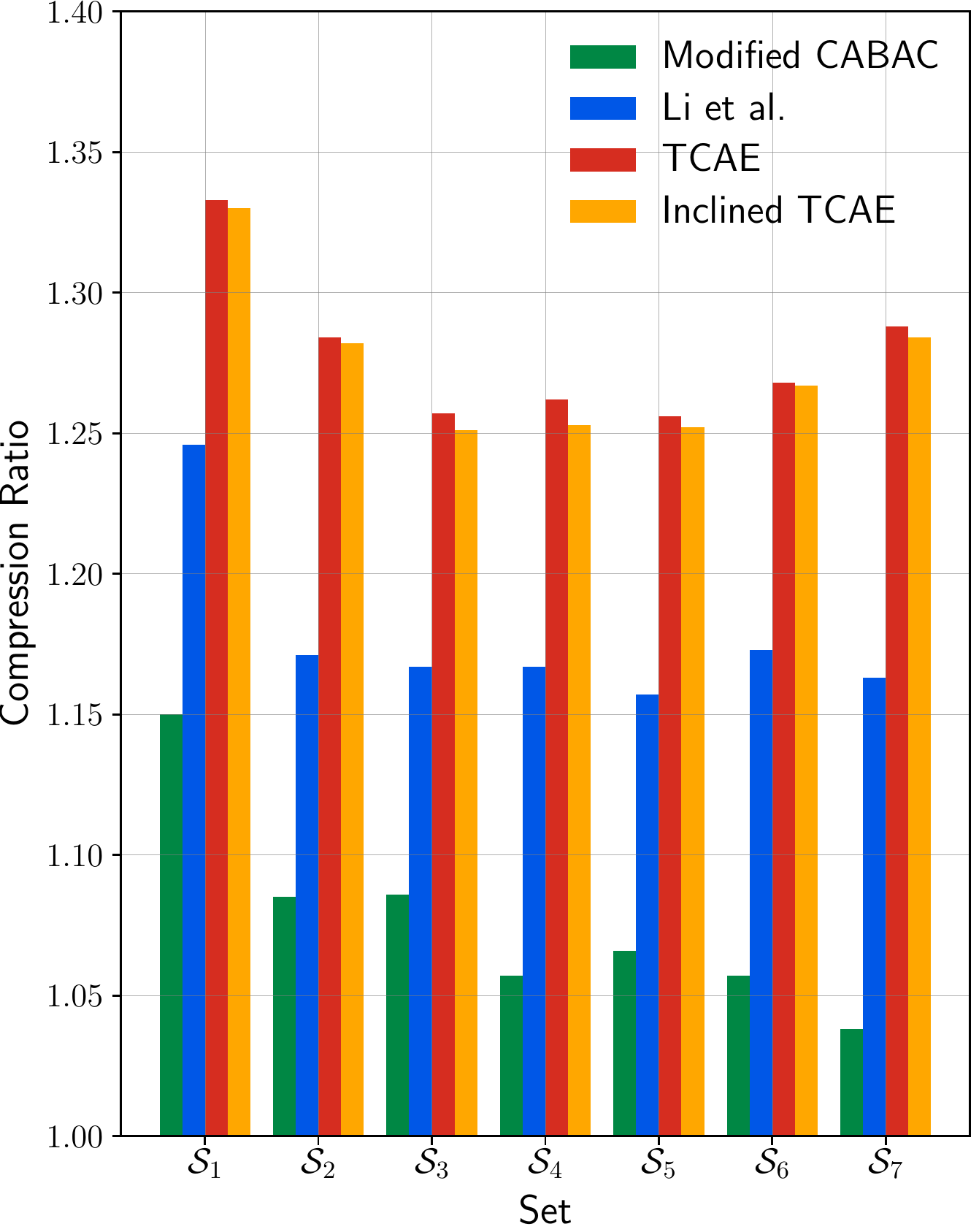}
\scriptsize{(a)}
\end{minipage}
\begin{minipage}{0.24\textwidth}
\center
\includegraphics[width=1.0\linewidth]{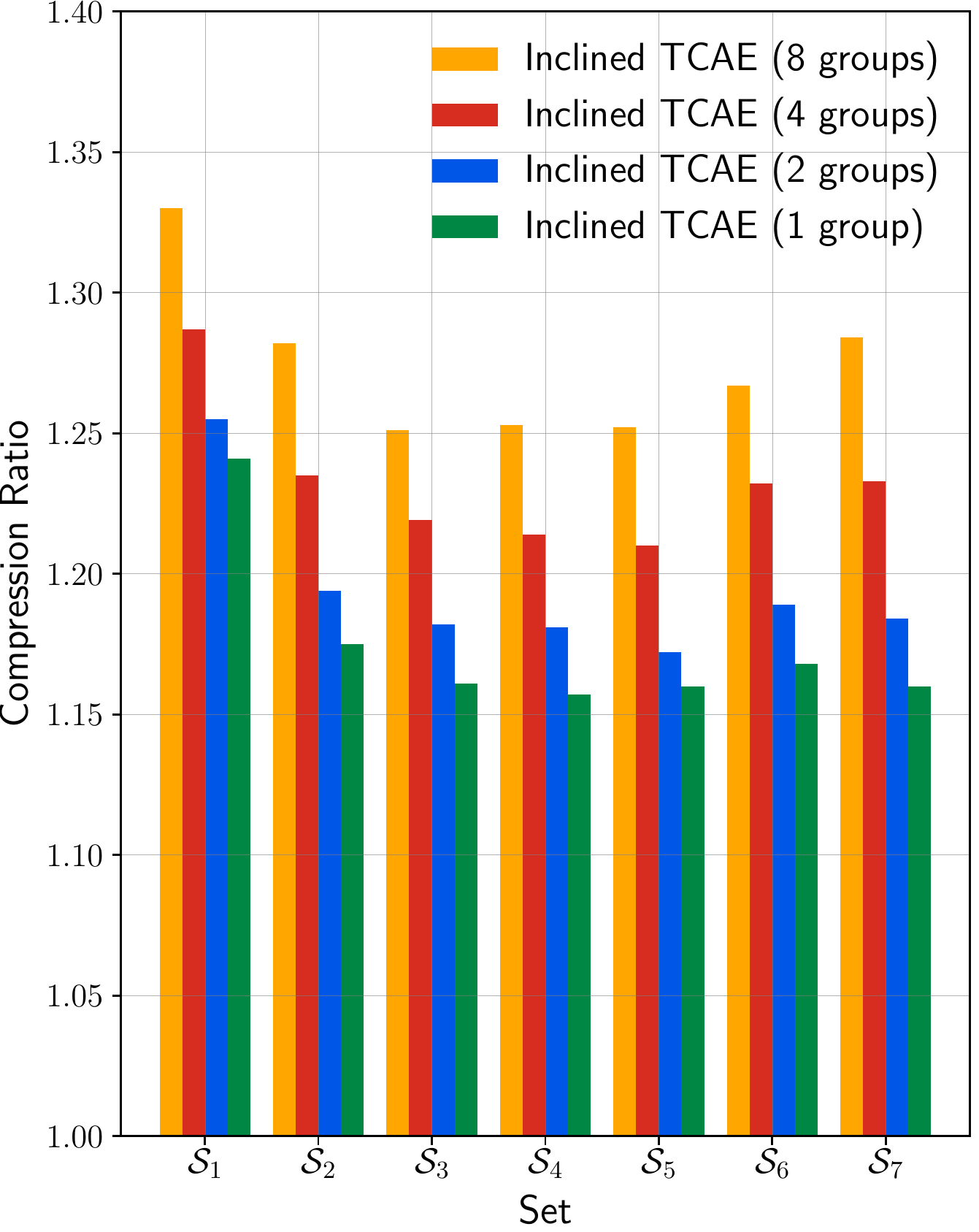}
\scriptsize{(b)}
\end{minipage}
\begin{minipage}{0.24\textwidth}
\center
\includegraphics[width=1.0\linewidth]{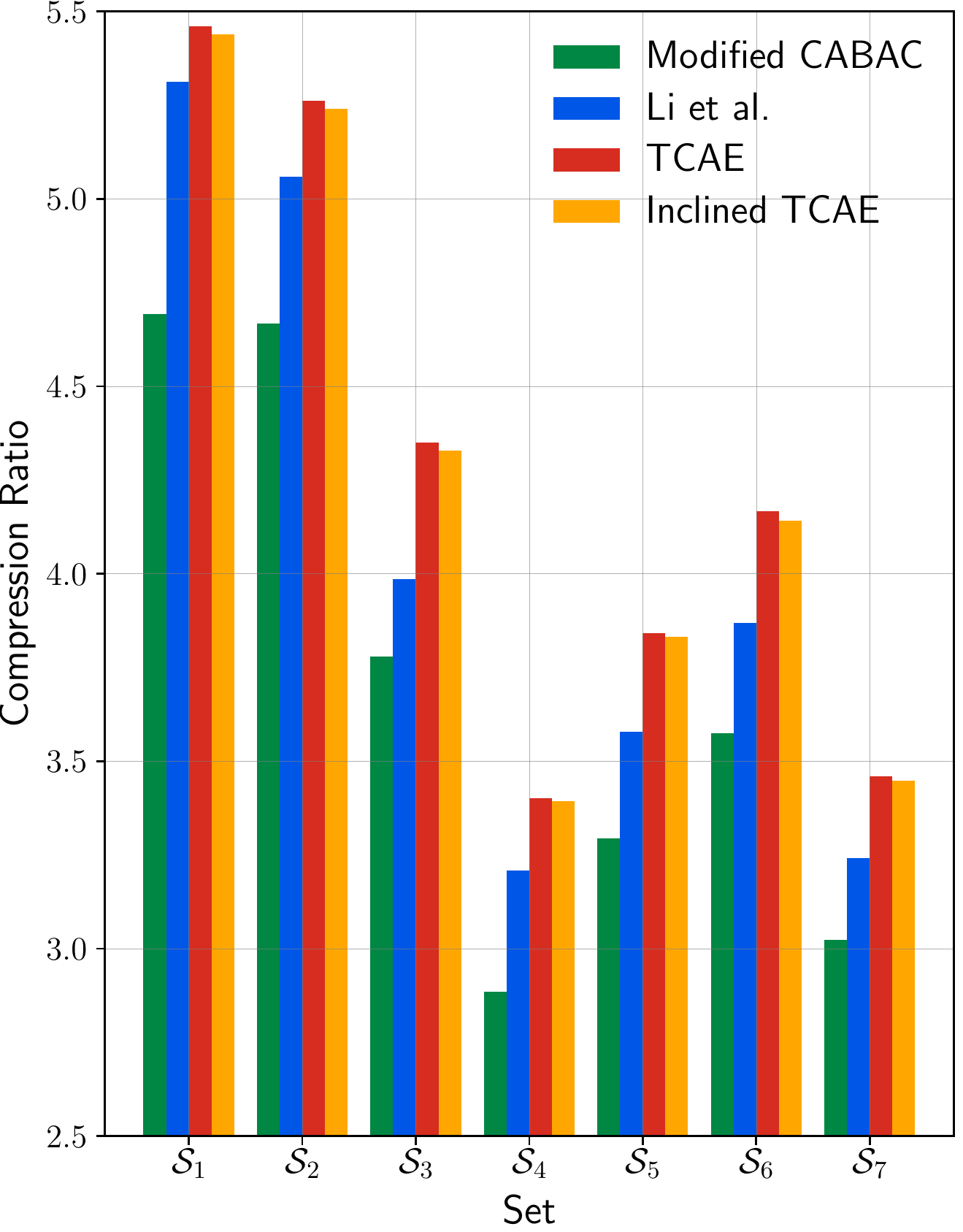}
\scriptsize{(c)}
\end{minipage}
\begin{minipage}{0.24\textwidth}
\center
\includegraphics[width=1.0\linewidth]{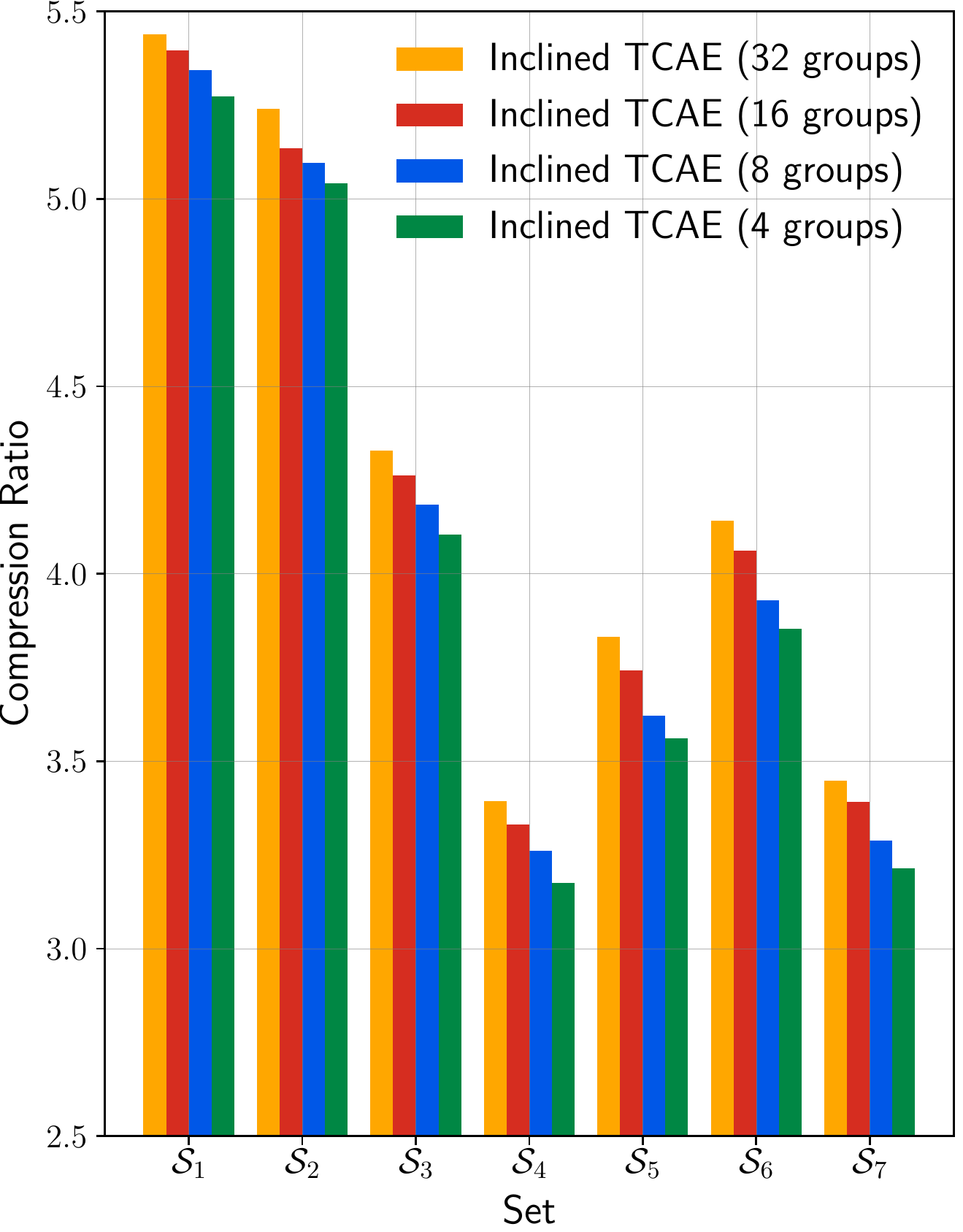}
\scriptsize{(d) }
\end{minipage}\par\medskip
\caption{{Lossless compression ratio of entropy prediction models. The data used to test the entropy prediction models are generated by our CWIC with 7 different parameter sets.
(a) and (c) respectively show the results of the four entropy prediction models on the {code $\mathbf{o}^\prime$} and the quantized importance map $\mathbf{p}^\prime$. %
(b) and (d) respectively show the results of inclined TCAE with different number of groups on {$\mathbf{o}^\prime$} and $\mathbf{p}^\prime$.  }}\label{fig:lossless}
\end{figure*}

Fig.~\ref{fig:ablation}(a) shows the rate-distortion curves of our CWIC with the four quantization variants on the Kodak dataset.
It can be seen that LCMQ gets the best performance followed by LMQ, while BIN exhibits the worst  performance.
We note that the rate-distortion results are consistent with the quantization error, where lower quantization error corresponds to lower distortion.
Therefore, the quantization loss in Sec.~\ref{loss}, which is introduced to minimize quantization error, can be expectantly beneficial to rate-distortion performance and to ease the gradient issue of quantization in back-propagation.


\subsubsection{Importance map}

To assess the effect of importance map, we introduce a baseline model by removing the importance map subnet from CWIC, and set the compression rate by adjusting the number of channels $n$ in the code $\mathbf{c}$.
In the experiments, we adopt $n=4,8,12,16,20,24$.
%
%
As shown in Fig.~\ref{fig:ablation} (b), the introduction of importance map can result in much better performance, clearly demonstrating the effectiveness of the spatially variant bit length allocation scheme.
It is also interesting to note that, due to the learnt channel-wise multi-valued quantization, our CWIC can also outperform BPG by MS-SSIM.

To reveal the role of importance map, we visualize the importance maps of a representative image {at 6 kinds of bpps}.
The importance maps are mapped from the range of [0,1] to [0,255] and shown as gray images.
From Fig.~\ref{fig:imp}, it can be observed that, at lower bpp the learnt importance map only allocates more codes to the strong edges.
Along with the increase of bit rate, more codes will be allocated to weak edges and mid-scale textures.
With the further increase of bit rate, small-scale textures such as the wave start to be allocated with more codes.
Thus, the learnt importance map is consistent with the human visual perception, which also explains the superiority of our model in preserving the structure, edges and textures.

\subsubsection{Inclined TCAE for encoding and decoding}\label{sec:exp-tcae}

For entropy modeling, we compare TCAE and inclined TCAE with two counterparts, i.e., our pioneer work (convolutional entropy prediction model)~\cite{li2017learning} and a modified CABAC~\cite{marpe2003context} to compress the {code $\mathbf{o}^\prime=(1+\mathbf{o})\circ \mathbf{m}$} and the quantized importance map $\mathbf{p}^\prime = QI(\mathbf{p})$.
%
%
%
%
%
In~\cite{li2017learning}, a $4\times 5\times 5$ cuboid is extracted for the symbol $o^\prime_{r,p,q}$ with the importance mask $m_{r,p,q}=1$, while a $1\times5\times5$ cuboid is extracted for symbols in $\mathbf{p}^\prime$.
%
%
For CABAC~\cite{marpe2003context}, only two nearby symbols, i.e., $o^\prime_{k,i-1,j}$ and $o^\prime_{k,i,j-1}$, are considered as the context of $o^\prime_{k,i,j}$.
In our modification, we further consider the relation across channels and also include the symbol $o^\prime_{k-1,i,j}$ into the context.
For the quantized importance map with only $1$ channel, the context is kept the same as CABAC.

%
%

Fig.~\ref{fig:lossless} (a) and (c) show the lossless compression ratios of the four context modeling methods on $\mathbf{o}^\prime$ and the quantized importance map $\mathbf{p}^\prime$.
In particular, our TCAE and inclined TCAE achieve comparable compression ratio and outperform the two counterparts.
Moreover, convolutional entropy prediction~\cite{li2017learning} is also superior to modified CABAC.
Using {the code $\mathbf{o}^\prime$} as an example, the context sizes of TCAE, inclined TCAE, convolutional entropy prediction~\cite{li2017learning}, and modified CABAC are $k\times37\times37$, $k\times37\times37$, $4\times5\times5$ and $3$, respectively.
Here, $k$ is the channel index of {$\mathbf{o}^\prime$}.
By comparing the four context modeling methods, it can be seen that larger context generally is beneficial to better entropy prediction.
%
%
%

In Fig.~\ref{fig:lossless} (b) and (d), we test inclined TCAE with different number of groups of feature maps, i.e., $1$, $2$, $4$ and $8$ groups for the {code} $\mathbf{o}^\prime$ and $4$, $8$, $16$ and $32$ groups for the quantized importance map $\mathbf{p}^\prime$.
Naturally, more groups do benefit the performance of inclined TCAE but also increase the model parameters and running time.
%
%
In our implementation, we adopt $8$ groups for $\mathbf{o}^\prime$ and $32$ groups for $\mathbf{p}^\prime$.
{
It is worth noting that, on the parameter sets $\mathcal{S}_1$ and $\mathcal{S}_2$, the symbols in the quantized importance maps mainly are some small values, i.e., 0, 1, 2, 3.
Consequently, the quantized importance map is small by entropy and results in larger compression ratio.
}
\begin{table*}[]
\scriptsize
\begin{center}
\caption{Running time ($s$) of different entropy prediction models on the codes and quantized importance maps generated by CWIC trained on seven different parameter sets. 
}
\begin{tabular}{|C{1.3cm}|C{1.3cm}|C{1.3cm}|C{1.3cm}|C{1.3cm}|C{1.3cm}|C{1.3cm}|C{1.3cm}|C{1.3cm}|C{1.3cm}|}
\hline
\multirow{2}{*}{Set} & \multirow{2}{*}{ Task} & \multicolumn{2}{c|}{Modified CABAC }& \multicolumn{2}{c|}{Li et al. } & \multicolumn{2}{c|}{TCAE } & \multicolumn{2}{c|}{Inclined TCAE }\\
\cline{3-10}
&& $\mathbf{o}^\prime$ & $\mathbf{p}^\prime$ & $\mathbf{o}^\prime$ & $\mathbf{p}^\prime$ & $\mathbf{o}^\prime$ & $\mathbf{p}^\prime$ & $\mathbf{o}^\prime$ & $\mathbf{p}^\prime$\\
\hline
\multirow{2}{*}{$\mathcal{S}_1$}&Encoding&0.001&0.00003&0.092&0.021&0.021&0.005&0.021&0.005\\
\cline{2-10}
&Decoding&0.001&0.00003&32.3&6.73&202.8&6.58&0.923&0.161\\
\hline
\multirow{2}{*}{$\mathcal{S}_2$}&Encoding&0.001&0.00003&0.178&0.021&0.021&0.005&0.021&0.005\\
\cline{2-10}
&Decoding&0.001&0.00003&58.5&6.73&202.8&6.58&0.923&0.161\\
\hline
\multirow{2}{*}{$\mathcal{S}_3$}&Encoding&0.001&0.00003&0.368&0.021&0.021&0.005&0.021&0.005\\
\cline{2-10}
&Decoding&0.001&0.00003&123.2&6.73&202.8&6.58&0.923&0.161\\
\hline
\multirow{2}{*}{$\mathcal{S}_4$}&Encoding&0.001&0.00003&0.483&0.021&0.021&0.005&0.021&0.005\\
\cline{2-10}
&Decoding&0.001&0.00003&163.5&6.73&202.8&6.58&0.923&0.161\\
\hline
\multirow{2}{*}{$\mathcal{S}_5$}&Encoding&0.001&0.00003&0.665&0.021&0.021&0.005&0.021&0.005\\
\cline{2-10}
&Decoding&0.001&0.00003&225.1&6.731&202.8&6.58&0.923&0.161\\
\hline
\multirow{2}{*}{$\mathcal{S}_6$}&Encoding&0.001&0.00003&0.782&0.021&0.021&0.005&0.021&0.005\\
\cline{2-10}
&Decoding&0.001&0.00003&250.6&6.73&202.8&6.58&0.923&0.161\\
\hline
\multirow{2}{*}{$\mathcal{S}_7$}&Encoding&0.001&0.00003&0.931&0.021&0.021&0.005&0.021&0.005\\
\cline{2-10}
&Decoding&0.001&0.00003&288.2&6.73&202.8&6.58&0.923&0.161\\
\hline
\end{tabular}
\label{table:lossless-time}
\end{center}
\end{table*}

\subsubsection{Running time}

Using a computer with a Intel(R) Xeon(R) Processor E5-2620 v4, 64 GB of RAM and a NVIDIA TITAN Xp GPU, we test the running time (in seconds, $s$) of our method on the Kodak dataset with the image size $752 \times 496$ (or $496 \times 752$).
{The running time to generate the code $\mathbf{o}$ and to reconstruct input image from $\mathbf{o}$ are  $0.024$ and $0.032$ $s$, respectively.}

We further test the running time of lossless encoding and decoding of
the {code $\mathbf{o}^\prime$} and the quantized importance map $\mathbf{p}^\prime$.
In particular, we consider four models, i.e., modified CABAC, convolutional entropy prediction {model}~\cite{li2017learning}, TCAE, and inclined TCAE.
The running time of modified CABAC is tested on CPU, while the running time of the other three models are accelerated with GPU in the caffe framework.

Table~\ref{table:lossless-time} lists  the running time of the 4 lossless compression models.
For both encoding and decoding, modified CABAC is the  fastest model, but fails to exploit large context.
In terms of encoding, TCAE and inclined TCAE are the second fastest methods, and are effective in large context modeling.
However, when taking decoding into account, TCAE should {decode the symbols in a sequential order} and remains computationally inefficient.
Benefited from the inclined plane based context, inclined TCAE {is able to} parallel decode the {symbols} within each inclined planes, and thus can be more than 100$\times$ faster than TCAE for decoding.
In comparison, convolutional entropy prediction {model}~\cite{li2017learning} only utilizes limited size of context (i.e., $4\times 5\times 5$), is much inefficient for encoding, and is only comparable to TCAE for decoding.
Furthermore, the running time of convolutional entropy prediction { model}~\cite{li2017learning} gradually increases from $\mathcal{S}_1$  to $\mathcal{S}_7$ when encoding and decoding $\mathbf{o}^\prime$ but keeps the same when encoding and decoding $\mathbf{p}^\prime$.
This result can be ascribed to that convolutional entropy prediction{ model}~\cite{li2017learning} independently handles each symbol to encode.
From $\mathcal{S}_1$ to $\mathcal{S}_7$, more and more $1$s are generated in the importance mask, which indicates that more symbols in $\mathbf{o}^\prime$ are required to {be processed} by convolutional entropy prediction {model}~\cite{li2017learning}.
As for $\mathbf{p}^\prime$, all the symbols should be processed, and thus the parameter set has no effect on running time.

\vspace{-.1in}
\section{Conclusion}\label{sec:conclusion}
In this paper, we proposed a learning based content-weighted image compression framework by taking both spatial  variation  and  dependency into account.
For handling spatial variation of image content, an importance map subset is incorporated with the encoder-decoder network to produce the importance mask for locally adaptive bit rate allocation.
In addition, a learnt channel-wise multi-valued quantization is further presented to reduce quantization error as well as improve compression performance.
To exploit spatial dependency, arithmetic encoding is adopted for transforming the quantized codes and importance map into bit streams.
For better context modeling, TCAE is introduced to enlarge the context while maintaining the efficiency of the entropy prediction, and inclined TCAE is further presented to accelerate the decoding process.
Experimental results show that our CWIC performs favorably in comparison to the state-of-the-art deep image compression methods and traditional image compression standards, and is  effective  in recovering  salient  structures  and  rich  details  especially at lower bpp.


%

\vspace{-.1in}
\ifCLASSOPTIONcompsoc
  \section*{Acknowledgments}
\else
  \section*{Acknowledgment}
\fi

This work is supported in part by the NSFC Fund (61671182) and the Hong Kong RGC General Research Fund (PolyU 152212/14E). The authors would like to thank the support from NVIDIA Corporation for donating the TITAN Xp GPU used in this work.

\ifCLASSOPTIONcaptionsoff
  \newpage
\fi




\vspace{-.1in}
\bibliographystyle{IEEEtran}
\bibliography{IEEEabrv,egbib}

%

\begin{IEEEbiography}
[{\includegraphics[width=1in,height=1.25in,clip,keepaspectratio]{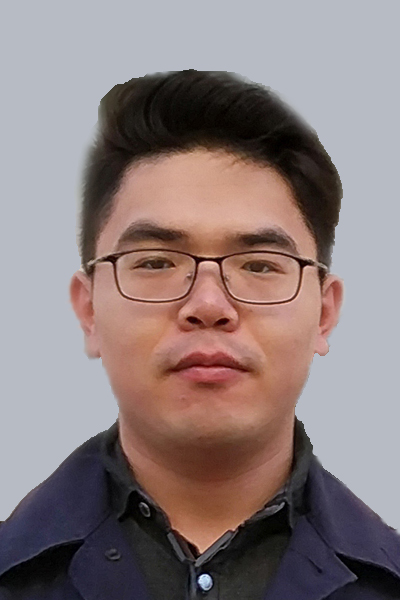}}]{Mu Li} received the BCS in Computer Science and Technology from Harbin Institute of Technology, Harbin, China, in 2015. He is the owner of Hong Kong PhD Fellowship and is currently pursuing the Ph.D. degree in Department of Computing with the Hong Kong Polytechnic University, Hong Kong, China, under the supervision of Prof. David Zhang, Prof. Jane You and Prof. Wangmeng Zuo. His research interests include deep learning and image processing.
\end{IEEEbiography}

\begin{IEEEbiography}
[{\includegraphics[width=1in,height=1.25in,clip,keepaspectratio]{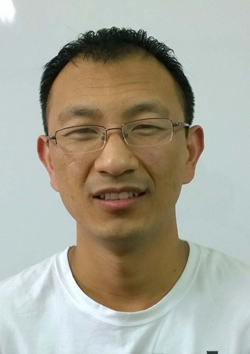}}]{Wangmeng Zuo} (M'09, SM'14) received the Ph.D. degree in computer application technology from the Harbin Institute of Technology, China, in 2007. From 2004 to 2006, he was a Research Assistant with the Department of Computing, The Hong Kong Polytechnic University. From 2009 to 2010, he was a Visiting Professor with Microsoft Research Asia. He is currently a Professor with the School of Computer Science and Technology, Harbin Institute of Technology. He has published over 80 papers in top-tier academic journals and conferences. His current research interests include image enhancement and restoration, image generation and editing, visual tracking, object detection, and image classification. He has served as a Tutorial Organizer in ECCV 2016, an Associate Editor of the IET Biometrics, and the Guest Editor of Neurocomputing, Pattern Recognition, IEEE Transactions on Circuits and Systems for Video Technology, and IEEE Transactions on Neural Networks and Learning Systems.
\end{IEEEbiography}

\begin{IEEEbiography}
[{\includegraphics[width=1in,height=1.25in,clip,keepaspectratio]{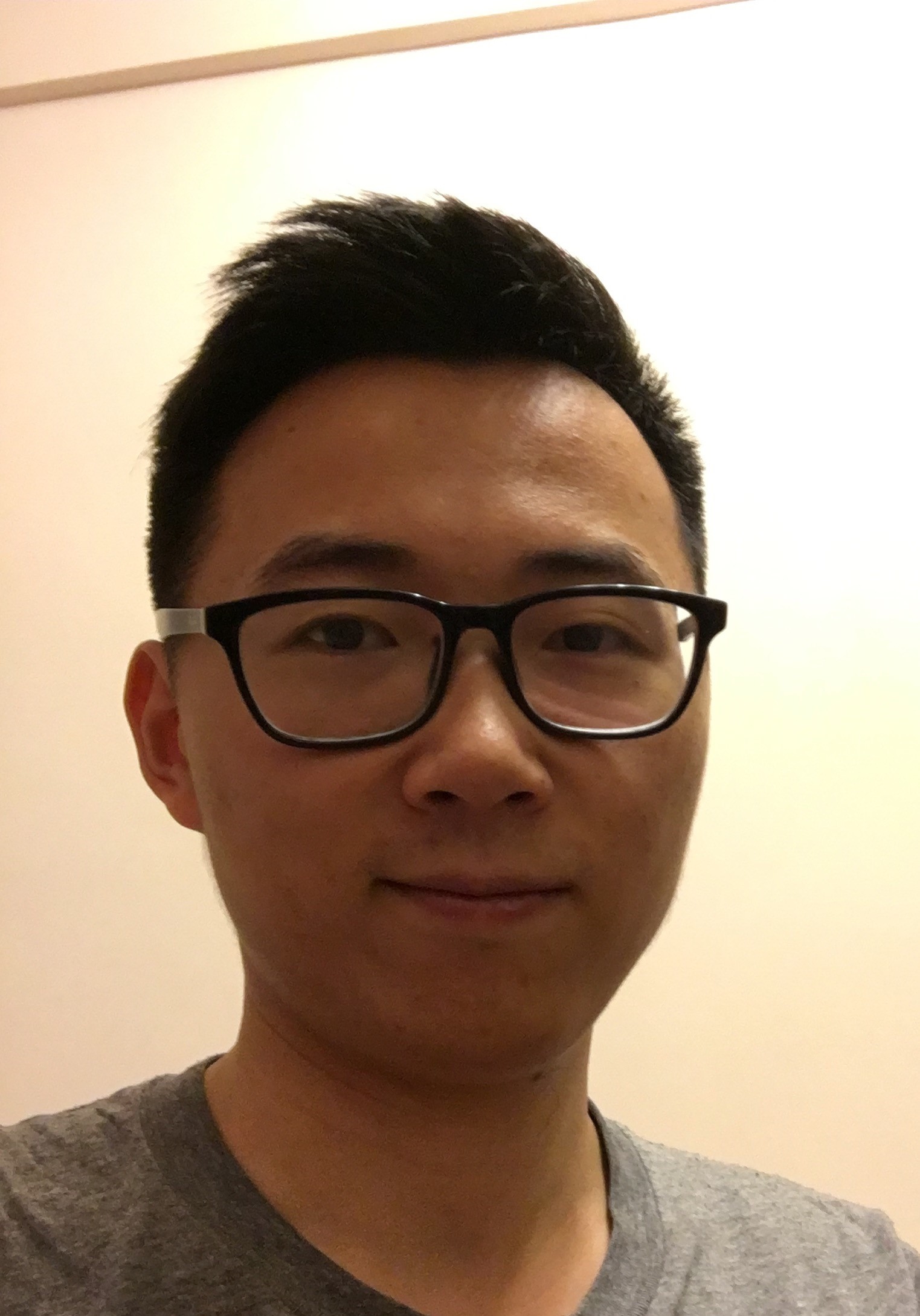}}]{Shuhang Gu} received the B.E. degree from the School of Astronautics, Beijing University of Aeronautics and Astronautics, China, in 2010, the M.E. degree from the Institute of Pattern Recognition and Artificial Intelligence, Huazhong University of Science and Technology, China, in 2013, and Ph.D. degree from the Department of Computing, The Hong Kong Polytechnic University, in 2017. He currently holds a post-doctoral position at ETH Zurich, Switzerland. His research interests include image restoration, enhancement and compression.
\end{IEEEbiography}

\begin{IEEEbiography}
[{\includegraphics[width=1in,height=1.25in,clip,keepaspectratio]{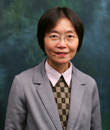}}]{Jane You} received the B.Eng. degree in electronics engineering from Xi'an Jiaotong University, Xi'an, China, in 1986, and the Ph.D. degree in computer science from La Trobe University, Melbourne, VIC, Australia, in 1992. She was a Lecturer with the University of South Australia, Adelaide SA, Australia, and a Senior Lecturer with Griffith University, Nathan, QLD, Australia, from 1993 to 2002. She is currently a Full Professor with The Hong Kong Polytechnic University, Hong Kong. Her current research interests include image processing, pattern recognition, medical imaging, biometrics computing, multimedia systems, and data mining.
\end{IEEEbiography}

\begin{IEEEbiography}
[{\includegraphics[width=1in,height=1.25in,clip,keepaspectratio]{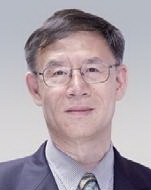}}]{David Zhang} received the Graduation degree in computer science from Peking University, China, the M.Sc. degree in computer science in 1982, and the Ph.D. degree in 1985 from the Harbin Institute of Technology (HIT), Harbin, China. In 1994, he received the second Ph.D. degree in electrical and computer engineering from the University of Waterloo,  Canada. From 1986 to 1988, he was a Postdoctoral Fellow with Tsinghua University and then an Associate Professor at the Academia Sinica, Beijing. He is currently a Chair Professor at the Hong Kong Polytechnic University and the Chinese University of Hong Kong (Shenzhen). He also serves as a Visiting Chair Professor at Tsinghua University, Beijing, and an Adjunct Professor at Peking University, Shanghai Jiao Tong University, Shanghai, China, HIT, and the University of Waterloo. His research interests are medical biometrics and pattern recognition.
\end{IEEEbiography}




\end{document}